\RequirePackage{fix-cm}
\documentclass[twocolumn]{svjour3}          
\smartqed  
\usepackage[colorlinks,linkcolor=blue,anchorcolor=blue,citecolor=blue]{hyperref}
\usepackage{amsmath}
\usepackage{cite}
\usepackage{bm}
\usepackage{color}
\usepackage{setspace}
\usepackage{algorithm}
\usepackage{algorithmic}

\usepackage{pdfpages}
\usepackage{caption}
\usepackage{graphicx}

\usepackage{multirow}
\usepackage{booktabs}
\usepackage{amsmath}
\usepackage{subcaption}
\usepackage{amssymb}

\usepackage{hyperref}
\hypersetup{
    colorlinks=true,       
    linkcolor=red,         
    citecolor=green,       
    urlcolor=blue          
}
\usepackage{xr} 
\begin{document}
\begin{sloppypar}
\title{Benchmarking Tabular Foundation Models as Surrogates in Expensive Evolutionary Optimization
\thanks{This work was supported in part by  the National Natural Science Foundation of China (No. 62636001, 62376202).}
}


\author{
        Lu~Han        \and
        Jin~Wang        \and
        Yuchen~Li         \and
        Haoran~Gu \and
        Shulei~Liu\and
        Ziyang~Shi\and
        Wenao~Lu\and
        Handing~Wang
}
\authorrunning{Short form of author list} 

\institute{ Lu~Han        \and
        Jin~Wang        \and
        Yuchen~Li         \and
        Haoran~Gu \and
        Shulei~Liu\and
        Ziyang~Shi\and
        Wenao~Lu\and
        Handing~Wang \at
              School of Artificial Intelligence, Xidian University, Xi'an, Shaanxi, 710071, China \\
              \email{hdwang@xidian.edu.cn}            \\
              (Lu Han and Jin Wang contributed equally to this work.) 
             (Corresponding author: Handing Wang)
}


\date{Received: date / Accepted: date}

\maketitle

\begin{abstract}
Surrogate-assisted evolutionary algorithms (SAEAs) are effective approaches for solving expensive optimization problems (EOPs), where surrogate models are used to reduce the number of expensive evaluations and can substantially influence optimization performance. In recent years, tabular foundation models have attracted increasing attention, and the Tabular Prior-data Fitted Network (TabPFN) has been explored as a surrogate model for EOPs, showing promising predictive and optimization performance. Motivated by its potential application in SAEAs, this work presents an empirical investigation of TabPFNv2 as a surrogate model in evolutionary optimization. Specifically, we evaluate its performance in both offline and online SAEA settings across various problem scenarios, including single-objective, multi-objective, constrained, combinatorial, mixed-variable, and engineering optimization problems. We also compare TabPFNv2 with conventional surrogate models to examine its predictive and optimization performance under different settings. The experimental results indicate that the effectiveness of TabPFNv2 is problem dependent, and it does not consistently outperform conventional surrogate models across all considered scenarios. Based on these observations, we discuss its applicability and limitations in different optimization settings and highlight several practical considerations for incorporating it into SAEAs.

\keywords{expensive optimization problems, surrogate-assisted evolutionary algorithms, surrogate models, Tabular Prior-data Fitted Network.}

\end{abstract}

\section{Introduction}

Expensive optimization problems (EOPs)~\cite{li2022evolutionary,zhan2020expected} widely exist in real-world optimization applications, for example, aerodynamic shape design of automobiles or aircraft~\cite{yu2018influence}, circuit parameter tuning of electronic systems~\cite{javaheripi2019peeking}, and antenna optimization on communication devices~\cite{PSAMP,xu2026self}. Many practical tasks require real physical experiments or computationally expensive simulations to evaluate candidate solutions, but time, computing power and physical resources are always limited. This makes EOPs very challenging. To address the difficulties of EOPs, surrogate-assisted evolutionary algorithms (SAEAs)~\cite{he2023review,gu2024surrogate} have been widely applied to reduce the required number of expensive evaluations.

Evolutionary algorithms (EAs)~\cite{vikhar2016evolutionary} possess gradient-free black-box optimization capability and population-based global search ability, which makes them well suited to EOPs. However, these advantages require EAs to perform a large number of evaluations, and the resulting high computational cost makes them difficult to apply in practice. To address this, SAEAs~\cite{pan2018classification} incorporate cheap surrogate models to replace or assist some expensive evaluations, thereby obtaining high-quality solutions within a limited evaluation budget.
Researchers have explored various surrogate models, including Gaussian processes (GP)~\cite{10478742,songKrigingAssistedTwoArchiveEvolutionary2021}, radial basis function network (RBFN)~\cite{li2024inverse,huang2021offline,lseo-s3}, support vector regression (SVR)~\cite{shi2020multi}, tree models~\cite{dasari2019random}, and neural networks (NN)~\cite{tahkola2020surrogate}.
They are mostly traditional small-scale machine learning models. Their simple structures result in limited representational capacity, which may make it more challenging for them to fit high-dimensional, strongly nonlinear, and complex fitness landscapes~\cite{liu2024surrogate}.

Recently, Transformer-based foundation models have advanced rapidly. Pretrained on large-scale data, they can capture complex patterns and long-range dependencies, showing strong generalization and adaptability across diverse tasks~\cite{zhou2025comprehensive}. The Tabular Prior-data Fitted Network (TabPFN)~\cite{tabpfnv1} has attracted considerable attention for its strong predictive performance on tabular data. It is pretrained on millions of synthetic tasks and then directly makes predictions on new tasks via in-context learning. On real-world tabular benchmarks, TabPFN achieves competitive normalized root-mean-square error (RMSE) without task-specific tuning and outperforms gradient-boosted decision tree baselines~\cite{friedman2001greedy,ke2017lightgbm}. Following the success of TabPFN, several new tabular foundation models have been developed in recent years, further advancing the capabilities of in-context learning for tabular data. These developments include TabPFNv2~\cite{tabpfnv2}, TabPFN-2.5~\cite{tabpfn25}, as well as TabICL~\cite{tabicl} and TabICLv2~\cite{tabiclv2}, which introduce different architectural designs to improve scalability and efficiency. Among these tabular foundation models, TabPFNv2 is one of the most widely used and extensively studied, and it is therefore the main focus of this study.

Compared with traditional surrogate models, TabPFNv2 leverages a pretrained architecture and prior knowledge acquired from synthetic tasks, which may provide greater capacity for modeling complex real-world fitness landscapes. Its pretrained prior knowledge may further improve predictive performance on previously unseen black-box EOPs, making it a promising surrogate candidate for real-world optimization. Several studies have incorporated TabPFN into Bayesian optimization (BO)~\cite{yu2026git,liu2025exploiting}, where it has demonstrated promising performance. TabPFN has also been used as a surrogate within an SAEA~\cite{du2026meta}, and large language models have been explored as training-free surrogates for evolutionary optimization~\cite{hao2024large}. However, BO and SAEAs differ substantially in both their optimization frameworks and the way surrogate models are integrated. BO typically follows a relatively structured surrogate-driven optimization pipeline, whereas SAEAs allow more flexible and diverse use of surrogates throughout the evolutionary process. Although TabPFN has been used in individual SAEA studies, systematic evidence across diverse SAEA settings remains limited. To address this gap, we conduct an empirical study of TabPFNv2 across a range of expensive optimization tasks in SAEAs and compare its performance with that of traditional surrogate models.

The main contributions of this paper are summarized as follows:
\begin{itemize}
    \item We conduct an empirical investigation of the fundamental prediction capabilities of TabPFNv2 across diverse fitness landscapes. By evaluating its performance under different conditions, including varying sample sizes and dimensionalities, we examine its predictive characteristics and identify its strengths and weaknesses compared with conventional surrogate models. This provides insights into the predictive behavior of TabPFN before its application to evolutionary optimization.
    
    \item By integrating TabPFNv2 into several representative SAEAs, we conduct extensive empirical studies across different expensive optimization scenarios. The results provide a comparative assessment of its optimization performance and reveal both scenarios in which TabPFN achieves competitive results and cases where its effectiveness is limited.
    
    \item Based on the empirical findings, we discuss the applicability and limitations of TabPFNv2 in SAEAs and summarize several practical considerations for its use. These observations may provide useful insights for researchers investigating the application of tabular foundation models to surrogate-assisted evolutionary optimization.
\end{itemize}

The remainder of this paper is organized as follows. Following this introduction, we present the preliminaries in Section \ref{S2}. The prediction performance of TabPFN is systematically evaluated in Section \ref{S3}. Section \ref{S4} conducts comprehensive empirical studies by integrating TabPFN into various SAEAs. Based on these evaluations, Section \ref{S5} summarizes the promises, analyzes the limitations, and provides practical guidelines for the effective use of TabPFN. Finally, conclusions and future work are discussed in Section \ref{S6}.

\section{Preliminaries}\label{S2} 

\subsection{Tabular Prior-data Fitted Network}

TabPFN is a Transformer-based foundation model for tabular data that performs in-context learning without any parameter fine-tuning \cite{tabpfnv1,tabpfnv2}. Based on the Prior-Data Fitted Network (PFN) framework, it is pretrained entirely on diverse synthetic datasets to approximate Bayesian inference \cite{PFN}. Thanks to this offline pretraining, given a support training set $\mathcal{D}_{train}=\{(\mathbf{x}_i,y_i)\}_{i=1}^{n}$ and query features $\mathbf{x}_{test}$, TabPFN directly conditions on $\mathcal{D}_{train}$ to predict the query targets via a single forward pass. Tabular inputs are first encoded into a high-dimensional embedding space, allowing the Transformer to capture complex data dependencies through self-attention mechanisms over the sample sequence. 

\begin{figure}[h] 
    \centering
    \includegraphics[width=\columnwidth]{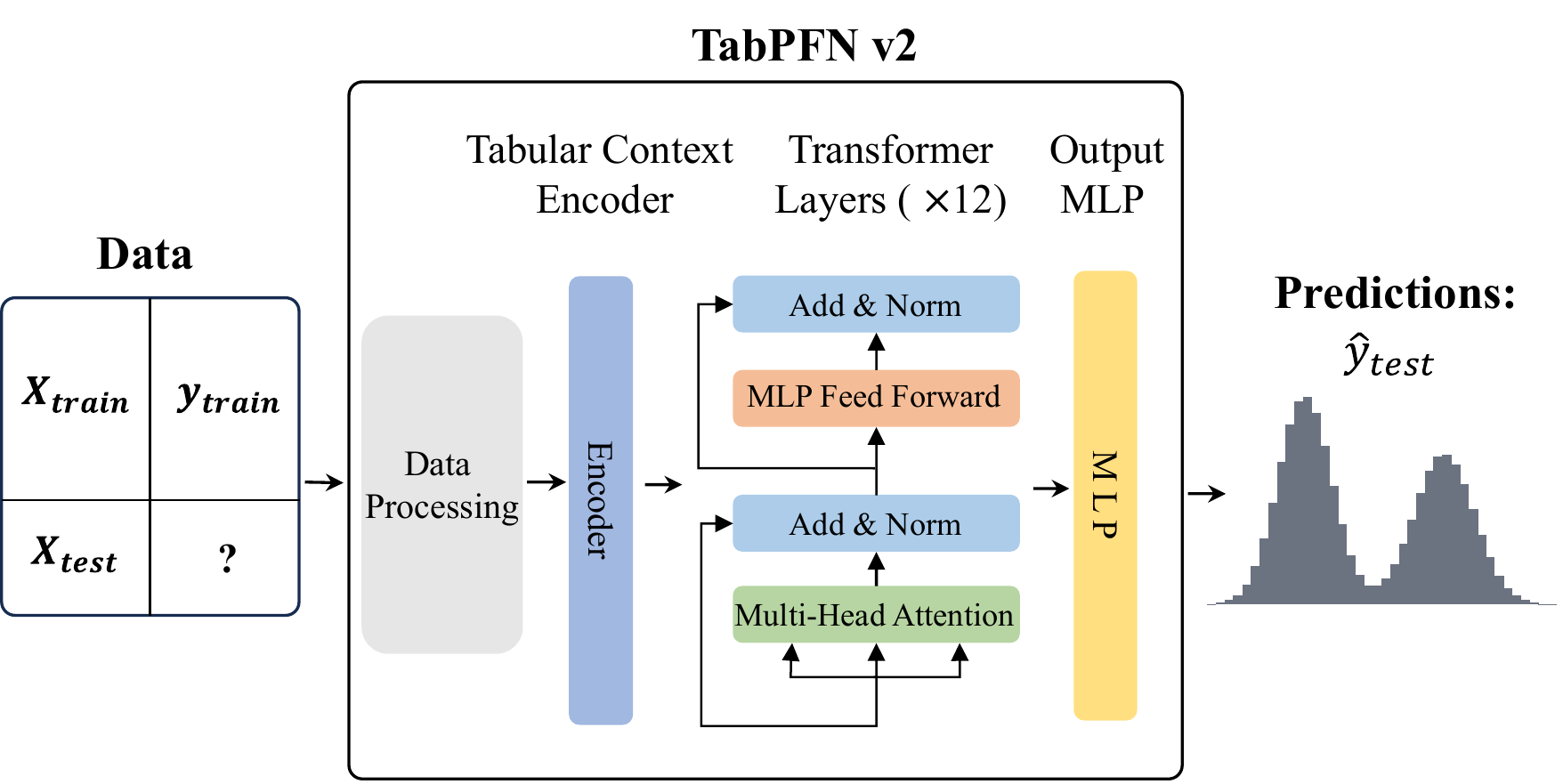} 
    \caption{Overview of the TabPFNv2 architecture and prediction workflow.}
    \label{fig:tabpfn}
\end{figure}

\begin{figure*}[!htbp] 
    \centering
    \includegraphics[width=\linewidth]{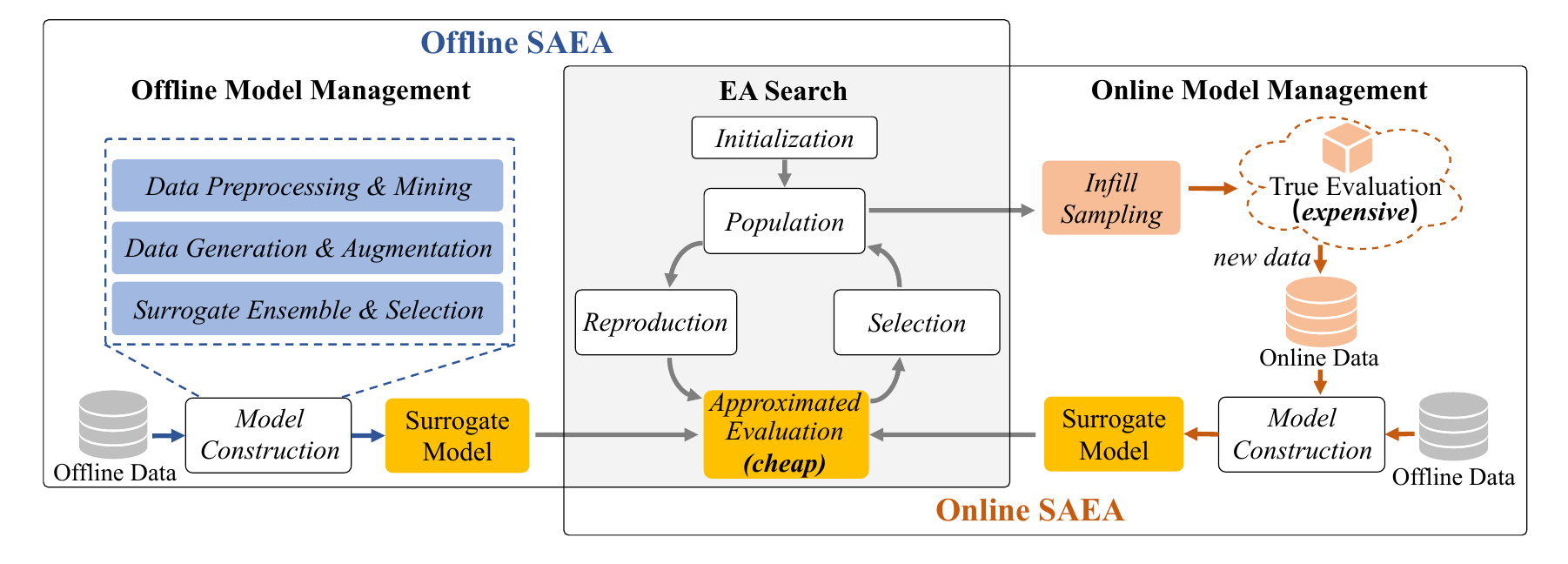} 
    \caption{General framework of offline and online SAEAs}
    \label{fig:SAEA}
\end{figure*}
 
TabPFNv2 extends this framework to a broader range of tasks, scaling up to 10,000 training samples and 500 features~\cite{tabpfnv2}. Furthermore, it inherently accommodates diverse tabular data challenges, such as categorical features, mixed data types, and missing values, without requiring complex manual preprocessing. Notably, for regression tasks, rather than directly outputting a point estimate, TabPFNv2 predicts a non-parametric bar distribution, as illustrated in Fig.~\ref{fig:tabpfn}. Specifically, the target space is partitioned into $K$ predefined bins, denoted by $\mathcal{B}=\{B_1,\ldots,B_K\}$. The model employs a classification-like head to output a probability vector $\mathbf{p}=(p_1,\ldots,p_K)$, where $p_k$ represents the probability mass assigned to the $k$-th bin. Accordingly,

\begin{equation}
P(y\in B_k \mid \mathbf{x}_{test},\mathcal{D}_{train})=p_k.
\end{equation}

The predictive mean is obtained as the probability-weighted average of the conditional means of all bins:

\begin{equation}
\hat{y}_{test}=\sum_{k=1}^{K}p_k\mu_k,
\end{equation}
where $\mu_k=\mathbb{E}[Y\mid Y\in B_k]$ denotes the conditional mean within the $k$-th bin.

To quantify the predictive uncertainty associated with this distribution, the predictive variance is computed as:

\begin{equation}
\sigma^2=
\sum_{k=1}^{K}p_k
\left[
v_k+(\mu_k-\hat{y}_{test})^2
\right],
\end{equation}
where $v_k=\operatorname{Var}(Y\mid Y\in B_k)$ denotes the conditional variance within the $k$-th bin. This formulation accounts for both the uncertainty within individual bins and the variability across different bins. By providing the full predictive distribution and its derived statistics, TabPFNv2 offers not only point predictions but also a measure of predictive uncertainty and other distributional information.

\subsection{Surrogate-Assisted Evolutionary Algorithms}

SAEAs integrate surrogate models in EAs to accelerate computationally expensive optimization tasks~\cite{SAEA-case,liang2025survey}. Fig.~\ref{fig:SAEA} illustrates the overall framework of SAEAs. Both paradigms rely on an underlying EA, such as genetic algorithm (GA) \cite{schmitt2001theory}, differential evolution (DE) \cite{das2010differential}, or particle swarm optimization (PSO) \cite{MGO-SLPSO}, as the base search engine. This engine drives the optimization through a continuous loop of population selection, reproduction, and evaluation. In this loop, SAEAs utilize surrogate models to replace the expensive true evaluations. To seamlessly incorporate the surrogate model into this evaluation phase and the broader evolutionary search, SAEAs require refined model management strategies to fully and effectively utilize the available data, which are typically closely tailored to the characteristics of the underlying surrogate \cite{DSI,DSAEA}. Fundamentally, depending on whether new data can be acquired via true evaluations during the search process, SAEAs are structurally categorized into the distinct offline and online paradigms shown in Fig.~\ref{fig:SAEA} \cite{wang2016data}.

\textbf{Offline SAEAs} are employed when true evaluations are entirely inaccessible during the evolutionary search, relying exclusively on static, pre-collected offline data. Its primary technical bottleneck is the vulnerability to the “false optima" problem, where the EA converges toward regions the surrogate misleadingly predicts as promising due to out-of-distribution (OOD) errors. Since the offline data cannot be dynamically updated, the \textit{Offline Model Management} module acts as a critical safeguard. To mitigate false optima, existing offline model management methods within the \textit{Surrogate Construction} phase can generally be classified into three main categories. The first two approaches tackle the challenge from the perspective of the training data: 1) \textit{Data Preprocessing \& Mining} is utilized to filter noises and extract high-value patterns, and 2) \textit{Data Generation \& Augmentation} is applied to artificially synthesize samples to compensate for sparse regions \cite{wang2016data,li2020boosting, li2020data}. The third approach focuses on the modeling strategy itself, specifically \textit{Surrogate Ensemble \& Selection}. Although grouped together, these are distinct mechanisms: surrogate ensembling integrates multiple models to enhance overall prediction robustness and implicitly estimate uncertainty, whereas surrogate selection adaptively chooses the most suited model for the current optimization stage \cite{wang2018offline, huang2021offline}. Together, these methods maximize the generalization capability of the surrogate within the strict confines of offline data.

\textbf{Online SAEAs} allow for limited interactions with the true evaluation to dynamically refine the surrogate. Because the offline paradigm is essentially a special case of online SAEAs where the evaluation budget is strictly zero, the \textit{Surrogate Construction} designed for offline scenarios are inherently applicable within the online frameworks. The fundamental distinction of the \textit{Online Model Management} module lies in its active data acquisition mechanism: the \textit{Infill Sampling} strategy. After the surrogate acts as a preliminary filter for the offspring population, this sampling strategy is applied to select a small and critical subset of candidates for true evaluation. It evaluates the comprehensive utility of candidates by explicitly balancing two competing dimensions: 1) \textit{Exploitation (Convergence)}, which evaluates the predicted fitness to drive the search toward currently known promising regions; and 2) \textit{Exploration (Uncertainty \& Diversity)}, which targets underexplored areas or regions with high prediction variance to enhance the model's global reliability \cite{MGO-SLPSO,songKrigingAssistedTwoArchiveEvolutionary2021}. Once evaluated, these exact solutions are appended to the online data repository. This new data subsequently triggers the dynamic updating or retraining of the surrogate model. By continuously correcting the surrogate's landscape through this closed-loop feedback mechanism, online SAEAs effectively navigate the complex search space while maximizing the utility of a strict computational budget.

\section{Prediction Performance of TabPFN} \label{S3} 
We select the Ellipsoid and Rastrigin functions and the CEC 2017 benchmark suite to evaluate the predictive performance of TabPFN and two classic surrogate models, RBFN and GP. The latter two serve as comparison models for TabPFN.
The results on the Ellipsoid and Rastrigin functions are visualized using 2D landscapes.
RMSE and Kendall's $\tau$ are used to evaluate prediction error and rank correlation, respectively.
We also test the training and inference time costs of the three models. In this study, the TabPFN used is TabPFNv2.

In the subsequent empirical studies, we evaluate algorithm performance using statistical significance tests. Specifically, we use the Wilcoxon rank-sum test \cite{ranksum} at a significance level of $\alpha = 0.05$. In the reported results, ``$+$'', ``$-$'', and ``$=$'' indicate that the control algorithm performs significantly better than, significantly worse than, or shows no significant difference compared to baseline algorithms, respectively.
In the experiments below in this section, the three models TabPFN, RBFN, and GP use consistent parameter settings. For detailed parameters, please refer to Section S-I-A in the Supplementary Material.

\subsection{2D Visualization}

We select two single-objective functions, Ellipsoid and Rastrigin, with the input dimension $D=2$. We test the predictive performance of TabPFN and two classic surrogate models, GP and RBFN, under different training data sizes and visualize the results as 2D landscapes. The number of training samples is denoted by $N_{\mathrm{tr}}$, which is set to $N_{\mathrm{tr}}=20$ and $N_{\mathrm{tr}}=300$. The test points are sampled on a $200 \times 200$ grid, and the total number of test samples is denoted by $N_{\mathrm{te}}=40000$. We compute RMSE and $\tau$ as evaluation metrics.

\begin{figure}[!ht]
\centering
\subfloat[$N_{\mathrm{tr}}=20$]{\includegraphics[width=\linewidth]{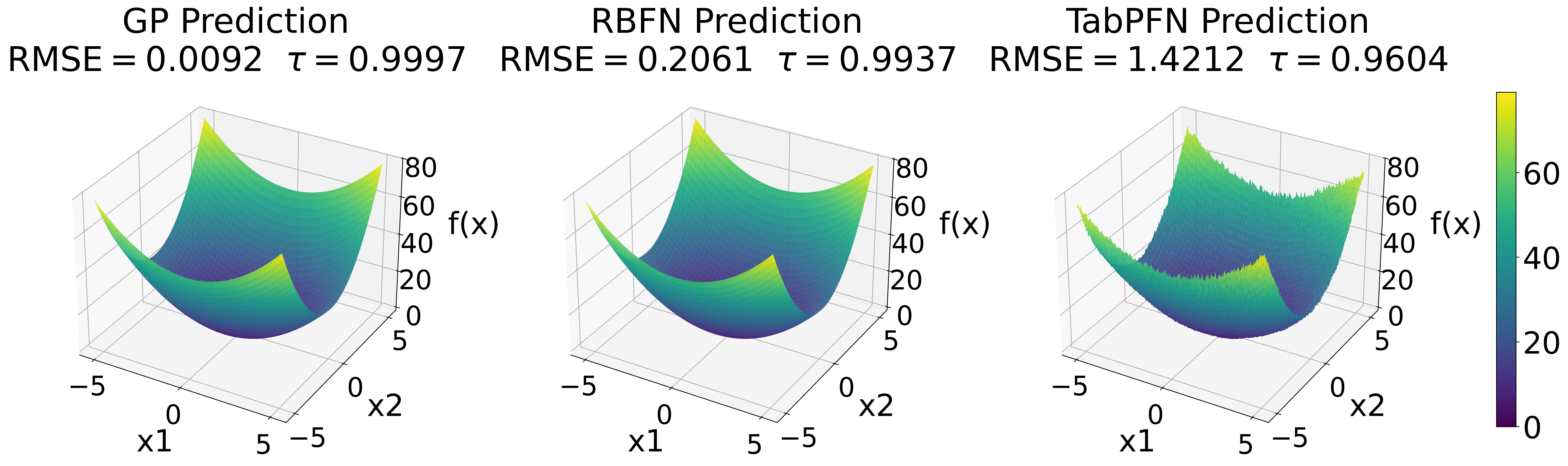}
\label{elltlsm1}}
\vspace{10pt}
\subfloat[$N_{\mathrm{tr}}=300$]{\includegraphics[width=\linewidth]{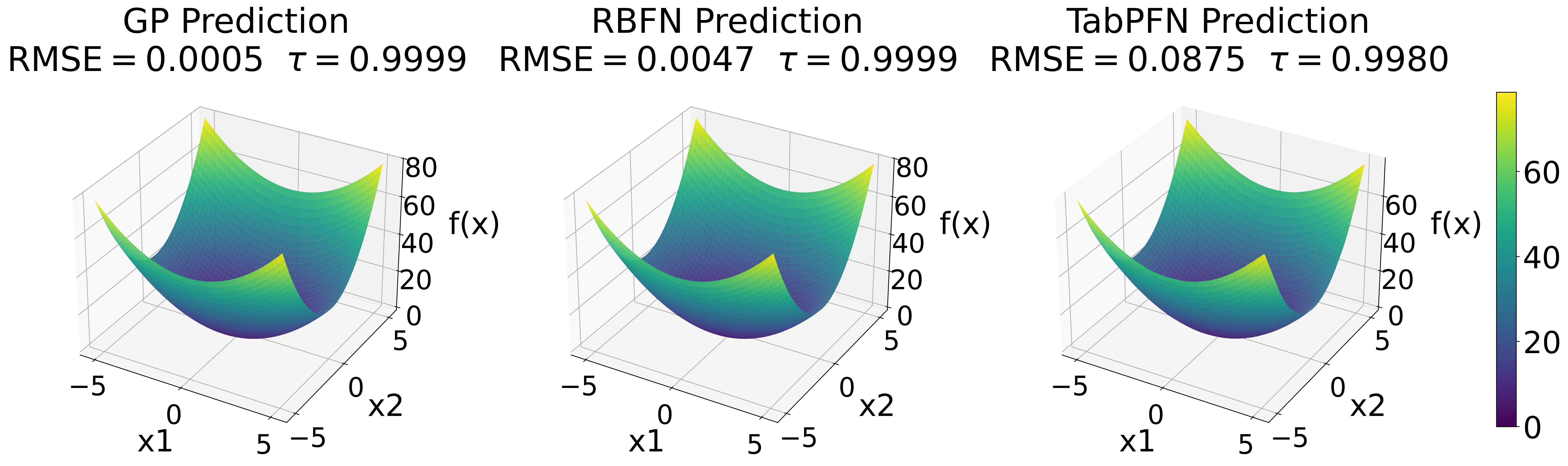}
\label{elltlsm2}}
\caption{The 2D landscape predicted by the surrogate models on the Ellipsoid function.}
\label{elltlsm}
\end{figure}

The landscapes of the three surrogate models on the Ellipsoid function are shown in Fig.~\ref{elltlsm}. When $N_{\mathrm{tr}}=20$, all three models fit the Ellipsoid function well, and their overall trends closely match the true landscape. Among them, GP achieves the optimal RMSE and $\tau$ values.
When $N_{\mathrm{tr}}=300$, the landscapes of all three models are highly accurate, and all evaluation metrics improve substantially compared with the case of $N_{\mathrm{tr}}=20$.
In this scenario, GP also achieves the best performance.

\begin{figure}[!ht]
\centering
\subfloat[$N_{\mathrm{tr}}=20$]{\includegraphics[width=\linewidth]{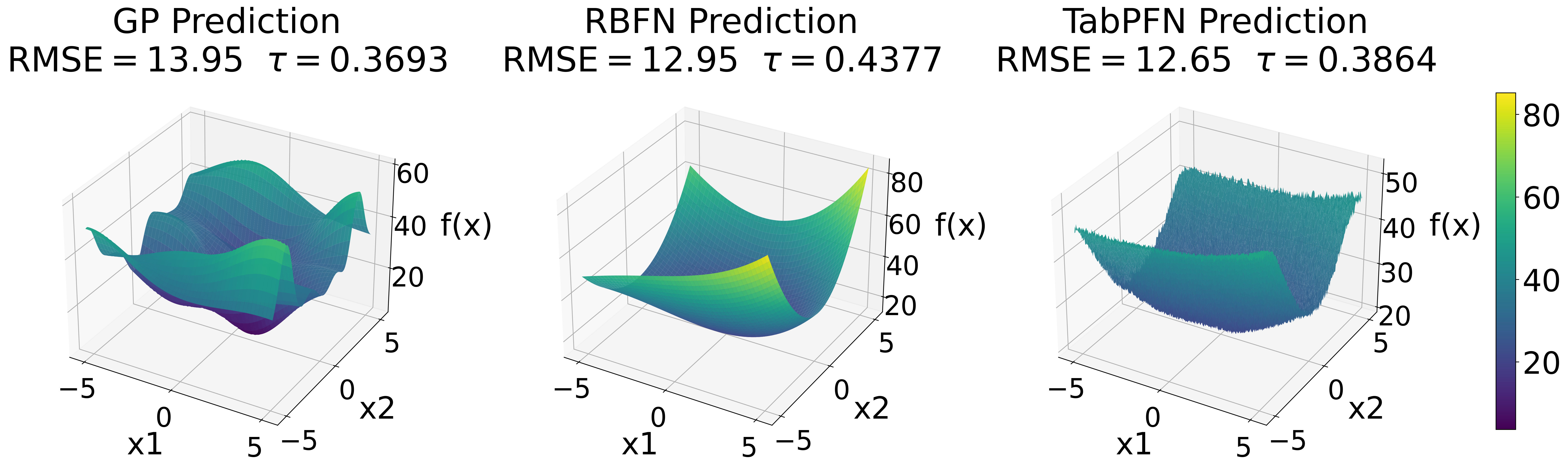}
\label{rastlsm1}}
\vspace{10pt}
\subfloat[$N_{\mathrm{tr}}=300$]{\includegraphics[width=\linewidth]{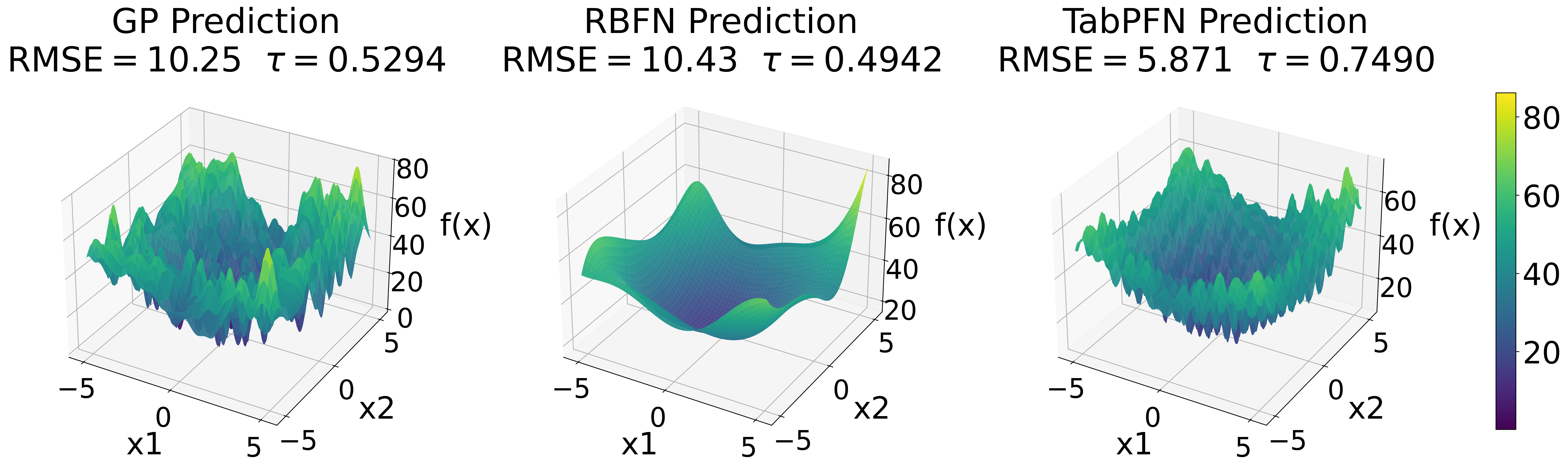}
\label{rastlsm2}}
\caption{The 2D landscape predicted by the surrogate models on the Rastrigin function.}
\label{rastlsm}
\end{figure}

The landscapes of the three surrogate models on the Rastrigin function are shown in Fig.~\ref{rastlsm}.
At $N_{\mathrm{tr}}=20$, all three models perform poorly and can only capture the general trend, with TabPFN being relatively better.
At $N_{\mathrm{tr}}=300$, TabPFN fits the real landscape best and obtains optimal RMSE and $\tau$, capturing both global trends and local details.
RBFN and GP also improve, but their gains are much smaller than that of TabPFN.

Overall, on the simpler unimodal Ellipsoid function, all three models predict very accurately. On the more difficult multimodal Rastrigin function, as the number of training samples increases, TabPFN clearly outperforms GP and RBFN. This shows that TabPFN has strong learning ability and achieves better predictive performance than GP and RBFN on more challenging function tasks.

\subsection{Regression Accuracy and Correlation}

We evaluate the prediction performance of three models, namely TabPFN, GP, and RBFN, on 29 of the 30 functions in the CEC 2017 benchmark~\cite{awad2017cec} suite (excluding F2) using RMSE and $\tau$ as evaluation metrics. For each function, each model runs 25 independent times, and the average result is taken as the final metric. We set three different input dimensions, i.e., $D=10, 30, 100$, and use the Wilcoxon rank-sum test to compare the models under each metric at each dimension. Across all test functions of different dimensions, we use a training sample size of $N_{\mathrm{tr}} = 10D$ and a test sample size of $N_{\mathrm{te}} = 1000$ for the three models.

\begin{table}[!ht]
  \centering
  \caption{Wilcoxon rank-sum significance comparison of GP and RBFN against TabPFN on the CEC 2017 benchmark suite. Each dimension aggregates all 29 evaluated functions, excluding F2.}
  \resizebox{\linewidth}{!}{%
  \begin{tabular}{clcc}
    \toprule
    \multirow{2}{*}{\textbf{Metric}} &
    \multirow{2}{*}{\(\boldsymbol{D}\)} &
    \multicolumn{2}{c}{\textbf{Significance counts \((+/-/=)\)}} \\
    \cmidrule(lr){3-4}
     & & \textbf{GP vs. TabPFN} & \textbf{RBFN vs. TabPFN} \\
    \midrule
    \multirow{3}{*}{RMSE}
      & 10  & 12/11/6 & 3/25/1 \\
      & 30  & 14/14/1 & 6/23/0 \\
      & 100 & 14/12/3 & 1/28/0 \\
    \midrule
    Total & -- & 40/37/10 & 10/76/1 \\
    \midrule
    \multirow{3}{*}{\(\tau\)}
      & 10  & 8/20/1  & 5/23/1 \\
      & 30  & 12/17/0 & 8/20/1 \\
      & 100 & 13/15/1 & 1/27/1 \\
    \midrule
    Total & -- & 33/52/2 & 14/70/3 \\
    \bottomrule
  \end{tabular}%
  }
  \label{wilcoxoncec2017}
\end{table}

The results of the Wilcoxon rank-sum test are summarized in Table~\ref{wilcoxoncec2017}, while the full detailed results for each input dimension are provided in Tables S-I, S-II, S-III in the Supplementary Material.
As shown in the table, across the three dimensionalities of the CEC 2017 benchmark suite, TabPFN has a significant advantage over RBFN in both RMSE and $\tau$. Compared with GP, however, GP achieves more significant wins in RMSE, whereas TabPFN achieves more significant wins in $\tau$. Overall, TabPFN and GP perform comparably. GP gives lower errors, while TabPFN has better ranking performance. TabPFN is significantly better than RBFN in prediction performance.

\subsection{Training and Inference Time}

Using the CEC 2017 F1 as the test function, we evaluate the training and inference time costs of GP, RBFN, and TabPFN under different dimensionalities and sample sizes. We also test their time costs on different GPU hardware, where the GTX 1650 is an entry-level GPU, the RTX 4090 is a flagship GPU, and the RTX A6000 is a professional GPU. For each setting, we conduct 25 independent runs for all three models and use the average training and inference time per sample as the final time performance. Note that TabPFN requires no training from scratch. It only performs data fit equivalent to the training of other models. We thus consistently use “training” to describe “fit”.

\begin{figure}[!ht]
    \centering
    \includegraphics[width=\columnwidth]{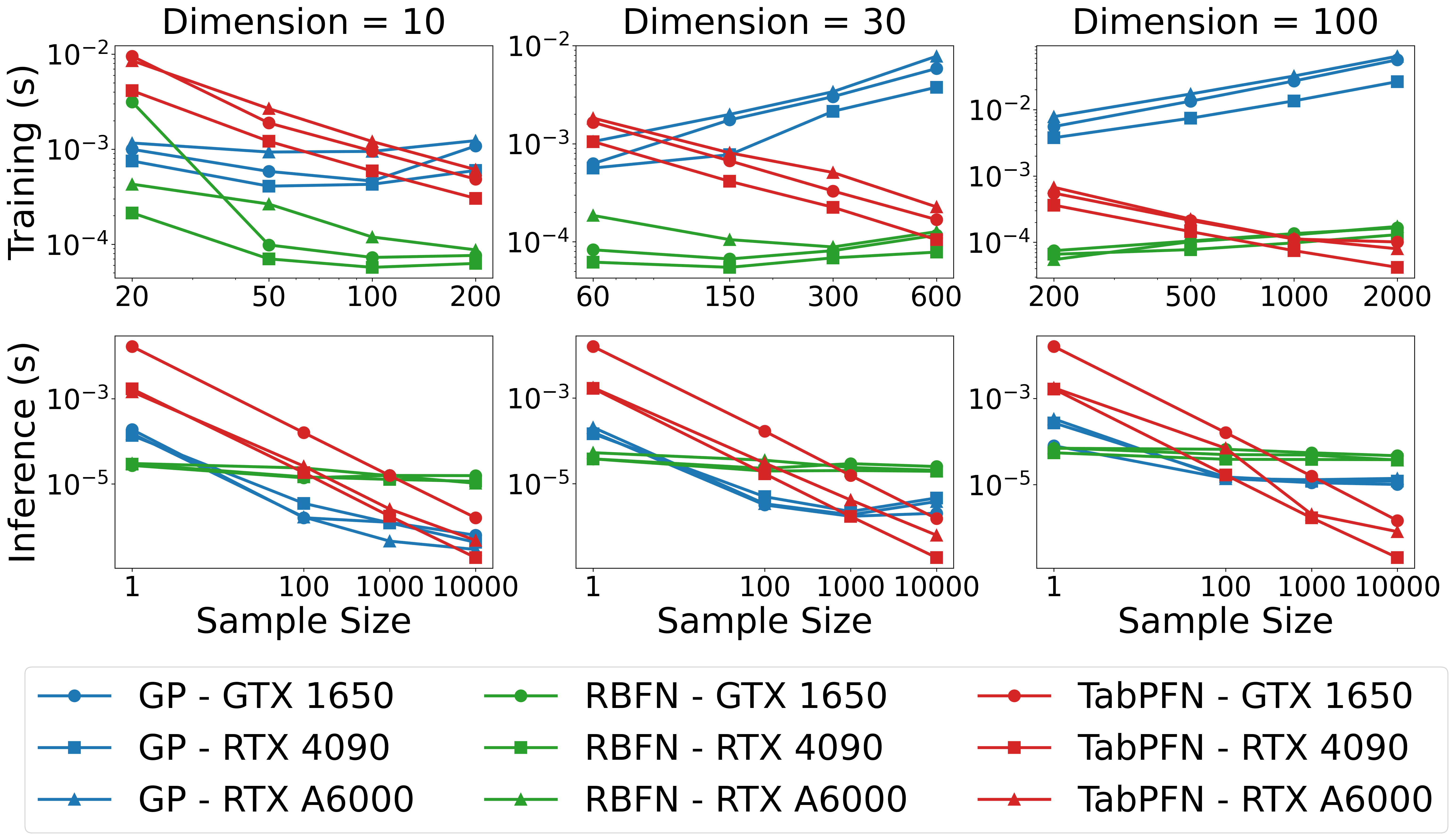}
    \caption{Training and Inference Time Curves for GP, RBFN, and TabPFN on the CEC 2017 F1.}
    \label{fig:time}
\end{figure}

We summarize and compare the training and inference time of different models in Fig.~\ref{fig:time}. Regarding training time, GP generally becomes slower as dimensionality and sample size increase, particularly in high-dimensional settings. RBFN maintains relatively low training costs across different settings, while TabPFN exhibits a noticeable decrease in training time as the sample size increases. Although TabPFN can be relatively costly on small datasets, it becomes increasingly competitive in large-sample and high-dimensional settings, sometimes even outperforming RBFN. For the inference time, RBFN has roughly similar time costs across different dimensions and sample sizes. In contrast, the per-sample inference time of both GP and TabPFN decreases substantially as the sample size increases, with TabPFN exhibiting a particularly pronounced reduction. Consequently, TabPFN achieves the lowest per-sample inference time in most large-sample settings. Regarding inference time, GP and RBFN show almost no dependence on the GPU hardware. TabPFN, however, runs significantly faster on the RTX 4090 and RTX A6000 than on the GTX 1650. In summary, TabPFN is particularly advantageous for high-dimensional and large-sample scenarios, offering highly efficient inference and increasingly competitive training time as the sample size grows, although its training cost can be relatively high on small datasets.

\subsection{Variance Visualization}

\begin{figure}[!ht]
    \centering
    \includegraphics[width=\columnwidth]{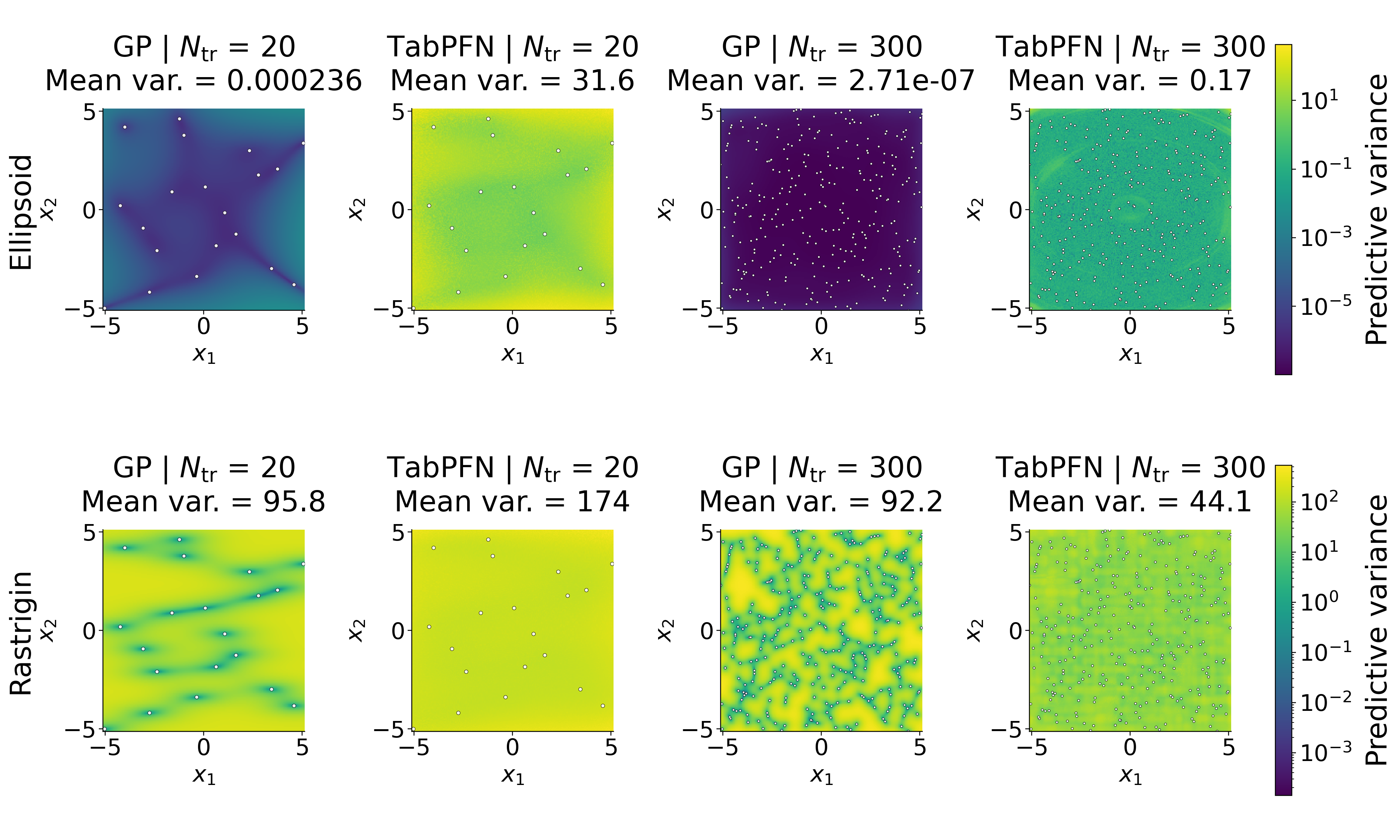}
    \caption{Predictive variance distributions of GP and TabPFN on the Ellipsoid and Rastrigin functions with different training sample sizes.}
    \label{fig:variance}
\end{figure}

We visualize the predictive variance of GP and TabPFN using two test functions, Ellipsoid and Rastrigin. For each function, models are fitted using 20 and 300 training samples, respectively, with the same training samples shared by both models. Predictions are then evaluated on a regular \(200\times200\) grid. The predictive variances are visualized as two-dimensional heatmaps on the original target scale, with training sample locations overlaid to illustrate their spatial distributions and the effects of sample size.

As shown in Fig.~\ref{fig:variance}, the two models exhibit distinct spatial patterns of predictive variance. For the Ellipsoid function, GP shows relatively smooth variance distributions, with a substantial reduction in overall variance as the sample size increases. TabPFN exhibits comparatively higher variance with 20 samples, but its variance also decreases considerably with 300 samples and becomes more spatially homogeneous. For the Rastrigin function, GP displays more complex and spatially heterogeneous variance patterns, characterized by localized low-variance regions around training samples. These patterns become more detailed as the sample size increases, while the mean variance remains relatively stable. In contrast, TabPFN exhibits a more broadly distributed variance pattern, with a noticeable reduction in mean variance as the sample size increases. Overall, the spatial distribution and magnitude of predictive variance depend on both the model, the underlying function, and the number of training samples.

\section{Empirical Studies} \label{S4}

To comprehensively benchmark the performance of TabPFN in SAEAs, this section evaluates TabPFN as a direct replacement for traditional machine learning surrogates within established SAEA frameworks. This straightforward substitution corresponds to the highlighted “Surrogate Model” modules illustrated in Fig.~\ref{fig:SAEA}, allowing us to objectively assess its empirical optimization capabilities. Our experimental study systematically spans both offline and online SAEA paradigms. Furthermore, to rigorously test its versatility and robustness, the online SAEA covers a broad and challenging spectrum of problem scenarios. These encompass single-objective, multi-objective, constrained, and combinatorial optimization, alongside mixed-variable spaces, large-scale decision variables, and one engineering case.

\subsection{Single-Objective Optimization}
\subsubsection{Offline Optimization Problems}

In the offline optimization scenario, only an initial dataset is available, and no additional true evaluations are allowed during the search.

We consider TT-DDEA~\cite{huang2021offline}, a representative offline framework using three RBFN surrogates trained on different data subsets, denoted as TT-DDEA-RBFN. Its model management strategy updates each surrogate with pseudo-labeled individuals selected according to the prediction consistency of the other surrogates. We replace RBFN with TabPFN while keeping the same model management strategy, resulting in TT-DDEA-TabPFN. To isolate the effect of the model management strategy, we also introduce DDEA-TabPFN, which removes the model management mechanism and trains a single TabPFN on the entire offline dataset.
The CEC 2017 benchmark suite with 10, 30, and 100 decision variables is used for evaluation~\cite{awad2017cec}, which includes simple functions (F1, F3--F10), hybrid functions (F11--F20), and composition functions (F21--F30). Each algorithm is run 25 independent times on each function.
Table~\ref{tab:offline_win} summarizes the results of the Wilcoxon rank-sum test at a significance level of 0.05, while the detailed results are provided in Table S-IV of the Supplementary Material.

\begin{table}[!ht]
  \centering
  \caption{Significance comparison of TT-DDEA-TabPFN and DDEA-TabPFN against TT-DDEA-RBFN on the CEC 2017 benchmark suite.}
  \label{tab:offline_win}
  \renewcommand{\arraystretch}{1.1}
  \setlength{\tabcolsep}{5pt}

  \resizebox{\columnwidth}{!}{%
  \begin{tabular}{@{}lccc@{}}
    \toprule
    \multirow{2}{*}{\textbf{Functions}}
    & \multirow{2}{*}{$\boldsymbol{D}$}
    & \multicolumn{2}{c}{\textbf{Significance counts} ($+/-/=$)} \\
    \cmidrule(lr){3-4}
    & & \textbf{TT-DDEA-TabPFN} & \textbf{DDEA-TabPFN} \\
    \midrule
    \multirow{3}{*}{\begin{tabular}[c]{@{}l@{}}
      Simple\\(F1, F3--F10)
    \end{tabular}}
    & 10  & 1/7/1 & 1/6/2 \\
    & 30  & 1/7/1 & 1/7/1 \\
    & 100 & 0/8/1 & 1/3/5 \\
    \midrule
    \multirow{3}{*}{\begin{tabular}[c]{@{}l@{}}
      Hybrid\\(F11--F20)
    \end{tabular}}
    & 10  & 8/0/2 & 9/0/1 \\
    & 30  & 6/3/1 & 7/1/2 \\
    & 100 & 2/5/3 & 4/5/1 \\
    \midrule
    \multirow{3}{*}{\begin{tabular}[c]{@{}l@{}}
      Composition\\(F21--F30)
    \end{tabular}}
    & 10  & 4/3/3 & 5/1/4 \\
    & 30  & 2/3/5 & 3/4/3 \\
    & 100 & 5/4/1 & 5/3/2 \\
    \midrule
    \textbf{Total} & -- & 29/40/18 & 36/30/21 \\
    \bottomrule
  \end{tabular}
  }
\end{table}

\begin{figure}[!ht] 
    \centering
    
    
    

    \includegraphics[width=\columnwidth]{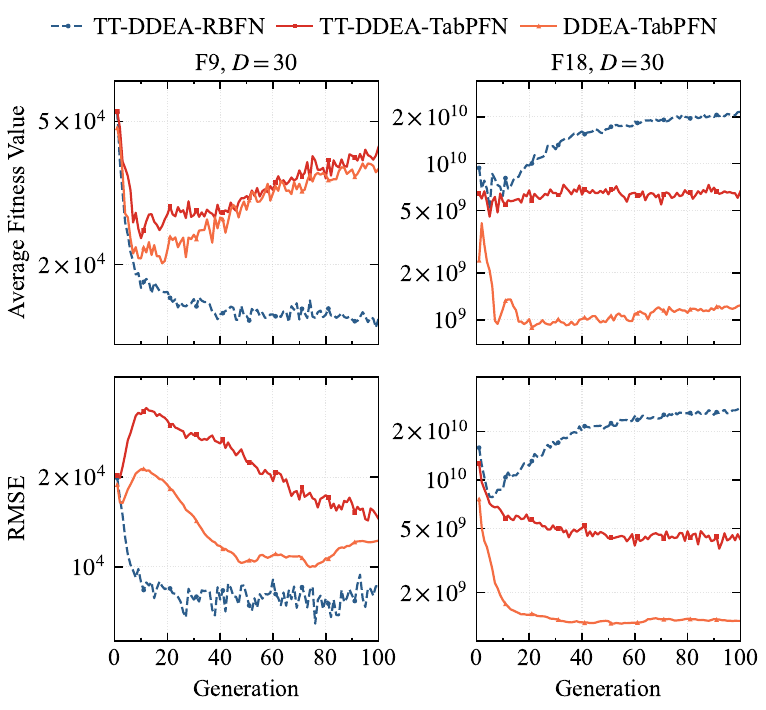}
    
    \caption{Performance on F9 and F18 with $D=30$. (Top) Convergence curves of the compared algorithms. (Bottom) RMSE profiles of the surrogate models over generations.}
    \label{fig:offline_performance}
\end{figure}

From Table~\ref{tab:offline_win}, DDEA-TabPFN achieves overall performance comparable to TT-DDEA-RBFN. The advantage of DDEA-TabPFN is mainly observed on hybrid functions and some composition functions, suggesting that TabPFN is effective in modeling complex fitness landscapes. In contrast, TT-DDEA-RBFN remains competitive on simple functions, where RBFN can already approximate the relatively smooth landscapes effectively.
To further examine these differences, Fig.~\ref{fig:offline_performance} presents the convergence curves and RMSE profiles on F9 and F18 over 25 independent runs. These two functions are selected as representative cases of relatively simple and complex fitness landscapes, respectively.
On F9, TT-DDEA-RBFN shows more stable convergence and the lowest RMSE, which explains its better performance on simple landscapes. On F18, the TabPFN-based variants, especially DDEA-TabPFN, achieve better convergence, with steadily decreasing RMSE values. In contrast, TT-DDEA-RBFN shows weaker convergence and increasing RMSE. These results further confirm that RBFN is effective for simple functions, whereas TabPFN provides more accurate surrogate modeling and more reliable search guidance on complex functions.

Therefore, TabPFN can work well as the main surrogate in offline SAEAs by providing more reliable fitness estimates on complex landscapes. However, directly combining it with the original model management strategy is less effective, probably because pseudo-labeling introduces additional uncertainty and data partitioning limits its use of the offline dataset. Thus, training data should not be unnecessarily split or augmented with pseudo-labels simply to follow conventional model management strategies.

\subsubsection{Small/Medium-Scale Online Optimization Problems}

To evaluate the effectiveness of TabPFN in online SAEAs, we consider MGP-SLPSO, a GP-assisted social learning PSO that employs a multi-objective infill criterion for model management \cite{MGO-SLPSO}. We replace the original GP surrogate with TabPFN while keeping the same optimization and model management strategies, resulting in TabPFN-SLPSO. The CEC 2017 benchmark suite with 10, 30, and 100 decision variables is used for evaluation. The detailed results and statistical tests are provided in Table S-V and S-VI of the Supplementary Material, while Table~\ref{tab:slpso_significance} summarizes the corresponding significance comparison counts.

  \begin{table}[!ht]
  \centering
  \caption{Significance comparison on the CEC 2017 benchmark suite.
    ``GP'' denotes MGP-SLPSO, and ``TabPFN'' denotes TabPFN-SLPSO.}
  \label{tab:slpso_significance}
  \renewcommand{\arraystretch}{1.1}
    \setlength{\tabcolsep}{10pt}

    \resizebox{\columnwidth}{!}{%
  \begin{tabular}{@{}lcc@{}}
    \toprule
    \textbf{Functions} & \boldmath$D$ & \textbf{TabPFN vs GP} ($+/-/=$) \\
    \midrule
    \multirow{3}{*}{\begin{tabular}[c]{@{}l@{}}Simple\\(F1, F3--F10)\end{tabular}}
    & 10  & 4/1/4 \\
    & 30  & 7/1/1 \\
    & 100 & 8/0/1 \\
    \midrule
    \multirow{3}{*}{\begin{tabular}[c]{@{}l@{}}Hybrid\\(F11--F20)\end{tabular}}
    & 10  & 8/0/2 \\
    & 30  & 9/0/1 \\
    & 100 & 10/0/0 \\
    \midrule
    \multirow{3}{*}{\begin{tabular}[c]{@{}l@{}}Composition\\(F21--F30)\end{tabular}}
    & 10  & 7/0/3 \\
    & 30  & 9/0/1 \\
    & 100 & 9/0/1 \\
    \midrule
    Total & -- & 71/2/14 \\
    \bottomrule
  \end{tabular}
  }
\end{table}

As shown in s~\ref{tab:slpso_significance}, TabPFN-SLPSO generally achieves better final objective values than MGP-SLPSO. On simple functions, the counts of significantly better, significantly worse, and no significant difference are 4/1/4, 7/1/1, and 8/0/1 at $D=10, 30$, and $100$, respectively. On hybrid and composition functions, TabPFN-SLPSO obtains significant advantages in most comparisons and no significant loss. Overall, the counts are 71/2/14, indicating a clear advantage under the same surrogate-assisted evolutionary framework.

Statistical results (Table S-V) indicate that this advantage grows with increasing dimensionality. The counts of significantly better, significantly worse, and no significant difference are 19/1/9 at $D=10$, 25/1/3 at $D=30$, and 27/0/2 at $D=100$. Unlike the offline setting, where TabPFN primarily excels on complex landscapes, the online results also show gains on relatively simple functions under this continually updated surrogate setting. Surrogate modeling results (Table~\ref{tab:rmse_tau_win}) further support these observations. TabPFN achieves lower RMSE on 29, 29, and 28 functions for $D=10, 30$, and $100$, providing substantially more accurate fitness estimates. It also obtains higher Kendall's $\tau$ on most functions. MGP-SLPSO's infill strategy jointly considers predicted fitness and uncertainty rather than relying solely on ranking. Consequently, the reduced RMSE serves as the primary driver of optimization gains, making the ranking results complementary evidence of surrogate quality.

\begin{table}[!ht]
  \centering
  \caption{Significance counts of population RMSE and $\tau$ during optimization between MGP-SLPSO and TabPFN-SLPSO on CEC 2017.}
  \setlength{\tabcolsep}{12pt}
  \renewcommand{\arraystretch}{1.1}

    \resizebox{\columnwidth}{!}{%
  \begin{tabular}{ccc}
    \toprule
    \boldmath$D$ & Metric & \textbf{TabPFN vs MGP} ($+/-/=$) \\
    \midrule
    \multirow{2}{*}{10} & RMSE & 29/0/0 \\
                        & $\tau$ & 29/0/0 \\
    \midrule
    \multirow{2}{*}{30} & RMSE & 29/0/0 \\
                        & $\tau$ & 25/0/4 \\
    \midrule
    \multirow{2}{*}{100} & RMSE & 28/1/0 \\
                         & $\tau$ & 20/1/8 \\
    \bottomrule
  \end{tabular}
  }
  \label{tab:rmse_tau_win}
\end{table}

Overall, TabPFN is a highly competitive alternative to GP for online SAEAs. Its consistent advantage across various landscapes and dimensionalities, alongside offline results, suggests that TabPFN's effectiveness depends not only on the optimization landscape characteristics, but also on how the surrogate is updated and integrated into the optimization process.

\subsubsection{Large-Scale Online Optimization Problems}
\label{sec_lso}
In large-scale optimization, surrogate models often struggle to provide sufficient predictive performance under limited data due to the massive number of decision variables. To address this, divide-and-conquer SAEAs \cite{saearfs,sadeamss,lseo-s3} commonly employ RBFN for modeling and searching within low-dimensional subspaces. Their promising results demonstrate that RBFN effectively fits fitness landscapes within subspaces formed by partial subsets of decision variables. Here, we investigate whether substituting RBFN with TabPFN can further improve optimization performance.

We compare the state-of-the-art LSEO-S3 \cite{lseo-s3} with LSEO-S3-TabPFN on five representative $1000D$ CEC2013 problems \cite{li2013benchmark} (F1, F4, F8, F12, and F15, being the first from each category). The variant replaces the original RBFN surrogate component with TabPFN while retaining the optimization framework. Detailed experimental settings are provided in Section S-I-D of the Supplementary Material. As shown in Table~\ref{tab:cec2013_comparison}, LSEO-S3-TabPFN performs significantly worse than LSEO-S3 on all five problems.

\begin{table}[!ht]
    \centering
    \caption{Optimization mean value results comparison between LSEO-S3 (RBFN) and LSEO-S3-TabPFN on CEC2013 benchmark functions.}
    \label{tab:cec2013_comparison}
    \resizebox{\columnwidth}{!}{%
    \begin{tabular}{lcc}
        \toprule
        \textbf{Function} & \textbf{LSEO-S3 (RBFN)} & \textbf{LSEO-S3-TabPFN} \\
        \midrule
        F1  & \textbf{7.39e+09 (4.53e+08)}   & 5.90e+10 (6.57e+09) $-$ \\
        F4  & \textbf{2.48e+11 (3.54e+10)}   & 8.23e+11 (3.22e+11) $-$ \\
        F8  & \textbf{9.62e+15 (1.21e+15)}   & 2.56e+16 (2.10e+16) $-$ \\
        F12 & \textbf{1.12e+11 (9.14e+09)}   & 2.39e+12 (2.05e+11) $-$ \\
        F15 & \textbf{6.88e+07 (1.54e+07)}   & 6.46e+13 (1.08e+14) $-$ \\
        \midrule
         $+$/$-$/$=$ & / &  0/5/0 \\
        \bottomrule
    \end{tabular}
    }
\end{table}

To examine whether this performance gap is associated with predictive accuracy, we compare RBFN and TabPFN in 100-dimensional subspaces under Global and Local test scenarios (Table \ref{tab:combined_comparison}). TabPFN achieves lower RMSE in all ten function–scenario combinations and higher Kendall’s \(\tau\) in nine of them. These predictive advantages do not translate into better optimization results, indicating that predictive accuracy alone is insufficient to explain surrogate effectiveness within LSEO-S3.

\begin{table}[!ht]
    \centering
    \caption{Comparison of RMSE and Kendall's $\tau$ mean between RBFN and TabPFN in Global and Local scenarios.}
    \label{tab:combined_comparison}
    \resizebox{\columnwidth}{!}{%
    \begin{tabular}{clcccc}
        \toprule
        \multirow{2}{*}{\textbf{Metric}} & \multirow{2}{*}{\textbf{Func}} & \multicolumn{2}{c}{\textbf{Global}} & \multicolumn{2}{c}{\textbf{Local}} \\
        \cmidrule(lr){3-4} \cmidrule(lr){5-6}
        & & \textbf{RBFN} & \textbf{TabPFN} & \textbf{RBFN} & \textbf{TabPFN} \\
        \midrule
        \multirow{5}{*}{RMSE} 
        & F1  & 2.6447e+11 & \textbf{4.2871e+10} & 1.4494e+12 & \textbf{2.6275e+11} \\
        & F4  & 3.9836e+14 & \textbf{1.7628e+14} & 1.1532e+15 & \textbf{2.6089e+14} \\
        & F8  & 1.4271e+19 & \textbf{1.0211e+19} & 1.6149e+19 & \textbf{1.3760e+19} \\
        & F12 & 8.1615e+12 & \textbf{5.8880e+11} & 4.8661e+13 & \textbf{8.0758e+12} \\
        & F15 & 1.1298e+19 & \textbf{5.9660e+18} & 3.7774e+19 & \textbf{1.3542e+19} \\
        \midrule
        \multirow{5}{*}{$\tau$} 
        & F1  & -0.0145 & \textbf{0.1286} & 0.0536 & \textbf{0.1273} \\
        & F4  & 0.0539  & \textbf{0.1208} & 0.0942 & \textbf{0.0973} \\
        & F8  & 0.0422  & \textbf{0.1074} & 0.0381 & \textbf{0.0556} \\
        & F12 & -0.0619 & \textbf{0.0873} & 0.0022 & \textbf{0.0687} \\
        & F15 & 0.0402  & \textbf{0.1489} & \textbf{0.0540} & -0.0013 \\
        \bottomrule
    \end{tabular}%
    }
\end{table}

\begin{table}[!ht]
    \centering
    \caption{Mean predictions in the F8 extrapolation diagnostic. The scale factor \(r\) specifies a nominal displacement \(r\ell\) from the best training point, where \(\ell\) is the median nearest-neighbor distance among training inputs.}
    \label{tab:dist_scale_preds}
    \resizebox{\columnwidth}{!}{%
    \begin{tabular}{lcccc}
        \toprule
        

         \(\boldsymbol{r}\) & \textbf{real\_dist} & \(\boldsymbol{\hat{y}}_{\textbf{RBFN}}\) & \(\boldsymbol{\hat{y}}_{\textbf{TabPFN}}\) & \(\boldsymbol{y}_{\textbf{train\_mean}}\) \\
        
        \midrule
        0.25  & 88.07  & 2.2757e+17  & 1.2106e+19 & 1.4145e+19 \\
        0.50  & 165.95 & -1.8353e+17 & 1.2454e+19 & 1.4145e+19 \\
        1.00  & 299.79 & -2.7424e+18 & 1.3200e+19 & 1.4145e+19 \\
        2.00  & 441.19 & -1.7237e+19 & 1.3914e+19 & 1.4145e+19 \\
        4.00  & 500.96 & -4.1182e+19 & 1.4378e+19 & 1.4145e+19 \\
        8.00  & 528.30 & -5.3035e+19 & 1.5132e+19 & 1.4145e+19 \\
        16.00 & 540.93 & -6.5718e+19 & 1.5101e+19 & 1.4145e+19 \\
        \bottomrule
    \end{tabular}%
    }
\end{table}

We further examine extrapolation behavior on F8 using runtime states collected during LSEO-S3 optimization. The two surrogates are evaluated at increasing distance scales from the best training point, following the protocol in Section S-I-D of the Supplementary Material. Table \ref{tab:dist_scale_preds} summarizes the mean predictions, the mean training label, and the mean distance to the nearest training sample at each scale. As this distance increases, RBFN produces increasingly extreme and favorable predicted objective values, whereas TabPFN remains close to the mean training label. Under candidate selection based on minimum predicted objective values, such RBFN predictions may favor distant candidates and encourage exploration. This difference provides a plausible explanation for the observed optimization gap, despite TabPFN’s better predictive accuracy in the subspace tests.

\subsection{Constrained Optimization}

To evaluate the performance of TabPFN in expensive constrained optimization, we integrated it into DSI ~\cite{DSI} by replacing RBFN. The comparison is based on the mean performance over 25 independent runs. For problems where an algorithm cannot find feasible solutions in all runs, only the success rate (SR) is reported, where a larger SR indicates a higher probability of locating the feasible region. Table \ref{tab:dsi_tabpfn_cf} details the results across ten benchmark problems selected from the widely used CEC2006 \cite{cec2006} and CEC2010 \cite{cec2010} test suites. The data reveals a systemic degradation when using TabPFN: DSI-RBFN significantly outperforms DSI-TabPFN on 8 out of 10 problems. DSI-TabPFN only achieves a statistical advantage on  g08. The most striking failure occurs on problem g12, where DSI-TabPFN's SR plummets to a mere $36\%$, whereas DSI-RBFN robustly maintains a $100\%$ SR. Furthermore, both algorithms struggle heavily on c15 (yielding an SR of $4\%$), indicating an intrinsically difficult constraint landscape that challenges both surrogates.

\begin{table}[!ht]
\centering
\caption{Comparison results of DSI-RBFN and DSI-TabPFN on ten constrained problems. The best result for each problem is highlighted in bold.}
\label{tab:dsi_tabpfn_cf}
\setlength{\tabcolsep}{10pt}
\resizebox{\columnwidth}{!}{%
\begin{tabular}{@{}ccc@{}}
\toprule
\textbf{Problem} & \textbf{DSI-RBFN} & \textbf{DSI-TabPFN} \\
\midrule
g02 & \textbf{-2.93e-01(6.46e-02)} & -2.28e-01(8.23e-02)${-}$ \\
g06 & \textbf{-6.96e+03(3.48e-03)} & -6.37e+03(2.75e+02)${-}$ \\
g08 & -7.11e-02(3.37e-02) & \textbf{-9.22e-02(8.09e-03)}${+}$ \\
g12 & \textbf{100\%} & 36\%${-}$ \\
g24 & \textbf{-5.51e+00(1.74e-09)} & -5.51e+00(1.82e-03)${-}$ \\
c01 & \textbf{-4.57e-01(1.01e-01)} & -2.51e-01(9.66e-02)${-}$ \\
c07 & \textbf{3.05e+06(4.70e+06)} & 3.13e+08(7.51e+08)${-}$ \\
c08 & \textbf{7.25e+07(9.63e+07)} & 1.68e+10(1.03e+10)${-}$ \\
c14 & \textbf{2.82e+12(5.58e+12)} & 7.46e+12(1.98e+13)${-}$ \\
c15 & 4\% & 4\%${=}$ \\
\midrule
$+$/$-$/$=$ & /&{1/8/1}  \\
\bottomrule
\end{tabular}%
}

\end{table}

\begin{table}[!ht]
\centering
\caption{Function complexity and constraint prediction range compression. ``RBFN'' denotes RBFN range, and ``TabPFN'' denotes TabPFN range.}
\label{tab:complexity_range}
\resizebox{\columnwidth}{!}{%
\begin{tabular}{@{}ccccc@{}}
\toprule
\textbf{Problem} & \textbf{Dim} & \textbf{Quad \(\boldsymbol{R^2}\)} & \textbf{RBFN(\%)} & \textbf{TabPFN(\%)} \\
\midrule
g02 & 20 & 0.067 & 265.37 & 53.36 \\
g06 & 2 & 1.000 & 100.12 & 99.48 \\
g08 & 2 & 1.000 & 99.53 & 99.81 \\
g12 & 3 & 0.286 & 147.78 & 48.06 \\
g24 & 2 & 0.630 & 96.63 & 97.43 \\
c01 & 10 & 0.518 & 119.93 & 63.25 \\
c07 & 10 & 0.037 & 100.60 & \textbf{11.15} \\
c08 & 10 & 0.012 & 91.30 & \textbf{1.63} \\
c14 & 10 & 0.054 & 96.20 & \textbf{12.22} \\
c15 & 10 & 0.012 & 109.73 & \textbf{8.61} \\
\bottomrule
\end{tabular}
}
\end{table}

The underlying cause of this widespread failure is quantitatively exposed in Table~\ref{tab:complexity_range}, which links function complexity with constraint prediction range compression. Here, function complexity is measured by Quad $R^2$ (the $R^2$ value of a quadratic polynomial fit), where a lower value indicates a highly non-linear and complex landscape. On simpler problems (e.g., g06 and g08), TabPFN maintains a normal prediction range and performs competitively. However, when confronting highly complex landscapes absent from its pre-training priors (characterized by low Quad $R^2$), TabPFN often produces predictions with a substantially compressed output range. Unlike the distance-based RBFN, which intrinsically preserves the scale and amplitude of the training data, TabPFN undergoes drastic variance shrinkage—collapsing to a mere 1.63\% range on c08. By severely compressing the variance, TabPFN completely blurs the critical mathematical boundaries between feasible ($g(x) \le 0$) and infeasible regions, rendering the constraint surrogate incapable of effectively guiding the EA.

\subsection{Multi-Objective Optimization}

\subsubsection{Unconstrained Problems}
To investigate the applicability of TabPFN to expensive multi-objective optimization, we select KTA2~\cite{songKrigingAssistedTwoArchiveEvolutionary2021}, a representative GP-assisted evolutionary algorithm, as the baseline. 
KTA2 maintains two archives to promote convergence and diversity and adaptively selects solutions for expensive evaluations according to the current optimization state. 
We replace its GP models with TabPFN while keeping the evolutionary search and model management strategies unchanged, resulting in KTA2-TabPFN. 
The two algorithms are evaluated on the unconstrained DTLZ1--DTLZ4 benchmark problems~\cite{deb2002scalable} with $M=3$ and $M=10$ objectives. 
The IGD$^{+}$ indicator is used to assess the quality of the obtained solution sets, where a smaller value indicates better.
Each algorithm is independently run 25 times on each problem setting, and the results are reported as the mean and standard deviation of IGD$^{+}$. Detailed experimental settings are provided in Section S-I-F of the Supplementary Material.

Table~\ref{tab:kta2_tabpfn} presents the statistical comparison results. 
KTA2-TabPFN obtains lower mean IGD$^{+}$ values in six of the eight settings, with two significant wins, one significant loss, and five ties against KTA2. 
For $M=3$, it significantly reduces the mean IGD$^{+}$ on DTLZ1 and DTLZ4 by approximately $20.5\%$ and $65.7\%$, respectively, and also achieves a lower mean on DTLZ3, while KTA2 performs slightly better on DTLZ2. 
For $M=10$, KTA2-TabPFN obtains slightly lower mean values on DTLZ1--DTLZ3 but is significantly worse on DTLZ4, with an increase of approximately $14.5\%$. 
Overall, replacing GP with TabPFN largely preserves the performance of KTA2 and provides clear improvements on some three-objective problems, although this advantage diminishes as the number of objectives increases. 
This may be because KTA2's model management strategy was originally designed around GP predictions and uncertainty estimates, suggesting that further adaptation may be needed for many-objective problems.

\begin{table}[!ht]
    \centering
    \caption{IGD$^{+}$ results of KTA2 and KTA2-TabPFN on the
    DTLZ1--DTLZ4 benchmark problems. The results are reported as
    mean (standard deviation), with the better mean shown in bold.}
    \label{tab:kta2_tabpfn}
    \renewcommand{\arraystretch}{1.1}
    \setlength{\tabcolsep}{5pt}

    \resizebox{\columnwidth}{!}{%
    \begin{tabular}{lccc}
        \toprule
        \textbf{Problem}
        & \(\boldsymbol{M}\)
        & \textbf{KTA2 (GP)}
        & \textbf{KTA2-TabPFN} \\
        \midrule
        \multirow{2}{*}{DTLZ1}
        & 3  & 6.00e+01 (1.61e+01)
             & \textbf{4.77e+01 (7.40e+00)} $+$ \\
        & 10 & 4.20e-01 (2.06e-01)
             & \textbf{3.74e-01 (2.89e-01)} $=$ \\
        \multirow{2}{*}{DTLZ2}
        & 3  & \textbf{3.98e-02 (3.05e-03)}
             & 4.19e-02 (4.16e-03) $=$ \\
        & 10 & 4.39e-01 (1.49e-02)
             & \textbf{4.37e-01 (2.04e-02)} $=$ \\
        \multirow{2}{*}{DTLZ3}
        & 3  & 1.54e+02 (4.05e+01)
             & \textbf{1.38e+02 (3.29e+01)} $=$  \\
        & 10 & 1.25e+00 (4.29e-01)
             & \textbf{1.21e+00 (4.22e-01)} $=$  \\
        \multirow{2}{*}{DTLZ4}
        & 3  & 1.71e-01 (1.10e-01)
             & \textbf{5.86e-02 (1.21e-02)} $+$  \\
        & 10 & \textbf{3.38e-01 (3.94e-02)}
             & 3.87e-01 (4.69e-02) $-$  \\
        \midrule
        \multicolumn{2}{l}{$+$/$-$/$=$} & / & 2/1/5 \\
        \bottomrule
    \end{tabular}
    }
\end{table}

\subsubsection{Constrained Problems}

To investigate the applicability of TabPFN to expensive constrained multi-objective optimization, we select KTS~\cite{songBalancingObjectiveOptimization2024}, a GP-assisted evolutionary algorithm with two search modes, as the baseline. 
KTS adaptively switches between unconstrained and constrained surrogate-assisted search according to the relationship between objective optimization and constraint satisfaction. We replace both the objective and constraint GP models with TabPFN while keeping the remaining search and model management strategies unchanged, resulting in KTS-TabPFN. 
To further identify the source of the performance difference, we construct two additional variants: KTS-TabPFN(obj), where only the objective surrogates are replaced by TabPFN, and KTS-TabPFN(cons), where only the constraint surrogates are replaced.

\begin{table}[!ht]
    \centering
    \caption{Comparison results of KTS and its three variants on ten CF
    test problems. The best result for each problem is highlighted in bold.}
    \label{tab:kts_tabpfn_cf}

    \renewcommand{\arraystretch}{1.1}
    \setlength{\tabcolsep}{1pt}

    \resizebox{\columnwidth}{!}{%
    \begin{tabular}{@{}ccccc@{}}
        \toprule
        \textbf{Problem}
        & \textbf{KTS}
        & \textbf{KTS-TabPFN}
        & \textbf{KTS-TabPFN(obj)}
        & \textbf{KTS-TabPFN(cons)} \\
        \midrule

        CF1
        & 1.04e-01(2.49e-02)
        & 1.94e-01(3.78e-02)$^{-}$
        & \textbf{9.76e-02(2.52e-02)}$^{=}$
        & 1.99e-01(4.03e-02)$^{-}$ \\

        CF2
        & 6.08e-02(2.03e-02)
        & 1.59e-01(4.84e-02)$^{-}$
        & 1.78e-01(7.97e-02)$^{-}$
        & \textbf{5.43e-02(1.82e-02)}$^{=}$ \\

        CF3
        & 1.32e+00(2.15e-01)
        & 1.91e+00(9.20e-01)$^{-}$
        & 1.82e+00(7.95e-01)$^{-}$
        & \textbf{1.22e+00(1.58e-01)}$^{+}$ \\

        CF4
        & 1.58e-01(4.43e-02)
        & 3.20e-01(1.23e-01)$^{-}$
        & 3.36e-01(8.68e-02)$^{-}$
        & \textbf{1.57e-01(5.07e-02)}$^{=}$ \\

        CF5
        & \textbf{7.84e-01(4.87e-01)}
        & 1.82e+00(9.00e-01)$^{-}$
        & 1.35e+00(5.17e-01)$^{-}$
        & 8.54e-01(6.35e-01)$^{=}$ \\

        CF6
        & \textbf{6.07e-02(2.37e-02)}
        & 2.32e-01(8.49e-02)$^{-}$
        & 2.15e-01(8.46e-02)$^{-}$
        & 7.71e-02(3.69e-02)$^{=}$ \\

        CF7
        & 9.81e-01(6.93e-01)
        & 3.76e+00(1.69e+00)$^{-}$
        & 3.36e+00(1.03e+00)$^{-}$
        & \textbf{9.70e-01(7.06e-01)}$^{=}$ \\

        CF8
        & \textbf{3.96e-01(1.28e-01)}
        & 3.98e-01(1.17e-01)$^{=}$
        & 4.49e-01(1.30e-01)$^{=}$
        & 4.11e-01(1.37e-01)$^{=}$ \\

        CF9
        & \textbf{1.36e-01(6.26e-02)}
        & 1.90e-01(6.80e-02)$^{-}$
        & 1.84e-01(8.91e-02)$^{-}$
        & 1.65e-01(8.64e-02)$^{=}$ \\

        CF10
        & \textbf{72.00\%}
        & 56.00\%$^{=}$
        & 64.00\%$^{=}$
        & 60.00\%$^{=}$ \\

        \midrule
        $+/-/=$
        & /
        & 0/8/2
        & 0/7/3
        & 1/1/8 \\
        \bottomrule
    \end{tabular}%
    }
\end{table}

The four algorithms are evaluated on the ten CF benchmark problems \cite{zhang2008multiobjective} under the same experimental settings.
The comparison is based on statistical significance tests over 25 independent runs.
For problems where an algorithm cannot find feasible solutions in all runs, only the SR is reported, where a larger SR indicates a higher probability of locating the feasible region. Detailed experimental settings are provided in Section S-I-G of the Supplementary Material. The comparison results are summarized in Table~\ref{tab:kts_tabpfn_cf}.

As shown in Table~\ref{tab:kts_tabpfn_cf}, KTS-TabPFN obtains a $+/-/=$ result of 0/8/2 against KTS, indicating that directly replacing all GP models with TabPFN generally degrades the performance. 
A similar result is observed for KTS-TabPFN(obj) with 0/7/3, whereas KTS-TabPFN(cons) achieves 1/1/8 and remains comparable to KTS on most problems. 
This suggests that the degradation mainly results from replacing the objective surrogates, which directly guide the search toward promising regions and are therefore more sensitive to changes in prediction behavior. 
In contrast, constraint surrogates mainly identify feasible regions, where TabPFN can provide useful information without substantially affecting the original search mechanism. These results indicate that TabPFN is more suitable for constraint modeling in KTS, while its use for objective modeling may require further adaptation.

\subsection{Combinatorial Optimization}
The decision variables of combinatorial optimization problems are discrete rather than continuous, such as binary, integer, or permutation variables. 
Due to the discrete nature, constructing an accurate surrogate model is often a major challenge when applying SAEAs to expensive combinatorial optimization problems \cite{liu2024surrogate}. To evaluate the performance of TabPFN in this setting, we use feature selection as a representative test problem. Feature selection aims to identify a subset of relevant features while maintaining or improving learning performance \cite{xue2015survey}. It can be formulated as a binary combinatorial optimization problem, where 1 and 0 indicate feature selection and exclusion, respectively.

We select SAEAPRG \cite{liu2022surrogate}, a representative offline SAEA for high-dimensional feature selection problems, as the baseline. Its key idea is to employ parallel random grouping to decompose the high-dimensional decision space into multiple lower-dimensional subspaces, thereby facilitating surrogate modeling and evolutionary search. We replace the original RBFN models in SAEAPRG with TabPFN while keeping all other components unchanged, resulting in SAEAPRG-TabPFN. The two algorithms are evaluated on 12 classification datasets, with the number of features ranging from 500 to 10,000. The training set size ($N_{tr}$) follows the original setting in SAEAPRG. Each algorithm is independently run 25 times on each dataset. A $k$-nearest neighbors (KNN) classifier is used to evaluate the selected feature subsets, and the classification error is adopted as the optimization objective.

Table \ref{tab:combi_fs} reports the classification errors obtained by SAEAPRG and SAEAPRG-TabPFN on the 12 feature selection datasets. Overall, replacing the original RBFN with TabPFN does not lead to any significant performance degradation. SAEAPRG-TabPFN achieves lower mean classification errors on 8 out of the 12 datasets and significantly outperforms SAEAPRG on ORL and Gutenberg, while no statistically significant difference is observed on the remaining 10 datasets. In particular, competitive performance is maintained across datasets with dimensionalities ranging from 500 to 10,000, and no clear deterioration of TabPFN is observed as the problem dimensionality increases. These results suggest that the binary and discrete nature of feature selection does not, by itself, prevent TabPFN from serving as an effective surrogate in combinatorial optimization. Such a training-data distribution may partially explain why the extrapolation limitation of TabPFN is less pronounced in feature selection.

\begin{table}[!ht]
    \centering
    \caption{Mean (standard deviation) of classification errors obtained by SAEAPRG and SAEAPRG-TabPFN. The best results are highlighted.}
    \label{tab:combi_fs}
    \resizebox{\columnwidth}{!}{%
    \begin{tabular}{lccc}
        \toprule
        \textbf{Datasets} & \(\boldsymbol{D}\) & \textbf{SAEAPRG} & \textbf{SAEAPRG-TabPFN} \\
        \midrule
        Madelon  & 500 & 1.59e-01(1.05e-02) & \textbf{1.53e-01(6.80e-03)}$=$\\
        Isolet  & 617 & 1.39e-01(1.28e-02) & \textbf{1.18e-01(1.38e-02)}$=$\\
        CNAE-9&856 & \textbf{1.06e-01(9.63e-03)} & 1.08e-01(9.53e-03)$=$\\
        ORL&1024 & 1.72e-01(2.12e-02) & \textbf{1.46e-01(2.09e-02)}$+$\\
        COIL20&1024& 8.69e-03(2.13e-03) & \textbf{7.33e-03(2.43e-03)}$=$\\
        Warp&2420 & 6.15e-02(1.72e-02) & \textbf{5.83e-02(1.44e-02)}$=$\\
        Lung&3312& \textbf{4.39e-02(2.73e-02)} & 4.87e-02(2.34e-02)$=$\\ 
        lymphoma&4026& \textbf{1.13e-01(3.88e-02)} & 1.21e-01(2.77e-02)$=$\\
        RELATHE&4322& 1.74e-01(8.59e-03) & \textbf{1.53e-01(1.45e-02)}$=$\\
        DLBCL&5469& \textbf{3.75e-02(2.26e-02)} & 3.81e-02(1.99e-02)$=$\\
        Gutenberg&8265& 3.17e-01(2.84e-02) & \textbf{2.72e-01(9.54e-03)}$+$\\
        arcene&10000& 1.14e-01(1.97e-02) & \textbf{1.01e-01(1.62e-02)}$=$\\
        \midrule
        $+$/$-$/$=$ & & / & 2/0/10 \\
        \bottomrule
    \end{tabular}
    }
\end{table}

The results shown in Table \ref{tab:combi_fs} contrast with those on the large-scale optimization problems in Section \ref{sec_lso}, where TabPFN performs worse than RBFN. This suggests that the effectiveness of TabPFN may depend less on whether the variables are continuous or discrete, and more on the degree of extrapolation required during optimization. Unlike the space-filling LHS used in large-scale optimization, SAEAPRG samples training solutions in a structured and non-uniform manner according to the number of selected features. Such a training-data distribution may partially explain why the extrapolation limitation of TabPFN is less pronounced in feature selection.

\subsection{Mixed-Variable Optimization}

Mixed-variable optimization problems involve both continuous and discrete variables, where different values of discrete variables partition the search space into distinct regions, resulting in discontinuous objective landscapes. To evaluate the applicability of TabPFN as a surrogate model for mixed-variable optimization, this study adopts SHEALED as the baseline algorithm~\cite{liu2023surrogate}, which employs RBFNmv with the Gower distance to measure sample similarity and combines global and local search to balance exploration and exploitation. We replace RBFNmv with TabPFN while keeping the evolutionary search and model management strategies unchanged, resulting in a variant termed SHEALED-TabPFN. The experiments consider the Type 1 and Type 2 mixed-variable benchmarks introduced in~\cite{liu2023surrogate}. These two benchmark classes differ in how the candidate values of the discrete variables are generated. For each class, five test problems are evaluated in 10, 30, and 50 dimensions.

\begin{table}[!ht]
	\centering
	\caption{Statistical comparison of the final best objective values obtained by SHEALED-TabPFN and SHEALED on the mixed-variable benchmark problems, $n_1$ and $n_2$ denote the numbers of continuous and discrete variables, respectively.}
	\label{tab:mixed_summary}
    \resizebox{\columnwidth}{!}{%

	\begin{tabular}{@{}ccc@{}}
		\toprule
		\textbf{Problem Type}
		& \(\boldsymbol{(n_1,n_2)}\)
		& \textbf{SHEALED-TabPFN vs SHEALED} \\
		\midrule

		\multirow{3}{*}{\shortstack{Simple\\(Type 1)}}
		& $(5,5)$   & 0/3/2 \\
		& $(15,15)$ & 1/4/0 \\
		& $(25,25)$ & 2/2/1 \\
		\midrule

		\multirow{3}{*}{\shortstack{Hard\\(Type 2)}}
		& $(5,5)$   & 0/5/0 \\
		& $(15,15)$ & 0/5/0 \\
		& $(25,25)$ & 0/2/3 \\
		\midrule

		Total & / & 3/21/6 \\
		\bottomrule
	\end{tabular}
    }
\end{table}

Table~\ref{tab:mixed_summary} summarizes the statistical comparison of the final best objective values obtained by SHEALED and SHEALED-TabPFN on the Type 1 and Type 2 mixed-variable benchmark problems, while the detailed optimization results are provided in Table S-VII of the Supplementary Material. The results indicate that the original SHEALED achieves better optimization performance on most test instances. To further investigate the performance discrepancy mentioned above, we compare the predictive performance of RBFNmv and TabPFN on the same mixed-variable benchmark instances. Table~\ref{tab:mixed_surrogate_statistical_comparison} summarizes their statistical comparison in terms of RMSE, $R^2$, and Kendall's $\tau$, while the detailed predictive results are provided in Table S-VIII of the Supplementary Material. The results show that RBFNmv achieves better predictive accuracy and candidate-ranking capability on most test instances.

\begin{table}[!ht]
	\centering
	\caption{Statistical comparison between TabPFN and RBFNmv on the mixed-variable benchmark problems, with $N_{\mathrm{tr}}=10D$ and $N_{\mathrm{te}}=2000$.}
	\label{tab:mixed_surrogate_statistical_comparison}
    \setlength{\tabcolsep}{7pt}
\resizebox{0.6\columnwidth}{!}{%
	\begin{tabular}{@{}lccc@{}}
		\toprule
		\multirow{2}{*}{\textbf{Metric}}
		& \multicolumn{3}{c}{\textbf{TabPFN vs RBFNmv}} \\
		\cmidrule(lr){2-4}
		& $\boldsymbol{(5,5)}$
		& $\boldsymbol{(15,15)}$
		& $\boldsymbol{(25,25)}$ \\
		\midrule

		RMSE   & 3/5/2 & 2/7/1 & 1/8/1 \\
		$R^2$  & 3/5/2 & 2/7/1 & 1/8/1 \\
		$\tau$ & 2/5/3 & 2/8/0 & 0/8/2 \\

		\bottomrule
	\end{tabular}
    }
\end{table}

\begin{figure}[!ht]
    \centering
    \includegraphics[width=\columnwidth]
    {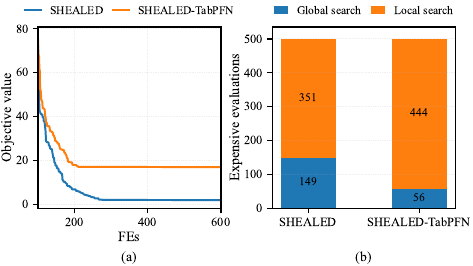}
    
    \caption{Search performance of SHEALED and SHEALED-TabPFN on the 10-dimensional Type 1 Rastrigin problem: (a) convergence curves and (b) expensive evaluation allocation between global and local search.}
    \label{fig:mixed_search_behavior}
\end{figure}

To investigate the impact of the surrogate model on the optimization performance, we compare the convergence processes of the two algorithms, as illustrated in Fig.~\ref{fig:mixed_search_behavior}. Fig.~\ref{fig:mixed_search_behavior}(a) presents the median best-so-far convergence curves, showing that SHEALED converges faster and achieves a better final solution than SHEALED-TabPFN. Fig.~\ref{fig:mixed_search_behavior}(b) further shows that SHEALED-TabPFN allocates fewer expensive evaluations to global search and more to local search than SHEALED. This indicates that replacing RBFNmv with TabPFN gradually shifts the search toward local exploitation. The weaker prediction and candidate-ranking capability of TabPFN makes globally generated candidates less competitive after expensive evaluation, while locally sampled solutions are more likely to remain highly ranked. As this effect accumulates, the parent population used by global search becomes increasingly concentrated around the current promising region, thereby further weakening global exploration and causing an imbalance between exploration and exploitation. Overall, the direct application of TabPFN as a surrogate model yields limited performance in mixed-variable optimization. Therefore, applying TabPFN to such problems necessitates the adoption of specialized model adaptation strategies.

\subsection{Case Studies:Embodied Robot Design}

\begin{figure}[!ht]
     \centering
     \includegraphics[width=0.95\columnwidth]{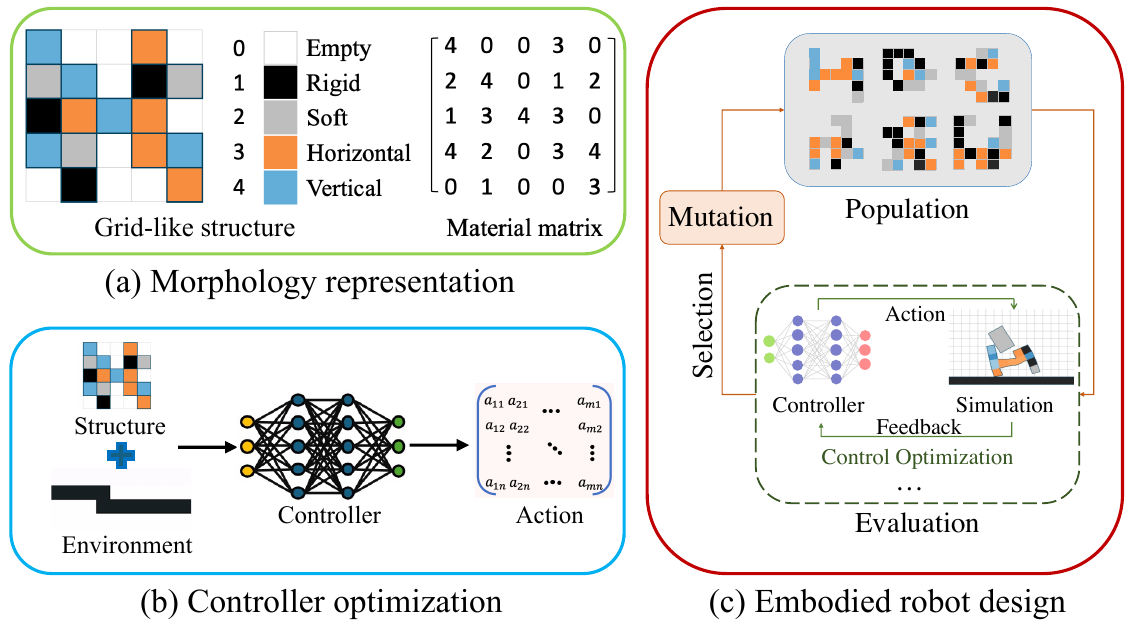}
     \caption{Overview of voxel-based soft robot co-design in EvoGym. (a) Morphology representation, (b) controller optimization, and (c) embodied robot co-design process.}
     \label{fig:robot}
\end{figure}

Voxel-based soft robot design jointly optimizes robot morphology and control policy. Evolutionary reinforcement learning is widely used for this purpose, with EAs searching over voxel-based morphologies and reinforcement learning optimizing the corresponding controllers. Since each candidate morphology requires controller training before its fitness can be evaluated, this problem is typically formulated as an expensive bilevel optimization problem. In this study, we consider voxel-based soft robots in EvoGym~\cite{bhatia2021evolution}, where each morphology is represented by a voxel grid, as shown in Fig.~\ref{fig:robot}(a). PPO is employed for controller optimization, as illustrated in Fig.~\ref{fig:robot}(b), and the overall co-design process is shown in Fig.~\ref{fig:robot}(c).

To evaluate TabPFN on soft robot design, we adopt AIEA \cite{liu2023rapidly}, an action inheritance-based SAEA, as the baseline and replace its performance approximation mechanism with TabPFN, yielding TabPFNEA. Since each morphology is represented as a \(5\times5\) voxel grid, it is flattened into a 25-dimensional vector before being used for TabPFN training and inference. We evaluate the algorithms under two settings:
\begin{enumerate}
\item Online setting: The algorithm first randomly samples and evaluates \(N_{\mathrm{pop}}\) morphologies to initialize the surrogate model. During evolution, \(N_{\mathrm{pop}}/2\) candidate morphologies are selected in each generation for true evaluation, and the resulting samples are added to the training set for surrogate updating. This process continues until the evaluation budget is exhausted.
\item Offline setting: The entire evaluation budget is used to construct the training set before optimization, and no additional true evaluations are performed during the subsequent search.
\end{enumerate}
The population size \(N_{\mathrm{pop}}\) is set to 20. The total evaluation budgets are set to 100, 200, and 300 for easy, medium, and hard tasks, respectively. Considering the computational cost of controller optimization and simulation, each algorithm is independently run five times on each task.


\begin{table}[!ht]
  \centering
  \caption{Significance comparison of AIEA and TabPFNEA on 30 EvoGym tasks across three difficulty levels under online and offline settings.}
  \label{tab:aiea_summary}
  \setlength{\tabcolsep}{15pt}
  \resizebox{\columnwidth}{!}{%
  \begin{tabular}{@{}lcc@{}}
    \toprule
    \textbf{Difficulty} & \textbf{Setting}
    & \textbf{TabPFNEA vs AIEA} ($+/-/=$) \\
    \midrule
    \multirow{2}{*}{Easy}
    & Online  & 1/2/7  \\
    & Offline & 4/2/4  \\
    \midrule
    \multirow{2}{*}{Medium}
    & Online  & 1/2/8  \\
    & Offline & 4/2/5  \\
    \midrule
    \multirow{2}{*}{Hard}
    & Online  & 0/6/3  \\
    & Offline & 5/1/3  \\
    \midrule
    \multirow{2}{*}{Overall}
    & Online  & 2/10/18 \\
    & Offline & 13/5/12 \\
    \bottomrule
  \end{tabular}
  }
\end{table}

Table \ref{tab:aiea_summary} summarizes the significance counts of AIEA and TabPFNEA on 30 EvoGym tasks, with detailed results reported in Table S-IX of the Supplementary Material. In the online setting, AIEA significantly outperforms TabPFNEA on 10 tasks, while TabPFNEA achieves significantly better performance on only 2 tasks, with no significant differences observed on the remaining 18 tasks. In contrast, under the offline setting, TabPFNEA achieves 13 significant wins compared with 5 for AIEA, while the remaining 12 tasks show no significant differences. For easy and medium tasks, the two methods exhibit no significant differences on most tasks, whereas more pronounced differences emerge on hard tasks. Specifically, AIEA achieves 6 significant wins against none for TabPFNEA in the online setting, while TabPFNEA achieves 5 significant wins against only 1 for AIEA in the offline setting.

The contrasting results under the online and offline settings suggest that the effectiveness of TabPFN as a surrogate model may depend on the availability of training data. In the online setting, TabPFN is initially trained on only 20 evaluated morphologies, whereas the offline setting provides 100, 200, and 300 training samples for easy, medium, and hard tasks, respectively. Unlike AIEA, which relies on morphologically similar evaluated individuals for action inheritance, TabPFN learns the relationship between morphology and task performance from the available training data. Consequently, larger training datasets may enable TabPFN to construct more effective surrogate models, particularly for challenging tasks. This interpretation is consistent with the substantial performance advantage of TabPFNEA on hard tasks in the offline setting. Overall, these results suggest that TabPFN is more effective when a relatively sufficient offline dataset is available, whereas AIEA is more competitive in online optimization scenarios with limited initial training data.

\section{Promises, Limitations, and Practical Guidelines of TabPFN} \label{S5}

We assess TabPFN as a surrogate model in both offline and online SAEAs, covering single-objective, multi-objective, constrained, combinatorial, and mixed-variable optimization problems, as well as practical engineering cases. Overall, although TabPFN shows certain advantages, it is not a universal replacement for conventional surrogate models. Whether its advantages translate into improved optimization performance depends on the data size, the complexity of the fitness landscape, the type of search space, the role of the surrogate model within the algorithm, and the degree of alignment between the model management strategy and TabPFN's predictive behavior. In the following, we summarize scenarios where TabPFN performs well, scenarios where it has limitations, and guidelines for its efficient use.

\subsection{Promising Application Scenarios of TabPFN}

\textbf{Offline Optimization with Complex Fitness Landscapes.}
TabPFN is more suitable for offline SAEAs with complex, multimodal, or nonlinear fitness landscapes. Results on hybrid and some composition functions in the CEC 2017 benchmark show that TabPFN can provide more accurate fitness approximation and more reliable search guidance.
For simple and smooth landscapes, conventional surrogates such as RBFN are often sufficient. Therefore, the model complexity should match the landscape complexity, rather than using TabPFN as a default replacement for traditional surrogates.

\textbf{Offline Optimization with Sufficient Historical Data.}
TabPFN is particularly promising for offline optimization when sufficient historical data are available before the search. The case study results show that TabPFN is less competitive with limited online data, but becomes clearly advantageous when the evaluation budget is used to construct the training set in advance. This suggests that TabPFN benefits more from sufficient data availability than from frequent model reconstruction with small incremental batches. Therefore, when historical evaluations are available, they should be exploited to build an informative surrogate before optimization rather than relying on repeated updates under data-scarce conditions.

\textbf{Online Small/Medium-Scale Single-objective Optimization.} 
TabPFN demonstrates significant potential as a surrogate in online single-objective optimization for small to medium-scale problems. When the surrogate is continuously updated with newly evaluated samples during the optimization process, TabPFN provides highly accurate fitness approximation and search guidance. Furthermore, its predictions can be effectively integrated with advanced model management strategies, such as multi-objective infill criteria that jointly consider predicted fitness and prediction uncertainty \cite{MGO-SLPSO}. These characteristics suggest that TabPFN is highly promising for developing novel SAEAs, offering a strong and flexible foundation for future algorithm design.

\textbf{Batch Candidate Evaluation Scenarios.} TabPFN is particularly well suited to algorithms that generate a large number of candidate solutions at once and then perform surrogate screening in a unified manner. TabPFN can leverage GPUs for efficient batch inference, making it faster than other conventional surrogate models.

\subsection{Limitations and Unsuitable Scenarios of TabPFN}

\textbf{Relatively High Training Time.} 
Although TabPFN does not involve conventional hyper-parameter optimization, its training process still requires encoding and caching the training context to construct the predictive mapping. As shown in Fig.~\ref{fig:time}, under small-sample settings, the training time of TabPFN remains substantially higher than that of RBFN, with an order-of-magnitude gap in some cases. Moreover, at low dimensionalities, TabPFN can even require more training time than GP. Although such differences are modest for a single surrogate construction, they can accumulate into substantial computational overhead when the surrogate is repeatedly rebuilt during the search. A typical example is multisubspace modeling and search in decomposition-based large-scale optimization. For instance, SADE-AMSS \cite{sadeamss} trains 20 surrogate models in each generation, resulting in on the order of $2\times10^{5}$ surrogate constructions over a complete run. Directly replacing its lightweight RBFN surrogate with TabPFN would therefore introduce considerable computational overhead. In contrast, frameworks such as LSEO-S3 \cite{lseo-s3}, which construct relatively few surrogate models per iteration, may incur lower surrogate-construction overhead when incorporating TabPFN.

\textbf{Conservative Extrapolation under Distribution Shifts.}
TabPFN exhibits relatively conservative extrapolation behavior when candidate solutions move away from the distribution of the training data. This characteristic is clearly reflected in the large-scale optimization experiments. 
This behavior is also observed in constrained optimization and appears to explain the performance degradation when replacing the objective surrogates with TabPFN.
Although TabPFN achieves better predictive accuracy than RBFN in most tested subspace scenarios, LSEO-S3-TabPFN performs significantly worse than the original RBFN-based LSEO-S3 on all five test problems, indicating that predictive accuracy alone is insufficient to determine the effectiveness of a surrogate for optimization. As shown in Table~\ref{tab:dist_scale_preds}, the predictions of RBFN become increasingly extreme as the distance from the training data increases, whereas TabPFN remains close to the mean of the training labels, exhibiting mean-regression under distribution shifts. Under the minimum-prediction-based candidate selection adopted by LSEO-S3, the aggressive extrapolation of RBFN implicitly favors solutions farther away from the observed data and consequently introduces additional global exploration. In contrast, the conservative extrapolation of TabPFN does not provide such an implicit search bias, which partly explains its inferior optimization performance despite its higher predictive accuracy.

\textbf{Vulnerability to Complex Constraints beyond Pre-training Priors.}
As a pre-trained Transformer model, TabPFN's effectiveness is fundamentally bounded by the synthetic priors it acquired during training. Our findings in constrained optimization highlight a critical failure mode: when confronted with highly irregular and severely non-linear constraint landscapes, which typically fall outside its pre-trained distribution, TabPFN experiences catastrophic variance shrinkage. This pronounced predictive variance collapse leads to excessively smoothed predictions, thereby blurring the strict boundaries required to distinguish feasible from infeasible regions. Consequently, the surrogate loses its ability to guide the search towards feasibility. This indicates that, without additional mechanisms to handle out-of-distribution topographies, TabPFN is ill-suited to serve as a direct constraint surrogate for optimization problems characterized by artificially complex or extremely non-linear constraint boundaries.

\textbf{Limited Prediction Performance in Mixed-variable Problems.} Although TabPFN can directly handle discrete variables without additional encoding, its prediction accuracy and ranking capability in mixed-variable optimization are still inferior to those of surrogate models specifically designed for mixed-variable problems. As shown in Table~\ref{tab:mixed_surrogate_statistical_comparison}, RBFNmv significantly outperforms TabPFN on most test instances. This discrepancy in predictive performance further affects the optimization process and disrupts the search balance in the original algorithm. Therefore, TabPFN is not suitable for directly replacing specialized mixed-variable surrogates.

\textbf{Limited Effectiveness in Data-scarce Online Optimization.}
TabPFN is less effective in online optimization when only a small number of evaluated samples are available and the training set grows gradually during the search. The results in Table \ref{tab:aiea_summary} show that under such data-scarce conditions, TabPFN does not consistently outperform alternative approximation mechanisms, especially in the early stages of optimization. Frequent surrogate updates with only a few newly collected samples cannot fully compensate for the lack of sufficient training data. Therefore, TabPFN is not well suited to online SAEAs with very limited initial data and small incremental updates.

\subsection{Practical Guidelines for Effective Use of TabPFN}

\textbf{Perform Batch Prediction and Use GPUs for Inference.} TabPFN is well suited for batch inference on large-scale data, with per-sample inference time decreasing as the batch size increases.

\textbf{Exploit Sufficient and Informative Data Whenever Possible.}
TabPFN should be provided with as much informative training data as reasonably available. Unlike conventional surrogates that are repeatedly fitted from scratch, TabPFN performs prediction through in-context learning, conditioning on the available labeled samples to infer the underlying mapping relationship. A richer training set therefore provides more evidence for identifying task-specific structures and enables more reliable predictions and rankings of candidate solutions. Therefore, when historical or offline evaluations are available, they should be incorporated into the training set rather than relying only on a small subset of recent samples.

\textbf{Select Surrogate Models According to Landscape Complexity.}
In practical SAEAs, conventional surrogates can be used for simple or relatively smooth fitness landscapes, while TabPFN is more suited for complex, nonlinear, or multimodal landscapes that require stronger approximation capability. Therefore, the surrogate model should be selected adaptively according to the problem landscape, rather than replacing traditional surrogates with TabPFN by default.
 
\textbf{Customized Model Management Strategies.} 
In SAEAs, model management strategies play a critical role in translating the predictive capability of a surrogate model into effective search behavior. Since these strategies are often closely tailored to the characteristics of the underlying surrogate, directly replacing a conventional surrogate with TabPFN without adapting the corresponding model management strategy may fail to fully exploit its predictive advantages. Therefore, TabPFN-specific model management strategies should be developed for different optimization scenarios. A representative example can be found in the large-scale optimization experiments. As discussed above, RBFN exhibits aggressive extrapolation away from the training data. When the infill criterion selects candidates solely according to their predicted objective values, such extrapolation implicitly favors distant solutions and thereby provides additional exploration, which contributes to its superior optimization performance. In contrast, TabPFN tends to produce more conservative predictions under distribution shifts and therefore lacks this implicit exploration mechanism. In this case, equipping TabPFN with a more exploration-oriented infill criterion, for example by explicitly incorporating uncertainty or solution diversity, may compensate for its conservative extrapolation behavior and achieve a better balance between exploration and exploitation.

\textbf{Differentiated Algorithm Design Guided by Role-specific Characteristics.} 
In complex scenarios such as constrained or multi-objective optimization, objective and constraint functions often exhibit fundamentally different landscape characteristics. Consequently, TabPFN's effectiveness is highly dependent on the specific role it plays within the evolutionary framework, and this behavior can vary drastically across problem domains. For instance, in single-objective constrained optimization (e.g., CEC2010), TabPFN struggles with highly non-linear constraints due to severe prediction range compression, making it highly unreliable as a constraint surrogate. Conversely, in constrained multi-objective optimization (e.g., the CF test suite), a contrasting behavior is observed: because the constraints are relatively simple, TabPFN models them effectively, whereas applying it to the objectives degrades performance, as the multi-objective search process is extremely sensitive to the predictions guiding the population toward the Pareto front. Given these contrasting behaviors, the design of future SAEAs must explicitly account for the distinct requirements of different algorithmic components, and a uniform application of TabPFN across all surrogate roles is generally not advisable. Instead, researchers should adopt a role-specific design philosophy. In practice, offline data can be leveraged to conduct a preliminary offline evaluation of TabPFN's modeling capability on the target problem. By analyzing metrics such as cross-validation error and prediction range compression on this initial data, designers can gain critical insights into the problem's specific characteristics. This evaluation can then inform a more rational algorithm design, ensuring that TabPFN is strategically incorporated into the roles (objectives or constraints) where it provides a clear advantage, while avoiding scenarios where its specific predictive behaviors might mislead the search process.

In summary, TabPFN exhibits both strengths and limitations in SAEAs, and its suitability depends on data availability, landscape characteristics, and its role within the optimization framework. Such scenario dependence highlights that its value as a surrogate lies not only in predicting fitness accurately but also in guiding the search effectively. A more accurate surrogate does not necessarily yield better optimization outcomes, as its extrapolation behavior and interaction with candidate selection also shape the balance between exploration and exploitation.   These findings highlight the importance of co-designing surrogate models and model management strategies when integrating foundation models into evolutionary optimization. Rather than simply replacing conventional surrogates, algorithm designers should tailor data utilization, infill sampling, and surrogate roles to the model’s characteristics, so that its predictive strengths can be effectively translated into improved search performance.

\section{Conclusions} \label{S6}
In this work, we conducted an empirical study to investigate the applicability and performance of TabPFNv2 as a surrogate model within SAEAs. We first examined its fundamental prediction capabilities and then integrated it into several representative SAEAs to evaluate its performance across different expensive optimization scenarios. Through these experiments, we obtained empirical observations regarding its predictive behavior and its impact on the optimization process. The results indicate that TabPFNv2 achieves competitive performance in some scenarios, while its effectiveness is limited in others. These findings highlight both the potential and limitations of applying TabPFNv2 in SAEAs. In particular, TabPFNv2 does not consistently outperform conventional surrogate models and should not be regarded as a universal, plug-and-play replacement. Instead, its applicability should be considered in relation to problem characteristics and the specific role of the surrogate model within the optimization algorithm. These observations provide useful insights for future research on incorporating tabular foundation models into surrogate-assisted evolutionary optimization.

\bibliographystyle{unsrt}
\bibliography{Reference}   

\end{sloppypar}
\end{document}


\title{Supplementary File for ``Benchmarking Tabular Foundation Models as Surrogates in Expensive Evolutionary Optimization"}

	
\markboth{}%
{Shell \MakeLowercase{\textit{et al.}}: A Sample Article Using IEEEtran.cls for IEEE Journals}


\maketitle

\section{Experimental Details} 

\subsection{Prediction Performance of TabPFN}
\label{ppoftabpfn}

GP uses an anisotropic squared exponential kernel \(k=\text{ConstantKernel}\times\text{RBF}\). It independently optimizes the length scale for each dimension, sets the initial length scale to \(\sqrt{D}\), uses a common search range of \([0.1\sqrt{D},\,10\sqrt{D}]\), sets \(\alpha=10^{-8}\), and restarts hyperparameter optimization 10 times. RBFN uses a Gaussian kernel, sets the number of hidden nodes to \(2\sqrt{N_{\mathrm{tr}}}\), determines centers by KMeans, sets the width to twice the average pairwise distance among centers, and uses pseudo-inverse linear fitting with bias in the output layer. TabPFN uses the TabPFNv2 regressor configuration and runs on a GPU without additional modification of its model hyperparameters. The predictive variance of the TabPFNv2 regressor is obtained directly from its full predictive distribution. Specifically, the model generates the full prediction outputs for the input samples, including the distribution-related logits and the corresponding criterion. The predictive variance is then computed by applying the variance function of the criterion to the logits. This variance quantifies the uncertainty associated with the model predictions and is subsequently used as an uncertainty measure in the analysis and evaluation of the regression results.

\subsection{TT-DDEA and DDEA-TabPFN}

TT-DDEA is implemented following the original paper. We compare TT-DDEA-RBFN with two TabPFNv2-based variants: TT-DDEA-TabPFN, which replaces the original RBFN surrogates while retaining the pseudo-labeling strategy, and DDEA-TabPFN, which trains a single TabPFNv2 model on the entire offline dataset without model management. All algorithms are evaluated on 29 CEC 2017 functions (excluding F2) with $D\in\{10,30,100\}$ over 25 independent runs. Each run uses $11D$ offline samples generated by Latin hypercube sampling, with no additional true evaluations during optimization. The population size and maximum number of generations are both set to 100, with crossover and mutation probabilities of 1 and $1/D$, respectively. The remaining evolutionary settings are kept unchanged.


\subsection{MGP-SLPSO and TabPFN-SLPSO}

MGP-SLPSO is implemented following the original paper. We evaluate MGP-SLPSO and its TabPFN-based variant on 29 CEC2017 functions (excluding F2) with $D\in\{10,30,100\}$, 25 independent runs, and a true-evaluation budget of $11D$. The population size is set to $\min(5D,100+D/10)$, resulting in 50, 103, and 110 particles for $D=10,30,100$, respectively. The LHS sample size equals the initial population size. The predictive mean and variance of TabPFNv2 are used in the original infill criterion in TabPFN-SLPSO. At each generation, all distinct particles in the first and last non-dominated fronts are selected for true evaluation, while duplicate particles reuse their archived fitness values. The TabPFN-based variant retains the original evolutionary operators, infill mechanisms, and model-management settings, with no additional evolutionary parameter tuning.

\subsection{LSEO-S3 and LSEO-S3-TabPFN}
\label{lseoexper}
LSEO-S3 is implemented following the original paper. We evaluate LSEO-S3 and its TabPFN-based variant on five representative CEC2013 functions (F1, F4, F8, F12, and F15), each with \(D=1000\), using 25 independent runs and a maximum budget of \(11D\) true function evaluations per run. LSEO-S3-TabPFN replaces the original RBFN surrogate component with TabPFNv2 while retaining the original optimization framework. Both algorithms use a population size of \(N=100\), an initial archive of \(N_{\mathrm{Arc}}=500\) true-evaluated samples generated by LHS, and an initial subproblem dimension of \(D_r=100\).
For the prediction experiments, both surrogate models are evaluated in 100-dimensional subspaces with training and test sizes of \(N_{\mathrm{tr}}=N_{\mathrm{te}}=300\). Training points are generated using LHS. Test points are generated under two scenarios: Global, using LHS over the subspace domain; and Local, using Gaussian perturbations centered at the mean of the training inputs, with a standard deviation equal to \(5\%\) of the search range. RMSE and Kendall’s \(\tau\) are averaged over three independent runs.
For the extrapolation diagnostic on F8, we collect runtime states, including the training samples, active subspace dimensions \(D_s\), and the global best solution, from 15 LSEO-S3 iterations, with a training size of \(N_s=3D_s\) for each state. The characteristic distance \(\ell\) is defined as the median nearest-neighbor distance among the training inputs, i.e., \(\ell=\mathrm{median}_i\min_{j\ne i}\lVert\mathbf{x}_i-\mathbf{x}_j\rVert_2\). At each state, test points are generated from the best training point along 30 random unit directions at nominal distances \(r\ell\), where \(r\in\{0.25,0.5,1,2,4,8,16\}\). This yields 450 test points per distance level across the 15 states. We report the mean prediction of each model ($\hat{y}_{\text{RBFN}}$ and $\hat{y}_{\text{TabPFN}}$), the mean training label $y_{\text{train\_mean}}$, and the mean minimum distance from the test points to the training set (real\_dist).

\subsection{DSI and DSI-TabPFN}

DSI-C$^2$oDE is implemented following the original paper. We compare DSI-RBFN with DSI-TabPFN on ten selected CEC2006/CEC2010 problems over 25 independent runs. Each run uses 30 initial samples, a population size of 30, a maximum of 300 true evaluations, and an initial $w_{\max}=10$, with $w_{\max}$ adjusted by the original mechanism within $[5,80]$. DSI-TabPFN replaces the original objective and constraint RBF surrogates with TabPFNv2, using one model for each output. The remaining C$^2$oDE settings are kept unchanged. To characterize constraint prediction-range compression, 5000 points are uniformly sampled from the decision bounds, and a degree-2 polynomial is fitted to each constraint, with the smallest value among the constraints reported. For the range analysis, 100 training and 50 test points are uniformly generated from the decision bounds. RBFN and TabPFNv2 use the same training and test points, and each constraint is modeled independently. The range-compression ratio is defined as $\rho=(\max\hat g-\min\hat g)/(\max g-\min g)$. These diagnostic samples are used only for analysis and do not update the archive or consume the true-evaluation budget.

\subsection{KTA2 and KTA2-TabPFN}
\label{kta2tabpfn}

The experiments consider only the unconstrained DTLZ1--DTLZ4 problems, with ten decision variables defined on \([0,1]^{10}\) and \(M\in\{3,10\}\) objectives. For each problem and objective setting, KTA2 and KTA2-TabPFN use the same evolutionary parameters: a maximum expensive-evaluation budget of \(300\), an initial population size and archive size of \(100\), \(\tau=0.75\), \(\phi=0.1\), \(w_{\max}=10\), and \(\mu=5\). The initial population is generated by Latin hypercube sampling and is fully evaluated, giving \(100\) initial function evaluations. KTA2 uses an KRG surrogate for each objective with a constant polynomial trend and a squared-exponential correlation function; the initial correlation parameters are \(5\) and are bounded by \([10^{-5},100]\), and the optimized parameters are reused as initial values in the next retraining step. KTA2-TabPFN replaces these objective-wise KRG models with default TabPFNv2 regressors and applies no additional input or output standardization; all other settings remain unchanged. The parameter \(p=1/M\) is used in distance calculations. Each problem and objective setting is independently repeated \(25\) times, and the final diversity archive is saved for IGD\(^{+}\) evaluation.

\subsection{KTS, KTS-TabPFN, KTS-TabPFN(obj) and KTS-TabPFN(cons)}
\label{ktstabpfn}

We evaluate the four algorithms on the ten CF benchmark problems, each with 10 decision variables; CF1--CF7 have two objectives, CF8--CF10 have three objectives, and CF6--CF7 have two constraints while the remaining problems have one. Every run uses high-fidelity evaluations, begins with a 100-point Latin hypercube design generated with a run-specific random seed, and terminates after 400 total function evaluations. KTS maintains constraint-aware and objective-oriented populations of size 100 and adaptively switches between its unconstrained and constrained search modes according to the correlation between convergence contribution and constraint violation: it selects constrained search when the correlation is below $-0.2$, unconstrained search when it exceeds $0.6$, and randomly selects either mode otherwise. The baseline uses independent KRG surrogates with a constant trend and squared-exponential kernel for each objective and constraint, whereas KTS-TabPFN replaces both surrogate sets with independent TabPFNv2 regressors; KTS-TabPFN(obj) replaces only the objective surrogates, and KTS-TabPFN(cons) replaces only the constraint surrogates. All variants retain the same initialization, adaptive switching, evolutionary operators, population updates, sampling procedures, and model-refitting schedule.

\subsection{SHEALED and SHEALED-TabPFN}

SHEALED is implemented following the original paper. We evaluate SHEALED and its TabPFN-based variant on ten mixed-variable benchmark problems, comprising ten functions, with dimension \(D \in \{10, 30, 50\}\) and \(n_1 = n_2 = D/2\), where \(n_1\) and \(n_2\) denote the continuous and discrete dimensions, respectively. Each discrete variable has five candidate values. Both algorithms use a population size of 100, an initial sample size of 100, and a total budget of 600 true evaluations. Each problem is independently run 25 times with random seeds ranging from 42 to 66. SHEALED-TabPFN replaces the original RBFNmv surrogate with TabPFNv2, while retaining the original search operators and model management strategies. Discrete variables are treated as categorical features using integer indices without one-hot encoding, and no external target scaling is applied. Each outer iteration performs up to three true evaluations via global search and local search. New samples are recorded and the corresponding surrogates are refitted when their search stages are activated. Optimization terminates when the evaluation budget is exhausted. For the prediction experiments, RBFNmv and TabPFN use identical training and test sets with sizes \(N_{\mathrm{tr}} = 10D\) and \(N_{\mathrm{te}} = 2000\), respectively. In both experiments, continuous variables are generated using Latin hypercube sampling, while discrete variables are sampled with balanced category frequencies. Each prediction setting is repeated 25 times.

\subsection{AIEA and TabPFNEA}
The experiments for embodied robot design were conducted on 30 tasks from the EvoGym benchmark, covering three difficulty levels: easy, medium, and hard. Each robot morphology was represented by a \(5 \times 5\) voxel grid, where each voxel was encoded as an integer from \(\{0,1,2,3,4\}\), corresponding to empty, rigid, soft, horizontal actuator, and vertical actuator, respectively. To ensure the validity of robot morphologies, each design was required to be connected and contain at least one actuator. The fitness of a robot morphology was defined as the cumulative task reward obtained by executing its controller, with higher fitness values indicating better performance. Specifically, a real evaluation involved training a controller using Proximal Policy Optimization (PPO) and evaluating the corresponding robot morphology with the trained controller. The population size was set to 20, and each algorithm was independently executed five times on each task.

Two experimental settings were considered: online and offline optimization. The real evaluation budgets were set to 100, 200, and 300 for easy, medium, and hard tasks, respectively, in both settings. In the online setting, the optimization process started with 20 evaluated robot morphologies, and promising candidates were progressively selected for real evaluations to update the training dataset. The shared algorithmic parameters were kept consistent with those used in AIEA. In the offline setting, datasets containing 100, 200, and 300 randomly generated and evaluated robot morphologies were constructed for easy, medium, and hard tasks, respectively. Subsequent optimization was performed exclusively using these fixed datasets, without any additional real evaluations.

\section{Detailed Experimental Data}

\begin{table}[htbp]
  \centering
  \caption{Predictive performance mean comparison of GP, RBFN, and TabPFN on the CEC 2017 benchmark suite, with input dimension $D=10$, training size $N_{\mathrm{tr}}=100$, and test size $N_{\mathrm{te}}=1000$. F2 is excluded. For GP and RBFN, the symbols \(+\), \(=\), and \(-\) indicate significantly better, not significantly different, or significantly worse than TabPFN, respectively, according to the Wilcoxon rank-sum test at $\alpha=0.05$.}
  \resizebox{\textwidth}{!}{%
  \begin{tabular}{ccccccc}
  \toprule
  \multirow{2}{*}{\textbf{F}}
    & \multicolumn{2}{c}{\textbf{GP}}
    & \multicolumn{2}{c}{\textbf{RBFN}}
    & \multicolumn{2}{c}{\textbf{TabPFN}} \\
  \cmidrule{2-7}
    & \textbf{RMSE} & $\boldsymbol{\tau}$
    & \textbf{RMSE} & $\boldsymbol{\tau}$
    & \textbf{RMSE} & $\boldsymbol{\tau}$ \\
  \midrule
  1
    & \textbf{4.071e+08 (8.886e+07)}\,\(+\)
    & \textbf{9.928e-01 (1.601e-03)}\,\(+\)
    & 1.261e+10 (5.808e+08)\,\(-\)
    & 7.573e-01 (1.631e-02)\,\(-\)
    & 8.168e+09 (5.714e+08)
    & 8.306e-01 (1.403e-02) \\
  3
    & \textbf{7.749e+11 (1.450e+11)}\,\(=\)
    & 4.784e-01 (8.851e-02)\,\(-\)
    & 1.044e+12 (1.207e+11)\,\(-\)
    & 1.774e-01 (7.661e-02)\,\(-\)
    & 8.493e+11 (1.616e+11)
    & \textbf{5.313e-01 (7.587e-02)} \\
  4
    & 1.010e+04 (1.496e+03)\,\(=\)
    & 6.905e-01 (3.374e-02)\,\(-\)
    & 1.265e+04 (7.953e+02)\,\(-\)
    & 6.466e-01 (3.228e-02)\,\(-\)
    & \textbf{9.527e+03 (1.873e+03)}
    & \textbf{7.406e-01 (3.217e-02)} \\
  5
    & 4.818e+01 (4.266e+00)\,\(-\)
    & 7.532e-01 (2.278e-02)\,\(-\)
    & 3.921e+01 (1.398e+00)\,\(-\)
    & 7.969e-01 (8.251e-03)\,\(-\)
    & \textbf{3.653e+01 (3.162e+00)}
    & \textbf{8.061e-01 (1.642e-02)} \\
  6
    & 5.140e+01 (3.068e+00)\,\(-\)
    & 3.361e-01 (4.645e-02)\,\(-\)
    & \textbf{4.102e+01 (1.393e+00)}\,\(+\)
    & \textbf{4.970e-01 (1.667e-02)}\,\(+\)
    & 4.370e+01 (1.862e+00)
    & 4.509e-01 (2.348e-02) \\
  7
    & 1.241e+02 (7.650e+01)\,\(+\)
    & 7.403e-01 (1.720e-01)\,\(+\)
    & \textbf{3.661e+01 (3.797e+00)}\,\(+\)
    & \textbf{9.291e-01 (8.107e-03)}\,\(+\)
    & 1.598e+02 (1.637e+01)
    & 6.810e-01 (4.229e-02) \\
  8
    & 4.599e+01 (3.768e+00)\,\(-\)
    & 5.719e-01 (3.906e-02)\,\(-\)
    & 4.118e+01 (1.782e+00)\,\(-\)
    & 6.121e-01 (1.795e-02)\,\(-\)
    & \textbf{3.636e+01 (1.982e+00)}
    & \textbf{6.580e-01 (2.224e-02)} \\
  9
    & 9.725e+03 (6.333e+02)\,\(-\)
    & 3.862e-01 (4.941e-02)\,\(-\)
    & 8.402e+03 (4.239e+02)\,\(=\)
    & \textbf{4.864e-01 (2.459e-02)}\,\(+\)
    & \textbf{8.271e+03 (4.927e+02)}
    & 4.657e-01 (3.486e-02) \\
  10
    & 7.020e+02 (2.890e+01)\,\(-\)
    & 2.226e-02 (2.515e-02)\,\(-\)
    & 6.884e+02 (2.611e+01)\,\(-\)
    & 5.307e-02 (3.012e-02)\,\(=\)
    & \textbf{6.374e+02 (2.078e+01)}
    & \textbf{5.898e-02 (3.219e-02)} \\
  \midrule
  11
    & \textbf{1.145e+08 (4.124e+07)}\,\(+\)
    & 8.714e-01 (1.532e-02)\,\(-\)
    & 1.936e+09 (1.141e+08)\,\(-\)
    & 5.295e-01 (2.959e-02)\,\(-\)
    & 2.661e+08 (1.404e+08)
    & \textbf{9.719e-01 (5.141e-03)} \\
  12
    & \textbf{3.875e+08 (6.015e+07)}\,\(+\)
    & \textbf{9.823e-01 (2.897e-03)}\,\(+\)
    & 6.201e+09 (3.289e+08)\,\(-\)
    & 6.300e-01 (1.942e-02)\,\(-\)
    & 4.367e+09 (5.260e+08)
    & 7.461e-01 (3.649e-02) \\
  13
    & \textbf{1.829e+08 (2.040e+07)}\,\(+\)
    & \textbf{9.851e-01 (1.963e-03)}\,\(+\)
    & 4.242e+09 (1.387e+08)\,\(-\)
    & 6.188e-01 (1.855e-02)\,\(-\)
    & 3.038e+08 (7.928e+07)
    & 9.792e-01 (4.461e-03) \\
  14
    & \textbf{5.269e+07 (1.386e+07)}\,\(+\)
    & 9.891e-01 (1.473e-03)\,\(-\)
    & 3.328e+09 (1.859e+08)\,\(-\)
    & 6.442e-01 (1.832e-02)\,\(-\)
    & 1.728e+08 (1.646e+08)
    & \textbf{9.913e-01 (2.301e-03)} \\
  15
    & \textbf{1.400e+08 (2.931e+07)}\,\(+\)
    & 9.883e-01 (1.466e-03)\,\(-\)
    & 9.360e+09 (5.580e+08)\,\(-\)
    & 2.962e-01 (3.844e-02)\,\(-\)
    & 3.681e+08 (3.230e+08)
    & \textbf{9.897e-01 (1.751e-03)} \\
  16
    & \textbf{3.129e+03 (9.060e+02)}\,\(=\)
    & 8.010e-01 (2.369e-02)\,\(-\)
    & 2.168e+04 (1.257e+03)\,\(-\)
    & 3.648e-01 (3.172e-02)\,\(-\)
    & 3.471e+03 (2.034e+03)
    & \textbf{8.794e-01 (1.676e-02)} \\
  17
    & \textbf{3.287e+05 (1.414e+05)}\,\(=\)
    & 7.127e-01 (4.271e-02)\,\(-\)
    & 1.206e+06 (9.320e+04)\,\(-\)
    & 4.155e-01 (2.872e-02)\,\(-\)
    & 3.468e+05 (1.873e+05)
    & \textbf{9.600e-01 (5.119e-03)} \\
  18
    & \textbf{5.340e+08 (8.254e+07)}\,\(+\)
    & \textbf{9.856e-01 (1.059e-03)}\,\(+\)
    & 1.224e+10 (7.126e+08)\,\(-\)
    & 6.857e-01 (1.918e-02)\,\(-\)
    & 1.616e+09 (3.944e+08)
    & 9.586e-01 (8.092e-03) \\
  19
    & \textbf{1.143e+09 (2.939e+08)}\,\(+\)
    & 9.496e-01 (7.420e-03)\,\(-\)
    & 9.873e+09 (5.378e+08)\,\(-\)
    & 7.781e-01 (1.682e-02)\,\(-\)
    & 1.282e+09 (4.367e+08)
    & \textbf{9.620e-01 (3.891e-03)} \\
  20
    & 3.739e+02 (1.575e+01)\,\(-\)
    & 9.697e-02 (3.899e-02)\,\(-\)
    & 3.503e+02 (1.224e+01)\,\(-\)
    & 1.696e-01 (4.004e-02)\,\(-\)
    & \textbf{3.295e+02 (1.079e+01)}
    & \textbf{2.110e-01 (5.134e-02)} \\
  \midrule
  21
    & 9.422e+02 (1.375e+02)\,\(-\)
    & 3.851e-01 (7.334e-02)\,\(-\)
    & 8.970e+02 (8.635e+01)\,\(-\)
    & 5.022e-01 (5.034e-02)\,\(-\)
    & \textbf{7.991e+02 (8.223e+01)}
    & \textbf{6.627e-01 (2.352e-02)} \\
  22
    & 7.979e+02 (4.725e+01)\,\(-\)
    & 1.321e-01 (6.620e-02)\,\(-\)
    & 7.839e+02 (3.130e+01)\,\(-\)
    & \textbf{1.868e-01 (3.556e-02)}\,\(+\)
    & \textbf{7.555e+02 (3.513e+01)}
    & 1.579e-01 (4.230e-02) \\
  23
    & 5.160e+02 (5.089e+01)\,\(-\)
    & 6.880e-01 (3.873e-02)\,\(-\)
    & 4.432e+02 (2.025e+01)\,\(-\)
    & 7.672e-01 (2.547e-02)\,\(-\)
    & \textbf{3.852e+02 (2.233e+01)}
    & \textbf{7.995e-01 (1.297e-02)} \\
  24
    & 5.866e+02 (7.252e+01)\,\(-\)
    & 5.479e-01 (6.477e-02)\,\(-\)
    & 5.085e+02 (3.719e+01)\,\(-\)
    & 6.500e-01 (3.185e-02)\,\(-\)
    & \textbf{4.794e+02 (4.457e+01)}
    & \textbf{7.116e-01 (2.297e-02)} \\
  25
    & 4.839e+03 (5.226e+02)\,\(-\)
    & 5.230e-01 (6.689e-02)\,\(-\)
    & \textbf{4.118e+03 (2.022e+02)}\,\(+\)
    & \textbf{6.091e-01 (2.176e-02)}\,\(+\)
    & 4.530e+03 (3.155e+02)
    & 5.623e-01 (3.890e-02) \\
  26
    & 2.425e+03 (4.008e+02)\,\(=\)
    & 5.459e-01 (6.852e-02)\,\(=\)
    & 2.486e+03 (1.737e+02)\,\(-\)
    & 5.458e-01 (4.018e-02)\,\(-\)
    & \textbf{2.300e+03 (2.093e+02)}
    & \textbf{5.721e-01 (2.907e-02)} \\
  27
    & \textbf{1.237e+03 (1.761e+02)}\,\(+\)
    & \textbf{7.453e-01 (2.512e-02)}\,\(+\)
    & 3.009e+03 (1.712e+02)\,\(-\)
    & 3.717e-01 (3.828e-02)\,\(-\)
    & 1.661e+03 (3.613e+02)
    & 7.178e-01 (5.499e-02) \\
  28
    & \textbf{1.569e+03 (1.460e+02)}\,\(+\)
    & \textbf{6.156e-01 (3.657e-02)}\,\(+\)
    & 1.936e+03 (1.466e+02)\,\(-\)
    & 5.342e-01 (2.782e-02)\,\(-\)
    & 1.687e+03 (2.014e+02)
    & 5.764e-01 (2.760e-02) \\
  29
    & 2.398e+07 (3.743e+07)\,\(=\)
    & 2.861e-01 (1.081e-01)\,\(-\)
    & 1.757e+07 (1.042e+07)\,\(-\)
    & 3.354e-01 (1.022e-01)\,\(-\)
    & \textbf{1.302e+07 (3.941e+06)}
    & \textbf{6.119e-01 (3.362e-02)} \\
  30
    & \textbf{1.466e+09 (1.437e+08)}\,\(+\)
    & \textbf{7.584e-01 (1.901e-02)}\,\(+\)
    & 4.104e+09 (3.646e+08)\,\(-\)
    & 2.845e-01 (5.928e-02)\,\(-\)
    & 1.866e+09 (2.331e+08)
    & 7.062e-01 (3.982e-02) \\
  \midrule
    $(+/-/=)$
    & \textbf{12/11/6}
    & \textbf{8/20/1}
    & \textbf{3/25/1}
    & \textbf{5/23/1}
    & -- & -- \\   
  \bottomrule
  \end{tabular}%
  }
  \label{cec2017_10d}%
\end{table}%

\begin{table}[htbp]
  \centering
  \caption{Predictive performance mean comparison of GP, RBFN, and TabPFN on the CEC 2017 benchmark suite, with input dimension $D=30$, training size $N_{\mathrm{tr}}=300$, and test size $N_{\mathrm{te}}=1000$. F2 is excluded. For GP and RBFN, the symbols \(+\), \(=\), and \(-\) indicate significantly better, not significantly different, or significantly worse than TabPFN, respectively, according to the Wilcoxon rank-sum test at $\alpha=0.05$.}
  \resizebox{\textwidth}{!}{%
  \begin{tabular}{ccccccc}
  \toprule
  \multirow{2}{*}{\textbf{F}}
    & \multicolumn{2}{c}{\textbf{GP}}
    & \multicolumn{2}{c}{\textbf{RBFN}}
    & \multicolumn{2}{c}{\textbf{TabPFN}} \\
  \cmidrule{2-7}
    & \textbf{RMSE} & $\boldsymbol{\tau}$
    & \textbf{RMSE} & $\boldsymbol{\tau}$
    & \textbf{RMSE} & $\boldsymbol{\tau}$ \\
  \midrule
  1 & \textbf{2.139e+10 (1.533e+09)}\,\(+\) & \textbf{7.746e-01 (1.752e-02)}\,\(+\) & 2.209e+10 (1.004e+09)\,\(+\) & 7.670e-01 (1.250e-02)\,\(+\) & 2.406e+10 (1.366e+09) & 7.401e-01 (1.564e-02) \\
  3 & \textbf{1.632e+15 (3.267e+14)}\,\(+\) & \textbf{4.091e-01 (7.010e-02)}\,\(+\) & 2.090e+15 (3.241e+14)\,\(-\) & 3.399e-02 (2.404e-02)\,\(-\) & 1.830e+15 (3.187e+14) & 3.330e-01 (7.701e-02) \\
  4 & 4.411e+04 (2.694e+03)\,\(-\) & 6.784e-01 (1.925e-02)\,\(-\) & 4.813e+04 (1.836e+03)\,\(-\) & 6.575e-01 (1.831e-02)\,\(-\) & \textbf{4.134e+04 (2.121e+03)} & \textbf{6.993e-01 (1.658e-02)} \\
  5 & 9.118e+01 (5.495e+00)\,\(-\) & 6.869e-01 (2.507e-02)\,\(-\) & \textbf{7.657e+01 (2.362e+00)}\,\(+\) & \textbf{7.381e-01 (1.336e-02)}\,\(+\) & 8.497e+01 (3.472e+00) & 7.058e-01 (1.545e-02) \\
  6 & 2.923e+01 (1.156e+00)\,\(-\) & 3.555e-01 (3.175e-02)\,\(-\) & \textbf{2.255e+01 (7.261e-01)}\,\(+\) & \textbf{5.242e-01 (2.166e-02)}\,\(+\) & 2.461e+01 (7.388e-01) & 4.706e-01 (2.657e-02) \\
  7 & 5.729e+02 (2.381e+01)\,\(-\) & 4.786e-01 (2.707e-02)\,\(-\) & \textbf{7.340e+01 (2.392e+01)}\,\(+\) & \textbf{9.408e-01 (2.045e-02)}\,\(+\) & 3.884e+02 (2.124e+01) & 6.636e-01 (2.135e-02) \\
  8 & 8.506e+01 (4.561e+00)\,\(-\) & 6.395e-01 (2.476e-02)\,\(-\) & \textbf{7.356e+01 (2.338e+00)}\,\(+\) & \textbf{6.922e-01 (1.314e-02)}\,\(+\) & 7.785e+01 (3.061e+00) & 6.662e-01 (1.716e-02) \\
  9 & 2.253e+04 (9.943e+02)\,\(-\) & 3.399e-01 (3.759e-02)\,\(-\) & \textbf{1.843e+04 (5.592e+02)}\,\(+\) & \textbf{4.809e-01 (1.869e-02)}\,\(+\) & 1.974e+04 (7.250e+02) & 4.256e-01 (2.358e-02) \\
  10 & 1.188e+03 (2.815e+01)\,\(-\) & 6.726e-03 (2.467e-02)\,\(-\) & 1.132e+03 (1.887e+01)\,\(-\) & \textbf{4.153e-02 (2.409e-02)}\,\(=\) & \textbf{1.083e+03 (1.712e+01)} & 3.982e-02 (2.500e-02) \\
  \midrule
  11 & \textbf{7.586e+09 (2.240e+09)}\,\(+\) & 7.282e-01 (2.445e-02)\,\(-\) & 5.953e+10 (4.607e+09)\,\(-\) & 4.133e-01 (2.501e-02)\,\(-\) & 2.553e+10 (7.847e+09) & \textbf{9.314e-01 (5.396e-03)} \\
  12 & \textbf{3.810e+09 (3.043e+08)}\,\(+\) & \textbf{8.971e-01 (8.383e-03)}\,\(+\) & 9.576e+09 (2.808e+08)\,\(-\) & 7.189e-01 (9.190e-03)\,\(-\) & 7.200e+09 (4.157e+08) & 7.953e-01 (1.278e-02) \\
  13 & \textbf{2.254e+09 (2.012e+08)}\,\(+\) & \textbf{9.753e-01 (1.644e-03)}\,\(+\) & 2.417e+10 (5.842e+08)\,\(-\) & 6.353e-01 (1.558e-02)\,\(-\) & 1.133e+10 (9.040e+08) & 8.396e-01 (1.388e-02) \\
  14 & \textbf{1.776e+08 (2.425e+07)}\,\(+\) & \textbf{9.668e-01 (2.264e-03)}\,\(+\) & 4.450e+09 (2.033e+08)\,\(-\) & 4.475e-01 (2.478e-02)\,\(-\) & 6.028e+08 (1.659e+08) & 9.403e-01 (6.373e-03) \\
  15 & \textbf{1.181e+09 (1.021e+08)}\,\(+\) & \textbf{9.727e-01 (2.007e-03)}\,\(+\) & 1.830e+10 (4.065e+08)\,\(-\) & 4.422e-01 (1.773e-02)\,\(-\) & 4.367e+09 (7.335e+08) & 8.868e-01 (1.787e-02) \\
  16 & \textbf{1.170e+04 (1.288e+03)}\,\(+\) & \textbf{8.753e-01 (7.528e-03)}\,\(+\) & 4.469e+04 (1.999e+03)\,\(-\) & 6.498e-01 (2.228e-02)\,\(-\) & 1.875e+04 (2.450e+03) & 8.464e-01 (1.264e-02) \\
  17 & \textbf{5.035e+07 (8.155e+06)}\,\(+\) & 5.623e-01 (3.621e-02)\,\(-\) & 8.843e+07 (1.125e+07)\,\(-\) & 4.885e-01 (3.058e-02)\,\(-\) & 5.898e+07 (1.245e+07) & \textbf{7.864e-01 (1.716e-02)} \\
  18 & \textbf{1.177e+09 (9.580e+07)}\,\(+\) & \textbf{9.552e-01 (3.175e-03)}\,\(+\) & 1.533e+10 (4.593e+08)\,\(-\) & 3.936e-01 (1.904e-02)\,\(-\) & 7.741e+09 (1.610e+09) & 7.159e-01 (6.348e-02) \\
  19 & \textbf{3.267e+09 (6.352e+08)}\,\(+\) & \textbf{9.258e-01 (1.399e-02)}\,\(+\) & 1.952e+10 (5.761e+08)\,\(-\) & 4.216e-01 (2.216e-02)\,\(-\) & 5.292e+09 (8.018e+08) & 8.740e-01 (2.227e-02) \\
  20 & 5.419e+02 (1.385e+01)\,\(-\) & 3.972e-02 (2.203e-02)\,\(-\) & 5.096e+02 (1.311e+01)\,\(-\) & \textbf{9.784e-02 (2.280e-02)}\,\(+\) & \textbf{4.931e+02 (1.194e+01)} & 8.509e-02 (2.386e-02) \\
  \midrule
  21 & 6.732e+02 (6.708e+01)\,\(-\) & 5.815e-01 (4.077e-02)\,\(-\) & 6.723e+02 (6.336e+01)\,\(-\) & 6.742e-01 (2.146e-02)\,\(-\) & \textbf{6.191e+02 (1.147e+02)} & \textbf{7.468e-01 (1.008e-02)} \\
  22 & 1.193e+03 (3.121e+01)\,\(-\) & 6.415e-02 (2.675e-02)\,\(-\) & 1.120e+03 (2.885e+01)\,\(-\) & \textbf{1.382e-01 (2.350e-02)}\,\(+\) & \textbf{1.101e+03 (2.469e+01)} & 1.146e-01 (3.370e-02) \\
  23 & 5.630e+02 (2.674e+01)\,\(-\) & 7.994e-01 (9.862e-03)\,\(-\) & 5.228e+02 (2.393e+01)\,\(-\) & 8.550e-01 (1.136e-02)\,\(-\) & \textbf{4.218e+02 (2.009e+01)} & \textbf{8.613e-01 (5.561e-03)} \\
  24 & 3.955e+02 (3.351e+01)\,\(-\) & 8.271e-01 (1.798e-02)\,\(-\) & 3.269e+02 (3.099e+01)\,\(-\) & 8.790e-01 (1.314e-02)\,\(-\) & \textbf{2.886e+02 (3.149e+01)} & \textbf{8.941e-01 (7.174e-03)} \\
  25 & 2.309e+04 (1.762e+03)\,\(-\) & 5.680e-01 (2.706e-02)\,\(-\) & 2.695e+04 (1.749e+03)\,\(-\) & 5.747e-01 (2.180e-02)\,\(-\) & \textbf{2.124e+04 (1.398e+03)} & \textbf{6.119e-01 (1.992e-02)} \\
  26 & 1.265e+04 (8.404e+02)\,\(-\) & 6.060e-01 (2.677e-02)\,\(-\) & 1.254e+04 (5.985e+02)\,\(-\) & 6.378e-01 (1.627e-02)\,\(-\) & \textbf{1.126e+04 (5.996e+02)} & \textbf{6.594e-01 (1.834e-02)} \\
  27 & \textbf{1.195e+03 (7.387e+01)}\,\(+\) & \textbf{7.924e-01 (1.326e-02)}\,\(+\) & 2.208e+03 (9.420e+01)\,\(-\) & 6.190e-01 (1.292e-02)\,\(-\) & 1.637e+03 (7.500e+01) & 7.036e-01 (1.573e-02) \\
  28 & \textbf{9.617e+03 (5.575e+02)}\,\(=\) & \textbf{6.136e-01 (2.561e-02)}\,\(+\) & 1.102e+04 (4.181e+02)\,\(-\) & 5.280e-01 (1.940e-02)\,\(-\) & 9.798e+03 (5.230e+02) & 5.918e-01 (2.428e-02) \\
  29 & \textbf{5.367e+08 (1.347e+08)}\,\(+\) & 4.033e-01 (6.820e-02)\,\(-\) & 7.886e+08 (1.412e+08)\,\(-\) & 4.372e-01 (5.438e-02)\,\(-\) & 6.359e+08 (1.587e+08) & \textbf{6.983e-01 (1.656e-02)} \\
  30 & \textbf{2.164e+09 (1.577e+08)}\,\(+\) & \textbf{9.075e-01 (5.323e-03)}\,\(+\) & 6.797e+09 (1.933e+08)\,\(-\) & 6.703e-01 (1.389e-02)\,\(-\) & 2.828e+09 (2.319e+08) & 8.808e-01 (8.452e-03) \\
  \midrule
  $(+/-/=)$
  & \textbf{14/14/1}
  & \textbf{12/17/0}
  & \textbf{6/23/0}
  & \textbf{8/20/1}
  & -- & -- \\
  \bottomrule
  \end{tabular}%
  }
  \label{cec2017_30d}%
\end{table}%

\begin{table}[htbp]
  \centering
  \caption{Predictive performance mean comparison of GP, RBFN, and TabPFN on the CEC 2017 benchmark suite, with input dimension $D=100$, training size $N_{\mathrm{tr}}=1000$, and test size $N_{\mathrm{te}}=1000$. F2 is excluded. For GP and RBFN, the symbols \(+\), \(=\), and \(-\) indicate significantly better, not significantly different, or significantly worse than TabPFN, respectively, according to the Wilcoxon rank-sum test at $\alpha=0.05$.}
  \resizebox{\textwidth}{!}{%
  \begin{tabular}{ccccccc}
  \toprule
  \multirow{2}{*}{\textbf{F}}
    & \multicolumn{2}{c}{\textbf{GP}}
    & \multicolumn{2}{c}{\textbf{RBFN}}
    & \multicolumn{2}{c}{\textbf{TabPFN}} \\
  \cmidrule{2-7}
    & \textbf{RMSE} & $\boldsymbol{\tau}$
    & \textbf{RMSE} & $\boldsymbol{\tau}$
    & \textbf{RMSE} & $\boldsymbol{\tau}$ \\
  \midrule
  1 & 6.170e+10 (2.403e+09)\,\(-\) & 6.217e-01 (1.526e-02)\,\(-\) & 6.724e+10 (5.096e+09)\,\(-\) & 5.740e-01 (3.827e-02)\,\(-\) & \textbf{5.487e+10 (1.585e+09)} & \textbf{6.684e-01 (1.160e-02)} \\
  3 & \textbf{2.452e+18 (4.078e+17)}\,\(+\) & \textbf{2.810e-01 (6.638e-02)}\,\(+\) & 2.778e+18 (3.989e+17)\,\(-\) & 5.726e-02 (2.535e-02)\,\(-\) & 2.714e+18 (4.224e+17) & 8.000e-02 (3.660e-02) \\
  4 & \textbf{1.179e+05 (4.613e+03)}\,\(+\) & \textbf{6.713e-01 (1.564e-02)}\,\(+\) & 1.822e+05 (1.042e+04)\,\(-\) & 4.639e-01 (3.730e-02)\,\(-\) & 1.271e+05 (5.283e+03) & 6.446e-01 (1.555e-02) \\
  5 & 1.836e+02 (5.639e+00)\,\(-\) & 6.054e-01 (1.493e-02)\,\(-\) & 2.040e+02 (1.290e+01)\,\(-\) & 5.436e-01 (3.820e-02)\,\(-\) & \textbf{1.657e+02 (4.127e+00)} & \textbf{6.475e-01 (1.032e-02)} \\
  6 & 1.490e+01 (3.361e-01)\,\(-\) & 3.884e-01 (2.184e-02)\,\(-\) & 1.396e+01 (4.771e-01)\,\(-\) & 4.188e-01 (3.440e-02)\,\(-\) & \textbf{1.350e+01 (2.899e-01)} & \textbf{4.530e-01 (1.540e-02)} \\
  7 & 1.249e+03 (2.857e+01)\,\(-\) & 4.449e-01 (1.654e-02)\,\(-\) & \textbf{7.179e+02 (6.362e+01)}\,\(+\) & \textbf{7.000e-01 (2.851e-02)}\,\(+\) & 9.780e+02 (3.187e+01) & 5.852e-01 (1.579e-02) \\
  8 & 1.845e+02 (5.959e+00)\,\(-\) & 6.569e-01 (1.216e-02)\,\(-\) & 2.326e+02 (1.597e+01)\,\(-\) & 5.446e-01 (3.558e-02)\,\(-\) & \textbf{1.600e+02 (4.492e+00)} & \textbf{7.073e-01 (1.162e-02)} \\
  9 & 4.974e+04 (1.145e+03)\,\(-\) & 3.480e-01 (1.995e-02)\,\(-\) & 4.656e+04 (1.208e+03)\,\(-\) & 3.861e-01 (2.495e-02)\,\(-\) & \textbf{4.467e+04 (9.304e+02)} & \textbf{4.244e-01 (2.158e-02)} \\
  10 & 2.140e+03 (5.073e+01)\,\(-\) & 1.201e-02 (2.057e-02)\,\(-\) & 2.030e+03 (4.734e+01)\,\(-\) & \textbf{3.130e-02 (1.620e-02)}\,\(=\) & \textbf{1.986e+03 (5.092e+01)} & 2.586e-02 (2.388e-02) \\
  \midrule
  11 & \textbf{1.347e+14 (1.852e+13)}\,\(=\) & 5.814e-01 (2.358e-02)\,\(-\) & 4.686e+14 (5.837e+13)\,\(-\) & 4.589e-01 (3.876e-02)\,\(-\) & 1.500e+14 (8.115e+13) & \textbf{9.098e-01 (1.236e-02)} \\
  12 & \textbf{4.515e+10 (1.373e+09)}\,\(+\) & \textbf{7.384e-01 (1.111e-02)}\,\(+\) & 7.766e+10 (5.797e+09)\,\(-\) & 5.119e-01 (4.117e-02)\,\(-\) & 4.718e+10 (1.431e+09) & 7.264e-01 (1.015e-02) \\
  13 & \textbf{6.243e+09 (5.724e+08)}\,\(+\) & \textbf{8.893e-01 (1.193e-02)}\,\(+\) & 2.367e+10 (1.372e+09)\,\(-\) & 5.090e-01 (3.354e-02)\,\(-\) & 9.651e+09 (4.187e+08) & 8.229e-01 (9.250e-03) \\
  14 & \textbf{3.576e+08 (2.773e+07)}\,\(+\) & \textbf{9.620e-01 (2.182e-03)}\,\(+\) & 5.562e+09 (2.063e+08)\,\(-\) & 2.802e-01 (3.030e-02)\,\(-\) & 3.018e+09 (1.902e+08) & 6.204e-01 (2.420e-02) \\
  15 & \textbf{1.808e+09 (1.107e+08)}\,\(+\) & \textbf{9.669e-01 (2.120e-03)}\,\(+\) & 1.930e+10 (9.206e+08)\,\(-\) & 5.034e-01 (3.292e-02)\,\(-\) & 6.264e+09 (3.361e+08) & 8.591e-01 (9.152e-03) \\
  16 & \textbf{1.865e+04 (1.039e+03)}\,\(+\) & \textbf{7.109e-01 (1.096e-02)}\,\(+\) & 3.242e+04 (1.864e+03)\,\(-\) & 4.318e-01 (3.679e-02)\,\(-\) & 2.271e+04 (1.149e+03) & 6.429e-01 (9.608e-03) \\
  17 & 1.188e+10 (2.324e+09)\(=\) & 4.735e-01 (4.057e-02)\,\(-\) & 1.473e+10 (2.768e+09)\,\(-\) & 3.684e-01 (2.997e-02)\,\(-\) & \textbf{1.155e+10 (2.640e+09)} & \textbf{6.222e-01 (1.421e-02)} \\
  18 & \textbf{1.585e+09 (8.652e+07)}\,\(+\) & \textbf{8.494e-01 (8.434e-03)}\,\(+\) & 7.214e+09 (2.910e+08)\,\(-\) & 2.277e-01 (3.344e-02)\,\(-\) & 5.242e+09 (2.341e+08) & 4.901e-01 (1.666e-02) \\
  19 & \textbf{1.845e+09 (1.051e+08)}\,\(+\) & \textbf{9.653e-01 (2.055e-03)}\,\(+\) & 1.975e+10 (1.218e+09)\,\(-\) & 4.780e-01 (4.715e-02)\,\(-\) & 6.253e+09 (2.624e+08) & 8.594e-01 (6.419e-03) \\
  20 & 9.883e+02 (2.714e+01)\,\(-\) & 6.229e-02 (2.175e-02)\,\(-\) & 9.236e+02 (1.929e+01)\,\(-\) & 1.091e-01 (2.099e-02)\,\(-\) & \textbf{9.081e+02 (1.670e+01)} & \textbf{1.229e-01 (2.583e-02)} \\
  \midrule
  21 & 9.065e+02 (8.311e+01)\,\(-\) & 7.419e-01 (1.552e-02)\,\(-\) & 1.322e+03 (1.399e+02)\,\(-\) & 5.321e-01 (4.637e-02)\,\(-\) & \textbf{6.981e+02 (6.715e+01)} & \textbf{8.022e-01 (6.370e-03)} \\
  22 & 2.147e+03 (4.815e+01)\,\(-\) & 7.355e-02 (2.829e-02)\,\(-\) & 2.012e+03 (3.891e+01)\,\(-\) & 1.075e-01 (2.595e-02)\,\(-\) & \textbf{1.978e+03 (3.324e+01)} & \textbf{1.253e-01 (2.595e-02)} \\
  23 & 7.928e+02 (2.868e+01)\,\(-\) & 8.310e-01 (6.198e-03)\,\(-\) & 1.796e+03 (1.417e+02)\,\(-\) & 5.524e-01 (4.036e-02)\,\(-\) & \textbf{6.756e+02 (2.501e+01)} & \textbf{8.520e-01 (5.714e-03)} \\
  24 & \textbf{8.891e+02 (2.119e+01)}\,\(+\) & \textbf{8.742e-01 (6.605e-03)}\,\(+\) & 2.694e+03 (2.191e+02)\,\(-\) & 5.459e-01 (4.500e-02)\,\(-\) & 9.978e+02 (3.671e+01) & 8.545e-01 (4.937e-03) \\
  25 & 6.143e+04 (2.244e+03)\,\(-\) & 5.273e-01 (2.013e-02)\,\(-\) & 6.903e+04 (3.474e+03)\,\(-\) & 4.535e-01 (3.047e-02)\,\(-\) & \textbf{5.879e+04 (2.167e+03)} & \textbf{5.536e-01 (1.606e-02)} \\
  26 & \textbf{3.323e+04 (1.644e+03)}\,\(=\) & \textbf{6.326e-01 (1.471e-02)}\,\(=\) & 4.462e+04 (2.637e+03)\,\(-\) & 4.449e-01 (5.115e-02)\,\(-\) & 3.346e+04 (1.753e+03) & 6.273e-01 (1.222e-02) \\
  27 & \textbf{2.548e+03 (8.719e+01)}\,\(+\) & \textbf{7.288e-01 (1.089e-02)}\,\(+\) & 4.958e+03 (2.112e+02)\,\(-\) & 4.118e-01 (3.434e-02)\,\(-\) & 3.591e+03 (1.190e+02) & 6.090e-01 (1.641e-02) \\
  28 & \textbf{2.027e+04 (5.907e+02)}\,\(+\) & \textbf{5.586e-01 (1.436e-02)}\,\(+\) & 2.582e+04 (1.083e+03)\,\(-\) & 3.956e-01 (3.602e-02)\,\(-\) & 2.068e+04 (5.945e+02) & 5.451e-01 (1.572e-02) \\
  29 & \textbf{2.605e+09 (5.700e+08)}\,\(+\) & 5.146e-01 (3.613e-02)\,\(-\) & 4.490e+09 (7.227e+08)\,\(-\) & 3.291e-01 (3.347e-02)\,\(-\) & 2.954e+09 (6.931e+08) & \textbf{6.815e-01 (1.065e-02)} \\
  30 & \textbf{9.556e+09 (1.208e+09)}\,\(+\) & \textbf{8.807e-01 (1.011e-02)}\,\(+\) & 3.272e+10 (1.364e+09)\,\(-\) & 4.696e-01 (1.896e-02)\,\(-\) & 1.238e+10 (9.811e+08) & 8.369e-01 (7.430e-03) \\
  \midrule
  $(+/-/=)$
  & \textbf{14/12/3}
  & \textbf{13/15/1}
  & \textbf{1/28/0}
  & \textbf{1/27/1}
  & -- & -- \\
  \bottomrule
  \end{tabular}%
  }
  \label{cec2017_100d}%
\end{table}%


\begin{table*}[t]
  \centering
  \caption{Objective performance of TT-DDEA-RBFN, TT-DDEA-TabPFN, and DDEA-TabPFN on CEC2017 for $D=10,30,100$ (25 runs). Each result is in the form of Mean (Standard Deviation). Lower values are better, and the best result for each metric is highlighted. The symbols $+$, $=$, and $-$ appended to the results of TT-DDEA-TabPFN and DDEA-TabPFN indicate that the algorithm performs significantly better than, similarly to, or significantly worse than the baseline TT-DDEA-RBFN, respectively, according to the Wilcoxon rank-sum test at a 0.05 significance level.}
  \setlength{\tabcolsep}{3.5pt} 
  \resizebox{\textwidth}{!}{%
    \begin{tabular}{l c ccc c l c ccc}
    \toprule
    \textbf{F} & $\boldsymbol{D}$ & \textbf{TT-DDEA-RBFN} & \textbf{TT-DDEA-TabPFN} & \textbf{DDEA-TabPFN} & \quad & \textbf{F} & $\boldsymbol{D}$ & \textbf{TT-DDEA-RBFN} & \textbf{TT-DDEA-TabPFN} & \textbf{DDEA-TabPFN} \\
    \midrule
    F1  & 10  & \textbf{1.38e+11 (3.98e+10)} & 2.69e+11 (8.29e+10)~$-$ & 1.87e+11 (4.53e+10)~$-$ & & F17 & 10  & 5.79e+04 (4.42e+04) & \textbf{2.70e+03 (3.18e+02)}~$+$ & 2.90e+03 (3.49e+02)~$+$ \\
        & 30  & \textbf{3.61e+11 (4.91e+10)} & 9.62e+11 (2.14e+11)~$-$ & 5.58e+11 (1.00e+11)~$-$ & &     & 30  & 3.05e+06 (1.02e+07) & \textbf{7.11e+03 (6.30e+03)}~$+$ & 9.22e+03 (9.71e+03)~$+$ \\
        & 100 & \textbf{2.15e+12 (9.49e+10)} & 3.44e+12 (3.87e+11)~$-$ & 2.93e+12 (7.06e+11)~$-$ & &     & 100 & 4.16e+08 (9.25e+08) & 4.29e+07 (1.10e+08)~$=$ & \textbf{4.23e+07 (1.52e+08)}~$+$ \\
    F3  & 10  & 1.17e+12 (2.87e+12) & 5.52e+09 (1.11e+10)~$+$ & \textbf{8.44e+08 (1.82e+09)}~$+$ & & F18 & 10  & 2.39e+10 (4.93e+09) & 2.98e+09 (6.17e+08)~$+$ & \textbf{1.52e+09 (1.04e+09)}~$+$ \\
        & 30  & 1.62e+15 (4.05e+15) & 6.15e+13 (1.18e+14)~$+$ & \textbf{2.44e+13 (6.11e+13)}~$+$ & &     & 30  & 2.57e+10 (1.32e+10) & 6.33e+09 (4.70e+09)~$+$ & \textbf{1.14e+09 (7.17e+08)}~$+$ \\
        & 100 & 1.16e+19 (1.67e+19) & 1.14e+19 (8.07e+18)~$=$ & \textbf{1.67e+18 (3.88e+18)}~$+$ & &     & 100 & 1.16e+10 (5.90e+09) & 1.34e+10 (5.22e+09)~$=$ & \textbf{3.56e+09 (2.28e+09)}~$+$ \\
    F4  & 10  & \textbf{1.42e+03 (6.12e+02)} & 3.59e+03 (2.04e+03)~$-$ & 3.52e+03 (2.43e+03)~$-$ & & F19 & 10  & 1.20e+11 (1.12e+10) & 1.34e+09 (2.88e+09)~$+$ & \textbf{7.92e+08 (2.72e+09)}~$+$ \\
        & 30  & \textbf{1.17e+04 (2.04e+03)} & 9.73e+04 (3.23e+04)~$-$ & 3.37e+04 (1.29e+04)~$-$ & &     & 30  & 6.63e+10 (4.68e+10) & 1.12e+10 (9.78e+09)~$+$ & \textbf{4.72e+09 (2.69e+09)}~$+$ \\
        & 100 & \textbf{5.66e+04 (7.28e+03)} & 1.02e+05 (2.33e+04)~$-$ & 6.58e+04 (2.30e+04)~$=$ & &     & 100 & \textbf{6.74e+10 (2.20e+10)} & 2.01e+11 (4.97e+10)~$-$ & 1.30e+11 (5.13e+10)~$-$ \\
    F5  & 10  & \textbf{6.34e+02 (2.95e+01)} & 6.56e+02 (2.21e+01)~$-$ & 6.53e+02 (1.97e+01)~$-$ & & F20 & 10  & 3.07e+03 (2.78e+02) & 3.10e+03 (3.25e+02)~$=$ & \textbf{2.96e+03 (3.98e+02)}~$=$ \\
        & 30  & \textbf{9.32e+02 (4.30e+01)} & 1.24e+03 (8.10e+01)~$-$ & 9.88e+02 (7.09e+01)~$-$ & &     & 30  & 4.71e+03 (5.22e+02) & \textbf{4.61e+03 (4.61e+02)}~$=$ & 4.69e+03 (5.25e+02)~$=$ \\
        & 100 & \textbf{2.14e+03 (7.36e+01)} & 2.38e+03 (1.33e+02)~$-$ & 2.43e+03 (2.59e+02)~$-$ & &     & 100 & \textbf{1.01e+04 (8.18e+02)} & 1.06e+04 (9.42e+02)~$=$ & 1.08e+04 (1.09e+03)~$-$ \\
    F6  & 10  & \textbf{6.94e+02 (2.04e+01)} & 7.91e+02 (4.75e+01)~$-$ & 7.76e+02 (4.48e+01)~$-$ & & F21 & 10  & 2.53e+03 (7.30e+01) & \textbf{2.46e+03 (1.91e+01)}~$+$ & 2.51e+03 (3.91e+01)~$=$ \\
        & 30  & \textbf{7.20e+02 (1.42e+01)} & 7.80e+02 (2.74e+01)~$-$ & 7.52e+02 (2.39e+01)~$-$ & &     & 30  & 2.85e+03 (7.24e+01) & \textbf{2.83e+03 (5.11e+01)}~$=$ & 2.89e+03 (6.57e+01)~$-$ \\
        & 100 & \textbf{7.32e+02 (6.97e+00)} & 7.48e+02 (1.38e+01)~$-$ & 7.35e+02 (1.54e+01)~$=$ & &     & 100 & 7.75e+03 (1.02e+03) & \textbf{4.23e+03 (1.32e+02)}~$+$ & 4.37e+03 (2.35e+02)~$+$ \\
    F7  & 10  & \textbf{8.39e+02 (3.53e+01)} & 1.41e+03 (2.22e+02)~$-$ & 9.39e+02 (1.29e+02)~$-$ & & F22 & 10  & \textbf{3.64e+03 (4.95e+02)} & 5.51e+03 (1.09e+03)~$-$ & 4.94e+03 (1.49e+03)~$-$ \\
        & 30  & \textbf{1.36e+03 (5.10e+01)} & 3.10e+03 (4.77e+02)~$-$ & 2.27e+03 (2.58e+02)~$-$ & &     & 30  & 1.42e+04 (9.70e+02) & 1.43e+04 (1.19e+03)~$=$ & \textbf{1.38e+04 (1.04e+03)}~$=$ \\
        & 100 & \textbf{3.74e+03 (1.56e+02)} & 7.60e+03 (7.59e+02)~$-$ & 4.37e+03 (1.55e+03)~$=$ & &     & 100 & 4.14e+04 (1.90e+03) & \textbf{4.13e+04 (1.74e+03)}~$=$ & 4.23e+04 (1.71e+03)~$=$ \\
    F8  & 10  & \textbf{9.28e+02 (2.08e+01)} & 9.65e+02 (3.92e+01)~$-$ & 9.43e+02 (3.83e+01)~$=$ & & F23 & 10  & 2.83e+03 (5.82e+01) & 2.81e+03 (3.34e+01)~$=$ & \textbf{2.78e+03 (3.79e+01)}~$+$ \\
        & 30  & \textbf{1.19e+03 (3.37e+01)} & 1.45e+03 (7.68e+01)~$-$ & 1.31e+03 (6.06e+01)~$-$ & &     & 30  & 3.77e+03 (3.25e+02) & 3.43e+03 (1.30e+02)~$+$ & \textbf{3.38e+03 (1.16e+02)}~$+$ \\
        & 100 & \textbf{2.47e+03 (7.44e+01)} & 2.64e+03 (1.30e+02)~$-$ & 3.08e+03 (2.40e+02)~$-$ & &     & 100 & 8.51e+03 (1.34e+02) & \textbf{4.85e+03 (1.27e+02)}~$+$ & 5.94e+03 (2.29e+02)~$+$ \\
    F9  & 10  & \textbf{4.27e+03 (2.18e+03)} & 1.35e+04 (8.17e+03)~$-$ & 9.23e+03 (4.66e+03)~$-$ & & F24 & 10  & 2.88e+03 (3.23e+01) & 2.91e+03 (2.92e+01)~$-$ & \textbf{2.87e+03 (3.66e+01)}~$=$ \\
        & 30  & \textbf{1.41e+04 (2.65e+03)} & 4.09e+04 (1.25e+04)~$-$ & 3.43e+04 (9.21e+03)~$-$ & &     & 30  & 3.34e+03 (9.92e+01) & \textbf{3.24e+03 (6.23e+01)}~$+$ & 3.54e+03 (1.01e+02)~$-$ \\
        & 100 & \textbf{9.50e+04 (1.06e+04)} & 1.37e+05 (2.73e+04)~$-$ & 9.71e+04 (2.46e+04)~$=$ & &     & 100 & 7.19e+03 (5.15e+02) & \textbf{6.10e+03 (1.65e+02)}~$+$ & 9.75e+03 (6.43e+02)~$-$ \\
    F10 & 10  & \textbf{4.80e+03 (6.03e+02)} & 4.83e+03 (6.46e+02)~$=$ & 5.09e+03 (7.45e+02)~$=$ & & F25 & 10  & \textbf{3.88e+03 (2.31e+02)} & 4.09e+03 (6.13e+02)~$=$ & 4.15e+03 (8.71e+02)~$=$ \\
        & 30  & 1.27e+04 (1.35e+03) & 1.28e+04 (1.20e+03)~$=$ & \textbf{1.26e+04 (1.14e+03)}~$=$ & &     & 30  & \textbf{5.60e+03 (5.55e+02)} & 1.97e+04 (9.34e+03)~$-$ & 9.94e+03 (4.79e+03)~$-$ \\
        & 100 & \textbf{3.95e+04 (2.21e+03)} & 4.10e+04 (1.83e+03)~$-$ & 4.04e+04 (1.85e+03)~$=$ & &     & 100 & \textbf{2.19e+04 (1.60e+03)} & 7.09e+04 (1.95e+04)~$-$ & 3.95e+04 (1.23e+04)~$-$ \\
    F11 & 10  & 9.53e+08 (3.45e+08) & 1.08e+05 (3.32e+05)~$+$ & \textbf{9.98e+04 (6.30e+04)}~$+$ & & F26 & 10  & \textbf{5.48e+03 (2.71e+02)} & 6.03e+03 (4.90e+02)~$-$ & 5.78e+03 (5.80e+02)~$=$ \\
        & 30  & 1.61e+11 (6.59e+10) & \textbf{2.51e+05 (5.43e+05)}~$+$ & 2.73e+06 (5.46e+06)~$+$ & &     & 30  & 1.04e+04 (6.76e+02) & 1.13e+04 (8.47e+02)~$-$ & \textbf{1.01e+04 (9.26e+02)}~$=$ \\
        & 100 & 2.43e+15 (8.85e+14) & 2.91e+11 (3.44e+11)~$+$ & \textbf{1.44e+10 (3.70e+10)}~$+$ & &     & 100 & 4.15e+04 (2.99e+03) & \textbf{3.24e+04 (3.18e+03)}~$+$ & 3.75e+04 (5.38e+03)~$+$ \\
    F12 & 10  & 2.35e+10 (1.74e+10) & 2.15e+10 (1.85e+10)~$=$ & \textbf{6.40e+09 (3.31e+09)}~$+$ & & F27 & 10  & 3.92e+03 (5.81e+02) & 3.25e+03 (7.27e+01)~$+$ & \textbf{3.20e+03 (5.70e+01)}~$+$ \\
        & 30  & \textbf{3.74e+10 (1.38e+10)} & 2.69e+11 (4.61e+10)~$-$ & 8.50e+10 (3.94e+10)~$-$ & &     & 30  & 6.76e+03 (1.31e+03) & 6.42e+03 (8.94e+02)~$=$ & \textbf{3.77e+03 (2.31e+02)}~$+$ \\
        & 100 & \textbf{9.39e+11 (1.27e+11)} & 1.98e+12 (2.64e+11)~$-$ & 1.34e+12 (4.24e+11)~$-$ & &     & 100 & 1.39e+04 (1.54e+03) & 1.73e+04 (3.07e+03)~$-$ & \textbf{1.10e+04 (1.64e+03)}~$+$ \\
    F13 & 10  & 3.99e+10 (1.86e+10) & 2.21e+09 (2.37e+09)~$+$ & \textbf{1.07e+08 (1.11e+08)}~$+$ & & F28 & 10  & 4.25e+03 (2.03e+02) & 4.18e+03 (1.88e+02)~$=$ & \textbf{3.95e+03 (1.56e+02)}~$+$ \\
        & 30  & 1.01e+11 (4.22e+10) & 4.35e+11 (8.47e+10)~$-$ & \textbf{6.61e+10 (4.95e+10)}~$+$ & &     & 30  & \textbf{7.05e+03 (4.95e+02)} & 1.17e+04 (1.38e+03)~$-$ & 1.05e+04 (1.66e+03)~$-$ \\
        & 100 & \textbf{2.34e+11 (4.67e+10)} & 3.67e+11 (5.26e+10)~$-$ & 3.30e+11 (9.18e+10)~$-$ & &     & 100 & \textbf{2.81e+04 (1.75e+03)} & 4.56e+04 (5.93e+03)~$-$ & 3.79e+04 (6.89e+03)~$-$ \\
    F14 & 10  & 4.57e+09 (8.88e+08) & 8.71e+05 (2.55e+06)~$+$ & \textbf{4.43e+05 (7.21e+05)}~$+$ & & F29 & 10  & 1.51e+07 (4.56e+07) & 9.22e+04 (1.28e+05)~$+$ & \textbf{8.70e+03 (1.67e+04)}~$+$ \\
        & 30  & 1.85e+10 (3.92e+09) & 2.53e+08 (9.72e+07)~$+$ & \textbf{5.32e+07 (2.69e+07)}~$+$ & &     & 30  & 1.68e+06 (3.78e+06) & 1.35e+06 (3.22e+06)~$=$ & \textbf{1.32e+06 (2.67e+06)}~$=$ \\
        & 100 & 1.57e+10 (8.21e+09) & 3.15e+09 (1.23e+09)~$+$ & \textbf{1.97e+09 (1.67e+09)}~$+$ & &     & 100 & \textbf{2.30e+06 (2.36e+06)} & 2.57e+07 (2.31e+07)~$-$ & 9.34e+06 (1.82e+07)~$=$ \\
    F15 & 10  & 9.39e+10 (5.81e+10) & 1.23e+08 (2.22e+08)~$+$ & \textbf{6.48e+07 (1.22e+08)}~$+$ & & F30 & 10  & 3.52e+10 (3.11e+10) & 1.78e+09 (1.46e+09)~$+$ & \textbf{9.52e+08 (6.40e+08)}~$+$ \\
        & 30  & 1.66e+11 (9.04e+10) & 7.81e+09 (5.49e+09)~$+$ & \textbf{4.80e+09 (3.61e+09)}~$+$ & &     & 30  & 2.23e+10 (1.51e+10) & 2.69e+10 (1.76e+10)~$=$ & \textbf{6.00e+09 (4.93e+09)}~$+$ \\
        & 100 & \textbf{6.72e+10 (1.68e+10)} & 1.69e+11 (3.81e+10)~$-$ & 1.24e+11 (4.72e+10)~$-$ & &     & 100 & 2.24e+11 (2.39e+10) & \textbf{1.83e+11 (3.36e+10)}~$+$ & 1.92e+11 (5.19e+10)~$+$ \\
        F16 & 10  & 1.96e+04 (1.47e+04) & 3.15e+03 (3.33e+02)~$+$ & \textbf{3.03e+03 (3.37e+02)}~$+$ & & \multicolumn{2}{l}{F1--F10} & / & 2 / 22 / 3 & 3 / 16 / 8 \\
        & 30  & 9.22e+03 (2.60e+03) & 1.19e+04 (2.48e+03)~$-$ & \textbf{8.13e+03 (2.86e+03)}~$=$ & & \multicolumn{2}{l}{F11--F20} & / & 16 / 8 / 6 & 20 / 6 / 4 \\
        & 100 & 1.80e+04 (1.60e+03) & 2.01e+04 (1.74e+03)~$-$ & \textbf{1.75e+04 (1.87e+03)}~$=$ & & \multicolumn{2}{l}{F21--F30} & / & 11 / 10 / 9 & 13 / 8 / 9 \\
        &     &                     &                           &                                        & & \multicolumn{2}{l}{$+$ / $-$ / $=$} & / & 29 / 40 / 18 & 36 / 30 / 21 \\
    \bottomrule
    \end{tabular}%
  }
  \label{tab:offline_detail}
\end{table*}

\begin{table*}
  \centering
  \small
  \caption{Objective performance of MGP-SLPSO and TabPFN-SLPSO on CEC2017 for $D=10,30,100$ (25 runs). Each result is in the form of Mean (Standard Deviation). Lower values are better, and the best result for each metric is highlighted. The symbols $+$, $=$, and $-$ appended to the results of TabPFN-SLPSO indicate that the algorithm performs significantly better than, similarly to, or significantly worse than the baseline MGP-SLPSO, respectively, according to the Wilcoxon rank-sum test at a 0.05 significance level. Subtotal and total significance counts ($+ / - / =$) are summarized at the end of each function group and at the bottom of the table.}
  \setlength{\tabcolsep}{2pt}
  \resizebox{\textwidth}{!}{%
    \begin{tabular}{c cc cc cc}
    \toprule
    \multirow{2}{*}{\textbf{F}} & \multicolumn{2}{c}{$\boldsymbol{D=10}$} & \multicolumn{2}{c}{$\boldsymbol{D=30}$} & \multicolumn{2}{c}{$\boldsymbol{D=100}$} \\
    \cmidrule(lr){2-3} \cmidrule(lr){4-5} \cmidrule(lr){6-7}
    & \textbf{MGP-SLPSO} & \textbf{TabPFN-SLPSO} & \textbf{MGP-SLPSO} & \textbf{TabPFN-SLPSO} & \textbf{MGP-SLPSO} & \textbf{TabPFN-SLPSO} \\
    \midrule
    F1 & 3.66e+10 (2.17e+10) & \textbf{3.01e+10 (1.95e+10)}~$=$ & 4.81e+11 (1.09e+11) & \textbf{1.30e+11 (4.95e+10)}~$+$ & 1.01e+12 (2.21e+11) & \textbf{5.72e+11 (1.50e+11)}~$+$ \\
    F3 & 5.93e+04 (3.28e+04) & \textbf{4.57e+04 (2.32e+04)}~$=$ & \textbf{2.82e+05 (2.10e+05)} & 3.35e+05 (2.09e+05)~$-$ & 8.62e+06 (3.63e+07) & \textbf{1.32e+06 (2.55e+06)}~$+$ \\
    F4 & 1.19e+03 (4.11e+02) & \textbf{5.93e+02 (1.02e+02)}~$+$ & 1.04e+04 (3.72e+03) & \textbf{1.75e+03 (5.77e+02)}~$+$ & 2.47e+04 (5.94e+03) & \textbf{9.04e+03 (3.26e+03)}~$+$ \\
    F5 & 6.01e+02 (1.82e+01) & \textbf{5.84e+02 (1.14e+01)}~$+$ & 9.11e+02 (3.86e+01) & \textbf{8.00e+02 (2.34e+01)}~$+$ & 1.84e+03 (5.07e+01) & \textbf{1.72e+03 (1.24e+02)}~$+$ \\
    F6 & 6.75e+02 (1.37e+01) & \textbf{6.57e+02 (1.51e+01)}~$+$ & 7.04e+02 (1.54e+01) & \textbf{6.72e+02 (1.24e+01)}~$+$ & 7.00e+02 (9.71e+00) & \textbf{6.94e+02 (7.56e+00)}~$+$ \\
    F7 & \textbf{8.29e+02 (2.33e+01)} & 8.76e+02 (6.65e+01)~$-$ & 2.04e+03 (1.69e+02) & \textbf{1.27e+03 (1.62e+02)}~$+$ & 6.76e+03 (5.86e+02) & \textbf{3.22e+03 (3.89e+02)}~$+$ \\
    F8 & 8.98e+02 (1.41e+01) & \textbf{8.85e+02 (1.24e+01)}~$+$ & 1.19e+03 (3.94e+01) & \textbf{1.10e+03 (1.99e+01)}~$+$ & 2.17e+03 (5.58e+01) & \textbf{2.06e+03 (7.30e+01)}~$+$ \\
    F9 & 2.51e+03 (5.91e+02) & \textbf{2.23e+03 (6.61e+02)}~$=$ & 1.70e+04 (3.41e+03) & \textbf{9.78e+03 (2.61e+03)}~$+$ & 6.55e+04 (1.30e+04) & \textbf{5.19e+04 (8.21e+03)}~$+$ \\
    F10 & 3.48e+03 (2.50e+02) & \textbf{3.41e+03 (3.67e+02)}~$=$ & \textbf{9.90e+03 (4.73e+02)} & 9.92e+03 (4.58e+02)~$=$ & 3.43e+04 (6.32e+02) & \textbf{3.41e+04 (6.31e+02)}~$=$ \\
    \midrule
    F1--F10 & / & 4 / 1 / 4 (9) & / & 7 / 1 / 1 (9) & / & 8 / 0 / 1 (9) \\
    \midrule
    F11 & 8.65e+03 (8.33e+03) & \textbf{4.09e+03 (2.49e+03)}~$+$ & 2.02e+04 (6.50e+03) & \textbf{1.45e+04 (5.47e+03)}~$+$ & 4.22e+05 (7.38e+04) & \textbf{3.28e+05 (5.41e+04)}~$+$ \\
    F12 & 1.18e+09 (7.41e+08) & \textbf{3.54e+08 (3.55e+08)}~$+$ & 5.46e+10 (1.62e+10) & \textbf{3.71e+09 (4.52e+09)}~$+$ & 2.87e+11 (1.15e+11) & \textbf{1.02e+11 (6.93e+10)}~$+$ \\
    F13 & 3.15e+07 (3.24e+07) & \textbf{6.97e+06 (1.63e+07)}~$+$ & 3.60e+10 (1.59e+10) & \textbf{9.94e+08 (1.57e+09)}~$+$ & 6.02e+10 (2.75e+10) & \textbf{8.81e+09 (9.22e+09)}~$+$ \\
    F14 & 4.20e+04 (7.68e+04) & \textbf{1.47e+04 (1.30e+04)}~$=$ & 1.09e+07 (5.77e+06) & \textbf{3.71e+06 (2.25e+06)}~$+$ & 1.31e+08 (5.91e+07) & \textbf{3.68e+07 (2.16e+07)}~$+$ \\
    F15 & 6.17e+05 (8.93e+05) & \textbf{1.42e+05 (1.44e+05)}~$=$ & 6.32e+09 (2.90e+09) & \textbf{1.40e+07 (1.67e+07)}~$+$ & 3.68e+10 (1.57e+10) & \textbf{1.93e+09 (3.61e+09)}~$+$ \\
    F16 & 2.52e+03 (1.33e+02) & \textbf{2.27e+03 (2.15e+02)}~$+$ & 5.32e+03 (5.30e+02) & \textbf{4.29e+03 (2.61e+02)}~$+$ & 1.26e+04 (5.61e+02) & \textbf{1.16e+04 (5.91e+02)}~$+$ \\
    F17 & 2.12e+03 (1.08e+02) & \textbf{2.00e+03 (9.28e+01)}~$+$ & 4.50e+03 (1.08e+03) & \textbf{3.55e+03 (4.51e+02)}~$+$ & 1.53e+06 (1.59e+06) & \textbf{3.68e+04 (3.41e+04)}~$+$ \\
    F18 & 2.22e+07 (2.81e+07) & \textbf{2.35e+06 (3.24e+06)}~$+$ & 1.69e+08 (1.08e+08) & \textbf{1.09e+07 (9.00e+06)}~$+$ & 2.17e+08 (9.20e+07) & \textbf{3.63e+07 (1.59e+07)}~$+$ \\
    F19 & 3.84e+06 (4.10e+06) & \textbf{1.06e+06 (1.45e+06)}~$+$ & 6.58e+09 (3.71e+09) & \textbf{5.58e+07 (9.52e+07)}~$+$ & 3.23e+10 (1.41e+10) & \textbf{4.78e+09 (8.53e+09)}~$+$ \\
    F20 & 2.46e+03 (9.86e+01) & \textbf{2.33e+03 (9.07e+01)}~$+$ & 3.34e+03 (1.69e+02) & \textbf{3.31e+03 (1.90e+02)}~$=$ & 8.20e+03 (3.76e+02) & \textbf{7.96e+03 (3.48e+02)}~$+$ \\
    \midrule
    F11--F20 & / & 8 / 0 / 2 (10) & / & 9 / 0 / 1 (10) & / & 10 / 0 / 0 (10) \\
    \midrule
    F21 & 2.41e+03 (1.91e+01) & \textbf{2.38e+03 (2.06e+01)}~$+$ & 2.71e+03 (4.07e+01) & \textbf{2.59e+03 (2.75e+01)}~$+$ & 3.92e+03 (9.05e+01) & \textbf{3.64e+03 (7.60e+01)}~$+$ \\
    F22 & 3.05e+03 (2.12e+02) & \textbf{2.94e+03 (2.62e+02)}~$=$ & \textbf{8.76e+03 (2.60e+03)} & 9.65e+03 (2.30e+03)~$=$ & 3.61e+04 (8.02e+02) & \textbf{3.57e+04 (7.39e+02)}~$=$ \\
    F23 & 2.74e+03 (2.67e+01) & \textbf{2.70e+03 (1.47e+01)}~$+$ & 3.22e+03 (8.23e+01) & \textbf{2.97e+03 (2.86e+01)}~$+$ & 4.56e+03 (1.09e+02) & \textbf{4.06e+03 (6.37e+01)}~$+$ \\
    F24 & 2.87e+03 (3.79e+01) & \textbf{2.81e+03 (1.19e+01)}~$+$ & 3.29e+03 (5.86e+01) & \textbf{3.10e+03 (2.07e+01)}~$+$ & 5.16e+03 (2.20e+02) & \textbf{4.52e+03 (6.36e+01)}~$+$ \\
    F25 & 3.49e+03 (2.08e+02) & \textbf{3.17e+03 (1.28e+02)}~$+$ & 7.37e+03 (1.26e+03) & \textbf{3.69e+03 (3.21e+02)}~$+$ & 2.49e+04 (4.87e+03) & \textbf{9.85e+03 (1.50e+03)}~$+$ \\
    F26 & \textbf{4.31e+03 (5.00e+02)} & 4.32e+03 (5.16e+02)~$=$ & 1.01e+04 (9.09e+02) & \textbf{7.35e+03 (2.57e+02)}~$+$ & 2.50e+04 (1.40e+03) & \textbf{2.03e+04 (1.03e+03)}~$+$ \\
    F27 & 3.24e+03 (5.05e+01) & \textbf{3.16e+03 (3.47e+01)}~$+$ & 3.68e+03 (1.36e+02) & \textbf{3.33e+03 (5.49e+01)}~$+$ & 4.42e+03 (1.71e+02) & \textbf{3.89e+03 (1.26e+02)}~$+$ \\
    F28 & 3.88e+03 (1.31e+02) & \textbf{3.70e+03 (1.44e+02)}~$+$ & 6.91e+03 (6.06e+02) & \textbf{4.99e+03 (6.10e+02)}~$+$ & 1.94e+04 (2.10e+03) & \textbf{1.59e+04 (2.52e+03)}~$+$ \\
    F29 & 3.84e+03 (1.50e+02) & \textbf{3.76e+03 (2.29e+02)}~$=$ & 8.50e+03 (2.11e+03) & \textbf{7.09e+03 (1.77e+03)}~$+$ & 4.47e+05 (3.28e+05) & \textbf{3.05e+04 (1.20e+04)}~$+$ \\
    F30 & 1.49e+08 (8.01e+07) & \textbf{6.67e+07 (1.10e+08)}~$+$ & 4.41e+09 (2.53e+09) & \textbf{8.14e+07 (5.53e+07)}~$+$ & 5.94e+10 (2.35e+10) & \textbf{4.85e+09 (7.88e+09)}~$+$ \\
    \midrule
    F21--F30 & / & 7 / 0 / 3 (10) & / & 9 / 0 / 1 (10) & / & 9 / 0 / 1 (10) \\
    \midrule
    Total ($+ / - / =$) & / & 19 / 1 / 9 (29) & / & 25 / 1 / 3 (29) & / & 27 / 0 / 2 (29) \\
    \bottomrule
    \end{tabular}%
  }
  \label{tab:slpso_detail}
\end{table*}

\begin{table}[htbp]
  \centering
  \scriptsize
  \caption{Full-run average surrogate quality (RMSE and $\tau$) on evaluated candidate batches for MGP-SLPSO and TabPFN-SLPSO on CEC 2017 at $D=10,30,100$ (25 runs). Each run first averages the final metric of every GP generation for which a metric could be computed (at least three evaluated candidates), then the 25 run-level values are averaged. Bold indicates the better mean. Symbols $+$, $-$, and $=$ appended to TabPFN-SLPSO indicate significantly better, significantly worse, and no significant difference relative to MGP-SLPSO, respectively (Wilcoxon rank-sum test, $\alpha=0.05$).}
  \label{tab:slpso_rmse_tau}
  \resizebox{\textwidth}{!}{%
    \begin{tabular}{c cccc cccc cccc}
    \toprule
    & \multicolumn{4}{c}{$\boldsymbol{D=10}$} & \multicolumn{4}{c}{$\boldsymbol{D=30}$} & \multicolumn{4}{c}{$\boldsymbol{D=100}$} \\
    \cmidrule(lr){2-5} \cmidrule(lr){6-9} \cmidrule(lr){10-13}
    & \multicolumn{2}{c}{\textbf{MGP}} & \multicolumn{2}{c}{\textbf{TabPFN}} & \multicolumn{2}{c}{\textbf{MGP}} & \multicolumn{2}{c}{\textbf{TabPFN}} & \multicolumn{2}{c}{\textbf{MGP}} & \multicolumn{2}{c}{\textbf{TabPFN}} \\
    \cmidrule(lr){2-3} \cmidrule(lr){4-5} \cmidrule(lr){6-7} \cmidrule(lr){8-9} \cmidrule(lr){10-11} \cmidrule(lr){12-13}
    \textbf{F} & RMSE & $\tau$ & RMSE & $\tau$ & RMSE & $\tau$ & RMSE & $\tau$ & RMSE & $\tau$ & RMSE & $\tau$ \\
    \midrule
    1 & 8.59e+11 & 5.08e-01 & \textbf{5.08e+10}~+ & \textbf{6.91e-01}~+ & 8.54e+11 & 6.64e-02 & \textbf{9.05e+10}~+ & \textbf{6.53e-01}~+ & 4.73e+11 & 3.53e-01 & \textbf{8.47e+10}~+ & \textbf{6.00e-01}~+ \\
    3 & 3.03e+12 & 1.38e-01 & \textbf{1.84e+11}~+ & \textbf{6.03e-01}~+ & 1.05e+15 & 1.80e-01 & \textbf{2.03e+14}~+ & \textbf{3.10e-01}~+ & \textbf{4.65e+17} & \textbf{1.19e-01} & 8.42e+17~- & 7.00e-02~- \\
    4 & 7.42e+04 & 3.75e-02 & \textbf{2.59e+03}~+ & \textbf{6.13e-01}~+ & 7.38e+04 & 1.11e-01 & \textbf{6.56e+03}~+ & \textbf{5.82e-01}~+ & 4.88e+04 & 2.80e-01 & \textbf{6.13e+03}~+ & \textbf{5.78e-01}~+ \\
    5 & 6.35e+02 & 1.27e-01 & \textbf{3.03e+01}~+ & \textbf{4.23e-01}~+ & 2.55e+02 & 1.32e-01 & \textbf{5.43e+01}~+ & \textbf{3.43e-01}~+ & 2.07e+02 & 2.22e-01 & \textbf{7.92e+01}~+ & \textbf{2.82e-01}~= \\
    6 & 4.04e+02 & 5.45e-02 & \textbf{3.69e+01}~+ & \textbf{4.45e-01}~+ & 6.68e+01 & 1.25e-01 & \textbf{1.77e+01}~+ & \textbf{3.24e-01}~+ & 1.97e+01 & 2.07e-01 & \textbf{8.45e+00}~+ & \textbf{2.49e-01}~+ \\
    7 & 3.17e+03 & -1.61e-02 & \textbf{1.08e+02}~+ & \textbf{5.82e-01}~+ & 1.98e+03 & -8.66e-02 & \textbf{1.95e+02}~+ & \textbf{5.16e-01}~+ & 1.60e+03 & 9.37e-02 & \textbf{1.84e+02}~+ & \textbf{5.16e-01}~+ \\
    8 & 5.65e+02 & 9.09e-02 & \textbf{3.10e+01}~+ & \textbf{4.23e-01}~+ & 2.43e+02 & 7.09e-02 & \textbf{5.14e+01}~+ & \textbf{3.28e-01}~+ & 2.08e+02 & \textbf{2.59e-01} & \textbf{8.47e+01}~+ & 2.53e-01~= \\
    9 & 6.50e+04 & -1.55e-02 & \textbf{3.90e+03}~+ & \textbf{4.78e-01}~+ & 3.32e+04 & 7.10e-02 & \textbf{6.91e+03}~+ & \textbf{3.18e-01}~+ & 3.63e+04 & 1.95e-01 & \textbf{1.30e+04}~+ & \textbf{2.58e-01}~+ \\
    10 & 5.62e+03 & 1.01e-02 & \textbf{6.23e+02}~+ & \textbf{7.77e-02}~+ & 2.61e+03 & 2.94e-02 & \textbf{1.05e+03}~+ & \textbf{3.69e-02}~= & 3.94e+03 & 6.42e-03 & \textbf{1.89e+03}~+ & \textbf{7.27e-03}~= \\
    11 & 8.53e+09 & 8.58e-02 & \textbf{2.54e+07}~+ & \textbf{6.39e-01}~+ & 3.53e+10 & 8.25e-02 & \textbf{2.26e+09}~+ & \textbf{5.38e-01}~+ & 6.48e+13 & 6.81e-02 & \textbf{6.69e+12}~+ & \textbf{2.47e-01}~+ \\
    12 & 2.39e+11 & 4.90e-01 & \textbf{7.83e+09}~+ & \textbf{7.52e-01}~+ & 2.08e+11 & 1.72e-01 & \textbf{1.58e+10}~+ & \textbf{6.83e-01}~+ & 3.19e+11 & 3.70e-01 & \textbf{4.45e+10}~+ & \textbf{6.12e-01}~+ \\
    13 & 1.44e+11 & 5.50e-01 & \textbf{1.91e+09}~+ & \textbf{8.36e-01}~+ & 3.77e+11 & 1.89e-01 & \textbf{2.72e+10}~+ & \textbf{6.84e-01}~+ & 1.05e+11 & 3.36e-01 & \textbf{1.16e+10}~+ & \textbf{5.96e-01}~+ \\
    14 & 1.36e+10 & 4.95e-01 & \textbf{3.50e+07}~+ & \textbf{8.35e-01}~+ & 3.45e+09 & 1.35e-01 & \textbf{2.24e+08}~+ & \textbf{5.86e-01}~+ & 1.77e+09 & 1.71e-01 & \textbf{2.27e+08}~+ & \textbf{4.38e-01}~+ \\
    15 & 2.68e+11 & 5.79e-01 & \textbf{1.23e+09}~+ & \textbf{7.95e-01}~+ & 1.87e+11 & 2.07e-01 & \textbf{1.15e+10}~+ & \textbf{7.12e-01}~+ & 8.20e+10 & 2.93e-01 & \textbf{6.73e+09}~+ & \textbf{5.49e-01}~+ \\
    16 & 1.06e+05 & 6.13e-02 & \textbf{1.59e+03}~+ & \textbf{4.99e-01}~+ & 4.16e+04 & 1.25e-01 & \textbf{2.21e+03}~+ & \textbf{2.98e-01}~+ & 7.84e+03 & \textbf{1.94e-01} & \textbf{1.58e+03}~+ & 1.78e-01~= \\
    17 & 5.80e+06 & 7.17e-02 & \textbf{3.88e+04}~+ & \textbf{4.49e-01}~+ & 1.00e+08 & 3.96e-02 & \textbf{3.50e+06}~+ & \textbf{2.57e-01}~+ & 2.20e+09 & 6.39e-02 & \textbf{9.37e+07}~+ & \textbf{2.07e-01}~+ \\
    18 & 6.11e+10 & 5.02e-01 & \textbf{1.02e+09}~+ & \textbf{7.43e-01}~+ & 1.18e+10 & 1.78e-01 & \textbf{1.86e+09}~+ & \textbf{5.53e-01}~+ & 2.30e+09 & 1.95e-01 & \textbf{3.79e+08}~+ & \textbf{3.83e-01}~+ \\
    19 & 5.53e+11 & 2.68e-01 & \textbf{1.75e+09}~+ & \textbf{8.35e-01}~+ & 2.05e+11 & 1.88e-01 & \textbf{1.21e+10}~+ & \textbf{6.59e-01}~+ & 8.17e+10 & 2.98e-01 & \textbf{7.04e+09}~+ & \textbf{5.47e-01}~+ \\
    20 & 3.94e+03 & 4.19e-02 & \textbf{3.03e+02}~+ & \textbf{2.28e-01}~+ & 1.14e+03 & \textbf{3.16e-02} & \textbf{4.87e+02}~+ & 1.78e-02~= & 1.77e+03 & \textbf{2.16e-02} & \textbf{8.53e+02}~+ & 1.50e-02~= \\
    21 & 3.43e+03 & 1.32e-01 & \textbf{1.51e+02}~+ & \textbf{3.31e-01}~+ & 5.93e+02 & 7.82e-02 & \textbf{7.33e+01}~+ & \textbf{3.42e-01}~+ & 4.32e+02 & 2.00e-01 & \textbf{1.09e+02}~+ & \textbf{2.48e-01}~+ \\
    22 & 6.19e+03 & 1.24e-01 & \textbf{8.49e+02}~+ & \textbf{4.74e-01}~+ & 2.48e+03 & 1.16e-01 & \textbf{1.07e+03}~+ & \textbf{1.29e-01}~= & 3.96e+03 & 6.62e-03 & \textbf{1.89e+03}~+ & \textbf{1.96e-02}~= \\
    23 & 1.68e+04 & 1.75e-01 & \textbf{9.83e+01}~+ & \textbf{4.89e-01}~+ & 1.19e+03 & 2.47e-01 & \textbf{9.82e+01}~+ & \textbf{3.76e-01}~+ & 6.52e+02 & 2.20e-01 & \textbf{1.05e+02}~+ & \textbf{2.32e-01}~= \\
    24 & 3.39e+03 & 1.48e-01 & \textbf{1.26e+02}~+ & \textbf{4.77e-01}~+ & 8.87e+02 & \textbf{3.05e-01} & \textbf{8.51e+01}~+ & 3.05e-01~= & 9.10e+02 & 2.83e-01 & \textbf{1.39e+02}~+ & \textbf{2.98e-01}~= \\
    25 & 3.87e+04 & 1.15e-01 & \textbf{1.30e+03}~+ & \textbf{6.52e-01}~+ & 3.44e+04 & 2.94e-02 & \textbf{2.79e+03}~+ & \textbf{5.87e-01}~+ & 2.63e+04 & 2.17e-01 & \textbf{2.77e+03}~+ & \textbf{5.86e-01}~+ \\
    26 & 1.38e+04 & 7.19e-02 & \textbf{8.10e+02}~+ & \textbf{5.06e-01}~+ & 1.34e+04 & 1.37e-01 & \textbf{1.37e+03}~+ & \textbf{4.13e-01}~+ & 1.02e+04 & 2.82e-01 & \textbf{1.71e+03}~+ & \textbf{3.09e-01}~+ \\
    27 & 1.19e+04 & 2.08e-01 & \textbf{4.99e+02}~+ & \textbf{7.02e-01}~+ & 2.26e+03 & 3.12e-01 & \textbf{2.67e+02}~+ & \textbf{6.35e-01}~+ & 1.37e+03 & 3.97e-01 & \textbf{2.95e+02}~+ & \textbf{5.33e-01}~+ \\
    28 & 1.00e+04 & 7.36e-02 & \textbf{4.83e+02}~+ & \textbf{5.72e-01}~+ & 1.24e+04 & 1.47e-01 & \textbf{1.40e+03}~+ & \textbf{6.05e-01}~+ & 7.03e+03 & 3.61e-01 & \textbf{1.38e+03}~+ & \textbf{5.84e-01}~+ \\
    29 & 1.19e+08 & 7.84e-02 & \textbf{4.65e+06}~+ & \textbf{3.26e-01}~+ & 4.34e+08 & 4.13e-02 & \textbf{2.21e+07}~+ & \textbf{2.31e-01}~+ & 5.64e+08 & 5.76e-02 & \textbf{2.42e+07}~+ & \textbf{2.09e-01}~+ \\
    30 & 1.18e+11 & 1.92e-01 & \textbf{6.10e+09}~+ & \textbf{6.67e-01}~+ & 1.10e+11 & 1.37e-01 & \textbf{5.23e+09}~+ & \textbf{6.12e-01}~+ & 1.26e+11 & 2.47e-01 & \textbf{8.87e+09}~+ & \textbf{5.33e-01}~+ \\
    \midrule
    \textbf{RMSE Total} & \multicolumn{2}{c}{/} & \multicolumn{2}{c}{29/0/0} & \multicolumn{2}{c}{/} & \multicolumn{2}{c}{29/0/0} & \multicolumn{2}{c}{/} & \multicolumn{2}{c}{28/1/0} \\
    \textbf{$\tau$ Total} & \multicolumn{2}{c}{/} & \multicolumn{2}{c}{29/0/0} & \multicolumn{2}{c}{/} & \multicolumn{2}{c}{25/0/4} & \multicolumn{2}{c}{/} & \multicolumn{2}{c}{20/1/8} \\
    \bottomrule
    \end{tabular}%
  }
\end{table}

\begin{table*}[!t]
	\centering
    \caption{Detailed final best objective values obtained by SHEALED and SHEALED-TabPFN on the Type 1 and Type 2 mixed-variable benchmark problems for $D=10,30,50$ (25 independent runs). The better mean result is highlighted in bold. The symbols $+$, $-$, and $=$ indicate that SHEALED-TabPFN performs significantly better than, significantly worse than, and shows no significant difference from SHEALED, respectively, according to the Wilcoxon rank-sum test at a significance level of 0.05.}
	\label{tab:supp_mixed_optimization}

	\begin{tabular}{@{}llccc@{}}
		\toprule
		\textbf{Type} & \textbf{Problem} & \(\boldsymbol{(n_1,n_2)}\)
		& \textbf{SHEALED} & \textbf{SHEALED-TabPFN} \\
		\midrule

		\multirow{15}{*}{Type 1}
		& \multirow{3}{*}{Ellipsoid}
		& $(5,5)$
		& \textbf{2.41e-25 (1.19e-24)}
		& 5.30e+02 (4.97e+02) $-$ \\
		&& $(15,15)$
		& \textbf{4.59e+02 (5.28e+02)}
		& 3.82e+03 (3.71e+03) $-$ \\
		&& $(25,25)$
		& \textbf{1.21e+04 (4.53e+03)}
		& 2.36e+04 (1.25e+04) $-$ \\

		& \multirow{3}{*}{Rosenbrock}
		& $(5,5)$
		& \textbf{1.38e+03 (3.13e+03)}
		& 1.15e+05 (3.89e+05) $=$ \\
		&& $(15,15)$
		& 1.09e+07 (7.75e+06)
		& \textbf{3.69e+06 (6.76e+06)} $+$ \\
		&& $(25,25)$
		& 1.80e+08 (1.00e+08)
		& \textbf{2.54e+07 (1.89e+07)} $+$ \\

		& \multirow{3}{*}{Ackley}
		& $(5,5)$
		& \textbf{1.90e+00 (1.65e+00)}
		& 2.69e+00 (1.98e+00) $=$ \\
		&& $(15,15)$
		& \textbf{3.94e+00 (1.56e+00)}
		& 5.27e+00 (1.75e+00) $-$ \\
		&& $(25,25)$
		& 6.52e+00 (1.36e+00)
		& \textbf{6.15e+00 (1.72e+00)} $=$ \\

		& \multirow{3}{*}{Griewank}
		& $(5,5)$
		& \textbf{5.99e-02 (6.04e-02)}
		& 7.30e-01 (3.77e-01) $-$ \\
		&& $(15,15)$
		& \textbf{1.17e+00 (6.45e-02)}
		& 2.40e+00 (1.10e+00) $-$ \\
		&& $(25,25)$
		& 7.92e+00 (1.67e+00)
		& \textbf{7.14e+00 (4.51e+00)} $+$ \\

		& \multirow{3}{*}{Rastrigin}
		& $(5,5)$
		& \textbf{1.79e+00 (1.22e+00)}
		& 1.69e+01 (6.12e+00) $-$ \\
		&& $(15,15)$
		& \textbf{1.80e+01 (4.66e+00)}
		& 5.64e+01 (1.17e+01) $-$ \\
		&& $(25,25)$
		& \textbf{5.96e+01 (1.37e+01)}
		& 1.04e+02 (2.53e+01) $-$ \\

		\midrule

		\multirow{15}{*}{Type 2}
		& \multirow{3}{*}{Ellipsoid}
		& $(5,5)$
		& \textbf{2.84e-24 (1.17e-23)}
		& 3.71e+03 (5.11e+03) $-$ \\
		&& $(15,15)$
		& \textbf{2.07e+04 (1.24e+04)}
		& 1.01e+05 (6.28e+04) $-$ \\
		&& $(25,25)$
		& \textbf{2.13e+05 (9.54e+04)}
		& 2.50e+05 (9.78e+04) $=$ \\

		& \multirow{3}{*}{Rosenbrock}
		& $(5,5)$
		& \textbf{2.21e+02 (1.90e+02)}
		& 1.90e+08 (2.18e+08) $-$ \\
		&& $(15,15)$
		& \textbf{8.89e+07 (7.67e+07)}
		& 1.36e+09 (1.56e+09) $-$ \\
		&& $(25,25)$
		& \textbf{1.41e+09 (5.98e+08)}
		& 4.58e+09 (3.33e+09) $-$ \\

		& \multirow{3}{*}{Ackley}
		& $(5,5)$
		& \textbf{1.78e+00 (2.74e+00)}
		& 6.48e+00 (3.92e+00) $-$ \\
		&& $(15,15)$
		& \textbf{8.71e+00 (2.77e+00)}
		& 1.14e+01 (2.26e+00) $-$ \\
		&& $(25,25)$
		& \textbf{1.34e+01 (1.20e+00)}
		& 1.35e+01 (2.05e+00) $=$ \\

		& \multirow{3}{*}{Griewank}
		& $(5,5)$
		& \textbf{8.77e-02 (5.71e-02)}
		& 4.09e+00 (6.69e+00) $-$ \\
		&& $(15,15)$
		& \textbf{2.70e+00 (2.28e+00)}
		& 3.58e+01 (3.17e+01) $-$ \\
		&& $(25,25)$
		& 1.04e+02 (6.65e+01)
		& \textbf{1.01e+02 (5.43e+01)} $=$ \\

		& \multirow{3}{*}{Rastrigin}
		& $(5,5)$
		& \textbf{1.74e+00 (1.30e+00)}
		& 1.79e+01 (8.21e+00) $-$ \\
		&& $(15,15)$
		& \textbf{3.49e+01 (1.56e+01)}
		& 8.02e+01 (2.70e+01) $-$ \\
		&& $(25,25)$
		& \textbf{1.19e+02 (2.62e+01)}
		& 1.71e+02 (3.67e+01) $-$ \\

		\midrule
		\multicolumn{3}{l}{$+/-/=$}
		& /
		& 3/21/6 \\
		\bottomrule
	\end{tabular}
\end{table*}

\begin{table*}[!t]
	\centering
	\small
	\caption{Detailed predictive performance of RBFNmv and TabPFN on the Type 1 and Type 2 mixed-variable benchmark problems for $D=10,30,50$ (25 independent runs), with $N_{\mathrm{tr}}=10D$ and $N_{\mathrm{te}}=2000$. Results are reported as mean (standard deviation). Lower RMSE values and higher $R^2$ and Kendall's $\tau$ values indicate better predictive performance, and the better mean result is highlighted in bold. The symbols $+$, $-$, and $=$ indicate that TabPFN performs significantly better than, significantly worse than, and shows no significant difference from RBFNmv, respectively, according to the Wilcoxon rank-sum test at a significance level of 0.05.}

	\label{tab:supp_mixed_prediction}

	\setlength{\tabcolsep}{2pt}
	\resizebox{\textwidth}{!}{%
	\begin{tabular}{@{}lllcccccc@{}}
		\toprule
        \multirow{2}{*}{\textbf{Type}} &
		\multirow{2}{*}{\textbf{Problem}} &
		\multirow{2}{*}{\(\boldsymbol{(n_1,n_2)}\)}
		& \multicolumn{2}{c}{\textbf{RMSE}}
		& \multicolumn{2}{c}{\(\boldsymbol{R^2}\)}
		& \multicolumn{2}{c}{\(\boldsymbol{\tau}\)} \\
		\cmidrule(lr){4-5}
		\cmidrule(lr){6-7}
		\cmidrule(lr){8-9}
		& & &
		\textbf{RBFNmv} & \textbf{TabPFN} &
		\textbf{RBFNmv} & \textbf{TabPFN} &
		\textbf{RBFNmv} & \textbf{TabPFN} \\
		\midrule

		\multirow{15}{*}{Type 1}
		& \multirow{3}{*}{Ellipsoid}
		& $(5,5)$
		& \textbf{4.9578e+03(7.7173e+02)}
		& 5.4019e+03(5.2081e+02) $=$
		& \textbf{9.8708e-01(4.1295e-03)}
		& 9.8485e-01(3.1160e-03) $=$
		& \textbf{9.2986e-01(1.1117e-02)}
		& 9.2430e-01(7.6053e-03) $=$ \\

		&& $(15,15)$
		& \textbf{1.8923e+04(2.3611e+03)}
		& 3.3882e+04(3.0328e+03) $-$
		& \textbf{9.8516e-01(3.7428e-03)}
		& 9.5286e-01(8.4829e-03) $-$
		& \textbf{9.2382e-01(1.0037e-02)}
		& 8.6449e-01(1.2924e-02) $-$ \\

		&& $(25,25)$
		& \textbf{4.7236e+04(3.0419e+03)}
		& 9.8022e+04(4.9759e+03) $-$
		& \textbf{9.8592e-01(1.8744e-03)}
		& 9.3943e-01(7.1071e-03) $-$
		& \textbf{9.2533e-01(5.2241e-03)}
		& 8.4501e-01(9.5025e-03) $-$ \\

		\cmidrule(lr){2-9}

		& \multirow{3}{*}{Rosenbrock}
		& $(5,5)$
		& 8.3429e+09(1.2520e+09)
		& \textbf{4.0947e+09(7.3409e+08)} $+$
		& 9.6541e-01(1.0120e-02)
		& \textbf{9.9157e-01(3.1256e-03)} $+$
		& 8.7933e-01(1.7824e-02)
		& \textbf{9.5394e-01(7.7969e-03)} $+$ \\

		&& $(15,15)$
		& 1.2267e+10(6.8406e+08)
		& \textbf{1.1036e+10(1.1188e+09)} $+$
		& 9.5600e-01(5.0788e-03)
		& \textbf{9.6410e-01(7.3208e-03)} $+$
		& 8.6671e-01(8.1587e-03)
		& \textbf{8.8331e-01(1.2913e-02)} $+$ \\

		&& $(25,25)$
		& 1.5245e+10(1.4375e+09)
		& \textbf{1.4573e+10(8.3915e+08)} $=$
		& 9.5528e-01(8.0601e-03)
		& \textbf{9.5923e-01(5.2093e-03)} $=$
		& 8.6585e-01(1.1926e-02)
		& \textbf{8.7321e-01(7.9169e-03)} $=$ \\

		\cmidrule(lr){2-9}

		& \multirow{3}{*}{Ackley}
		& $(5,5)$
		& 9.5195e-01(1.0302e-01)
		& \textbf{5.6686e-01(8.6244e-02)} $+$
		& 3.6914e-01(1.3397e-01)
		& \textbf{7.7276e-01(7.1230e-02)} $+$
		& 5.7429e-01(3.8531e-02)
		& \textbf{7.1859e-01(2.5134e-02)} $+$ \\

		&& $(15,15)$
		& 2.8449e-01(1.4695e-02)
		& \textbf{2.1884e-01(1.4391e-02)} $+$
		& 5.9969e-01(3.6350e-02)
		& \textbf{7.6328e-01(2.3826e-02)} $+$
		& 6.4292e-01(2.0578e-02)
		& \textbf{6.8271e-01(1.3338e-02)} $+$ \\

		&& $(25,25)$
		& 1.7055e-01(5.1200e-03)
		& \textbf{1.4742e-01(5.8238e-03)} $+$
		& 6.7281e-01(2.3393e-02)
		& \textbf{7.5594e-01(1.3353e-02)} $+$
		& 6.6538e-01(1.3419e-02)
		& \textbf{6.6625e-01(9.6244e-03)} $=$ \\

		\cmidrule(lr){2-9}

		& \multirow{3}{*}{Griewank}
		& $(5,5)$
		& \textbf{1.0823e+01(1.9338e+00)}
		& 1.1592e+01(2.5295e+00) $=$
		& \textbf{9.9244e-01(2.6635e-03)}
		& 9.9121e-01(3.8084e-03) $=$
		& 9.4667e-01(9.7754e-03)
		& \textbf{9.4867e-01(1.2289e-02)} $=$ \\

		&& $(15,15)$
		& \textbf{1.6714e+01(1.7759e+00)}
		& 4.3047e+01(4.1200e+00) $-$
		& \textbf{9.9285e-01(1.5770e-03)}
		& 9.5280e-01(8.7046e-03) $-$
		& \textbf{9.4664e-01(6.2774e-03)}
		& 8.6752e-01(1.3364e-02) $-$ \\

		&& $(25,25)$
		& \textbf{2.0885e+01(1.8589e+00)}
		& 7.3422e+01(3.7867e+00) $-$
		& \textbf{9.9340e-01(1.1385e-03)}
		& 9.1881e-01(8.1965e-03) $-$
		& \textbf{9.4861e-01(4.3817e-03)}
		& 8.2094e-01(1.0445e-02) $-$ \\

		\cmidrule(lr){2-9}

		& \multirow{3}{*}{Rastrigin}
		& $(5,5)$
		& \textbf{2.1795e+01(1.4976e+00)}
		& 2.4049e+01(1.0625e+00) $-$
		& \textbf{7.0650e-01(3.9955e-02)}
		& 6.4362e-01(3.0417e-02) $-$
		& \textbf{6.5167e-01(2.2799e-02)}
		& 6.0269e-01(2.4432e-02) $-$ \\

		&& $(15,15)$
		& \textbf{3.8261e+01(1.4195e+00)}
		& 4.2135e+01(1.2046e+00) $-$
		& \textbf{7.1405e-01(2.1043e-02)}
		& 6.5344e-01(1.9078e-02) $-$
		& \textbf{6.5464e-01(1.4028e-02)}
		& 6.0533e-01(1.3870e-02) $-$ \\

		&& $(25,25)$
		& \textbf{4.8505e+01(1.5939e+00)}
		& 5.3637e+01(1.5932e+00) $-$
		& \textbf{7.2219e-01(1.7740e-02)}
		& 6.6027e-01(2.1430e-02) $-$
		& \textbf{6.5836e-01(1.3077e-02)}
		& 6.0653e-01(1.5014e-02) $-$ \\

		\midrule

		\multirow{15}{*}{Type 2}
		& \multirow{3}{*}{Ellipsoid}
		& $(5,5)$
		& \textbf{6.5384e+03(1.0527e+03)}
		& 2.9146e+04(3.1467e+03) $-$
		& \textbf{9.9732e-01(8.7124e-04)}
		& 9.4734e-01(1.1491e-02) $-$
		& \textbf{9.6842e-01(5.2174e-03)}
		& 8.6278e-01(1.4872e-02) $-$ \\

		&& $(15,15)$
		& \textbf{3.6694e+04(6.7934e+03)}
		& 2.3571e+05(2.0548e+04) $-$
		& \textbf{9.9735e-01(1.0592e-03)}
		& 8.9340e-01(1.8788e-02) $-$
		& \textbf{9.6802e-01(6.0101e-03)}
		& 7.9803e-01(2.0624e-02) $-$ \\

		&& $(25,25)$
		& \textbf{7.8380e+04(9.3617e+03)}
		& 5.2451e+05(2.6426e+04) $-$
		& \textbf{9.9674e-01(8.5327e-04)}
		& 8.5586e-01(1.3842e-02) $-$
		& \textbf{9.6403e-01(4.3960e-03)}
		& 7.6169e-01(1.2592e-02) $-$ \\

		\cmidrule(lr){2-9}

		& \multirow{3}{*}{Rosenbrock}
		& $(5,5)$
		& \textbf{7.3078e+09(1.1018e+09)}
		& 8.6203e+09(1.2172e+09) $-$
		& \textbf{9.8106e-01(5.7682e-03)}
		& 9.7377e-01(7.2081e-03) $-$
		& \textbf{9.1269e-01(1.3667e-02)}
		& 8.9847e-01(1.3854e-02) $-$ \\

		&& $(15,15)$
		& \textbf{1.3038e+10(1.0598e+09)}
		& 2.4247e+10(1.7538e+09) $-$
		& \textbf{9.7522e-01(4.1958e-03)}
		& 9.1440e-01(1.2795e-02) $-$
		& \textbf{9.0068e-01(8.2205e-03)}
		& 8.1518e-01(1.4382e-02) $-$ \\

		&& $(25,25)$
		& \textbf{1.8445e+10(1.2721e+09)}
		& 3.5023e+10(2.0680e+09) $-$
		& \textbf{9.7413e-01(3.7253e-03)}
		& 9.0688e-01(1.1292e-02) $-$
		& \textbf{8.9781e-01(7.2092e-03)}
		& 8.0814e-01(1.2315e-02) $-$ \\

		\cmidrule(lr){2-9}

		& \multirow{3}{*}{Ackley}
		& $(5,5)$
		& 4.7158e-01(7.1614e-02)
		& \textbf{4.1967e-01(2.7815e-02)} $+$
		& 2.3565e-01(2.2316e-01)
		& \textbf{4.0098e-01(7.3269e-02)} $+$
		& 4.8336e-01(4.7083e-02)
		& \textbf{4.8793e-01(4.3466e-02)} $=$ \\

		&& $(15,15)$
		& 1.7059e-01(7.4321e-03)
		& \textbf{1.6551e-01(7.8835e-03)} $=$
		& 4.2642e-01(4.9133e-02)
		& \textbf{4.6047e-01(4.3384e-02)} $=$
		& \textbf{5.1132e-01(1.9314e-02)}
		& 4.8856e-01(2.4914e-02) $-$ \\

		&& $(25,25)$
		& \textbf{1.2167e-01(3.7607e-03)}
		& 1.3115e-01(3.1568e-03) $-$
		& \textbf{5.2038e-01(3.2050e-02)}
		& 4.4306e-01(2.8226e-02) $-$
		& \textbf{5.5265e-01(1.1288e-02)}
		& 4.7219e-01(1.8204e-02) $-$ \\

		\cmidrule(lr){2-9}

		& \multirow{3}{*}{Griewank}
		& $(5,5)$
		& \textbf{1.2381e+01(2.2748e+00)}
		& 5.3441e+01(5.4156e+00) $-$
		& \textbf{9.9525e-01(1.7572e-03)}
		& 9.1347e-01(1.7978e-02) $-$
		& \textbf{9.5774e-01(7.8998e-03)}
		& 8.1692e-01(1.9918e-02) $-$ \\

		&& $(15,15)$
		& \textbf{2.5080e+01(3.4109e+00)}
		& 1.5516e+02(1.0309e+01) $-$
		& \textbf{9.9570e-01(1.1476e-03)}
		& 8.3769e-01(1.9390e-02) $-$
		& \textbf{9.5888e-01(5.5326e-03)}
		& 7.4906e-01(1.3918e-02) $-$ \\

		&& $(25,25)$
		& \textbf{3.4425e+01(3.2721e+00)}
		& 2.0343e+02(1.1044e+01) $-$
		& \textbf{9.9492e-01(1.0007e-03)}
		& 8.2384e-01(1.8296e-02) $-$
		& \textbf{9.5499e-01(4.3108e-03)}
		& 7.3205e-01(1.5426e-02) $-$ \\

		\cmidrule(lr){2-9}

		& \multirow{3}{*}{Rastrigin}
		& $(5,5)$
		& \textbf{2.1532e+01(1.1940e+00)}
		& 2.7507e+01(2.1175e+00) $-$
		& \textbf{8.3853e-01(1.8463e-02)}
		& 7.3590e-01(3.9942e-02) $-$
		& \textbf{7.4079e-01(1.4435e-02)}
		& 6.7596e-01(2.6303e-02) $-$ \\

		&& $(15,15)$
		& \textbf{3.7664e+01(1.1423e+00)}
		& 5.3804e+01(3.6199e+00) $-$
		& \textbf{8.7197e-01(7.9722e-03)}
		& 7.3785e-01(3.5837e-02) $-$
		& \textbf{7.7104e-01(7.2267e-03)}
		& 6.7030e-01(2.0860e-02) $-$ \\

		&& $(25,25)$
		& \textbf{4.8684e+01(1.6833e+00)}
		& 7.2900e+01(2.6020e+00) $-$
		& \textbf{8.7217e-01(8.1096e-03)}
		& 7.1328e-01(2.0311e-02) $-$
		& \textbf{7.7065e-01(7.5218e-03)}
		& 6.4716e-01(1.4932e-02) $-$ \\

		\midrule
		\multicolumn{3}{l}{$+/-/=$}
		& / & 6/20/4
		& / & 6/20/4
		& / & 4/21/5 \\
		\bottomrule
	\end{tabular}%
	}
\end{table*}

\begin{table}[!t]
\centering
\caption{Mean (standard deviation) task performance obtained by AIEA and TabPFNEA on 30 tasks under online and offline settings. The best results are highlighted in bold. Each algorithm is independently run five times.}
\label{tab:aiea}
\setlength{\tabcolsep}{1pt}
\begin{tabular}{l|c|cc|cc}
\toprule
\multirow{2}{*}{\textbf{Tasks}}
& \multirow{2}{*}{\textbf{Difficulty}}
& \multicolumn{2}{c|}{\textbf{Online}}
& \multicolumn{2}{c}{\textbf{Offline}} \\
\cmidrule(lr){3-4} \cmidrule(lr){5-6}
&
& \textbf{AIEA} & \textbf{TabPFNEA}
& \textbf{AIEA} & \textbf{TabPFNEA} \\
\midrule
BridgeWalker-v0
& \multirow{10}{*}{Easy}
& 5.69(0.63) & \textbf{6.56(0.01)}$=$
& \textbf{5.96(0.59)} & 4.28(0.24)$-$ \\

Pusher-v0
&
& 8.68(0.73) & \textbf{9.59(0.42)}$=$
& 7.16(2.06) & \textbf{8.44(0.17)}$+$ \\

BeamToppler-v0
&
& 3.20(0.32) & \textbf{6.97(2.92)}$-$
& \textbf{6.87(4.36)} & 2.75(0.09)$-$ \\

Carrier-v0
&
& 7.32(1.01) & \textbf{8.02(1.16)}$=$
& 3.22(3.28) & \textbf{6.09(0.89)}$+$ \\

DownStepper-v0
&
& 8.26(1.14) & \textbf{9.06(0.01)}$=$
& 4.97(0.83) & \textbf{7.74(1.31)}$+$ \\

AreaMaximizer-v0
&
& \textbf{2.21(0.35)} & 2.10(0.17)$=$
& 1.81(0.12) & \textbf{1.82(0.50)}$=$ \\

WingspanMax-v0
&
& \textbf{0.76(0.13)} & 0.55(0.12)$=$
& \textbf{0.54(0.14)} & 0.45(0.02)$=$ \\

Flipper-v0
&
& \textbf{12.94(11.29)} & 7.61(3.1)$-$
& \textbf{4.86(0.21)} & 4.18(0.72)$=$\\

Jumper-v0
&
& 1.61(0.27) & \textbf{2.24(0.41)}$+$
& 2.03(0.77) & 2.03(0.78)$=$ \\

Balancer-v0
&
& \textbf{0.12(0.02)} & 0.09(0.01)$=$
& -0.2(0.19) & \textbf{0.10(0.02)}$+$ \\

\hline

BidrectWalker-v0
& \multirow{11}{*}{Medium}
& \textbf{7.44(0.34)} & 6.67(0.56)$-$
& \textbf{4.98(0.48)} & 4.86(0.33)$=$ \\

Pusher-v1
&
& \textbf{2.85(0.86)} & 2.48(0.87)$=$
& 0.72(0.42) & \textbf{1.06(0.27)}$=$ \\

Thrower-v0
&
& \textbf{1.77(0.40)} & 1.75(0.36)$=$
& 0.55(0.14) & \textbf{1.46(0.17)}$+$ \\

Climber-v0
&
& 0.47(0.02) & 0.47(0.02)$=$
& 0.18(0.10) & \textbf{0.23(0.01)}$=$ \\

Climber-v1
&
& \textbf{0.53(0.17)} & 0.49(0.11)$=$
& 0.26(0.01) & \textbf{0.34(0.05)}$=$ \\

UpStepper-v0
&
& \textbf{4.94(0.62)} & 4.52(0.55)$=$
& 2.61(1.42) & \textbf{3.45(0.43)}$+$ \\

ObstacleTraverser-v0
&
& 6.78(1.93) & \textbf{6.93(1.93)}$=$
& 2.11(0.58) & \textbf{6.61(2.85)}$+$ \\

CaveCrawler-v0
&
& \textbf{5.57(1.06)} & 4.18(1.09)$-$
& \textbf{5.23(0.99)} & 4.11(0.13)$-$ \\

AreaMinimizer-v0
&
& \textbf{1.06(0.16)} & 1.01(0.10)$=$
& \textbf{0.87(0.08)} & 0.71(0.01)$-$ \\

HeightMaximizer-v0
&
& \textbf{0.35(0.01)} & 0.32(0.02)$=$
& 0.18(0.03) & \textbf{0.26(0.04)}$=$ \\

Balancer-v1
&
& 0.57(0.06) & \textbf{0.64(0.11)}$+$
& 0.31(0.17) & \textbf{0.59(0.01)}$+$ \\

\hline

Carrier-v1
& \multirow{9}{*}{Hard}
& \textbf{3.65(0.01)} & 3.52(0.19)$=$
& 2.47(0.13) & \textbf{3.55(0.19)}$+$ \\

Catcher-v0
&
& \textbf{0.72(0.08)} & 0.70(0.10)$=$
& \textbf{0.21(0.08)} & -3.66(0.62)$-$ \\

BeamSlider-v0
&
& \textbf{2.93(0.23)} & 2.40(0.37)$-$
& 1.82(0.19) & \textbf{2.32(0.55)}$+$ \\

Lifter-v0
&
& \textbf{0.43(0.29)} & 0.27(0.32)$-$
& -0.03(0.48) & \textbf{0.34(0.40)}$+$ \\

Climber-v2
&
& \textbf{1.83(0.91)} & 1.26(0.37)$-$
& \textbf{1.69(0.96)} & 1.28(0.43)$=$ \\

ObstacleTraverser-v1
&
& \textbf{4.27(0.65)} & 3.64(1.22)$-$
& 1.18(0.29) & \textbf{2.41(0.82)}$+$ \\

Hurdler-v0
&
& \textbf{4.09(0.91)} & 3.15(0.45)$-$
& 1.18(0.51) & \textbf{2.15(0.45)}$=$ \\

PlatformJumper-v0
&
& \textbf{3.93(1.96)} & 3.18(2.02)$-$
& 2.83(0.16) &2.83(0.22)$=$\\

GapJumper-v0
&
& 6.07(0.86) & \textbf{6.19(0.95)}$=$
& 1.99(0.86) & \textbf{3.42(0.45)}$+$ \\

\hline

\textbf{$+$/$-$/$=$}
&
& - & 2/10/18
& - & 13/5/12 \\

\bottomrule
\end{tabular}
\end{table}